%% file: main.tex
\documentclass[10pt,twocolumn,letterpaper]{article}

\usepackage{iccv}              % To produce the CAMERA-READY version
\definecolor{iccvblue}{rgb}{0.21,0.49,0.74}
\usepackage[pagebackref,breaklinks,colorlinks,allcolors=iccvblue]{hyperref}

\usepackage{multirow}
\usepackage{booktabs}
\usepackage{algorithm}
\usepackage{algcompatible}
\usepackage{pifont}

\def\paperID{13192} % *** Enter the Paper ID here
\def\confName{ICCV}
\def\confYear{2025}

\title{DEFNet: Multitasks-based Deep Evidential Fusion Network for Blind Image Quality Assessment}

\author{Yiwei Lou, Yuanpeng He, Rongchao Zhang, Yongzhi Cao, Hanpin Wang, Yu Huang\\
Key Laboratory of High Confidence Software Technologies (Peking University), Ministry of Education;\\
School of Computer Science, Peking University
}

\begin{document}
\maketitle
\input{sec/0_abstract}

\input{sec/1_intro}

\input{sec/3_preliminary}
\input{sec/4_method}

\input{sec/5_experiments}

\input{sec/6_conclusion}

\clearpage
{
    \small
    \bibliographystyle{ieeenat_fullname}
    \bibliography{main}
}

\input{sec/7_supplementary}

\end{document}

%% file: sec/0_abstract.tex
\begin{abstract}
    Blind image quality assessment (BIQA) methods often incorporate auxiliary tasks to improve performance. However, existing approaches face limitations due to insufficient integration and a lack of flexible uncertainty estimation, leading to suboptimal performance. To address these challenges, we propose a multitasks-based \textbf{D}eep \textbf{E}vidential \textbf{F}usion \textbf{Net}work (DEFNet) for BIQA, which performs multitask optimization with the assistance of scene and distortion type classification tasks. To achieve a more robust and reliable representation, we design a novel trustworthy information fusion strategy. It first combines diverse features and patterns across sub-regions to enhance information richness, and then performs local-global information fusion by balancing fine-grained details with coarse-grained context. Moreover, DEFNet exploits advanced uncertainty estimation technique inspired by evidential learning with the help of normal-inverse gamma distribution mixture. Extensive experiments on both synthetic and authentic distortion datasets demonstrate the effectiveness and robustness of the proposed framework. Additional evaluation and analysis are carried out to highlight its strong generalization capability and adaptability to previously unseen scenarios.
\end{abstract}

%% file: sec/1_intro.tex
\section{Introduction}
Blind image quality assessment (BIQA) is a pivotal area in the field of image processing. Its primary goal is to objectively and consistently assess the quality of images without relying on reference images for comparison. 
The pursuit of more accurate and efficient methods helps to improve the overall quality of experience for end-users.
This technique is of great importance in a wide range of application areas, such as real-time multimedia processing \cite{guo2025asymmetric,bai2025lensnet,guo2025underwater,yi2025mac,underwater2025guo,chen2025high} and medical image analysis \cite{chen2025semi,lou2024mr,lou2024no,zhang2024curriculum,qin2022ensemble}.

\begin{figure}[t] 
    \centering
    \includegraphics[width=\linewidth]{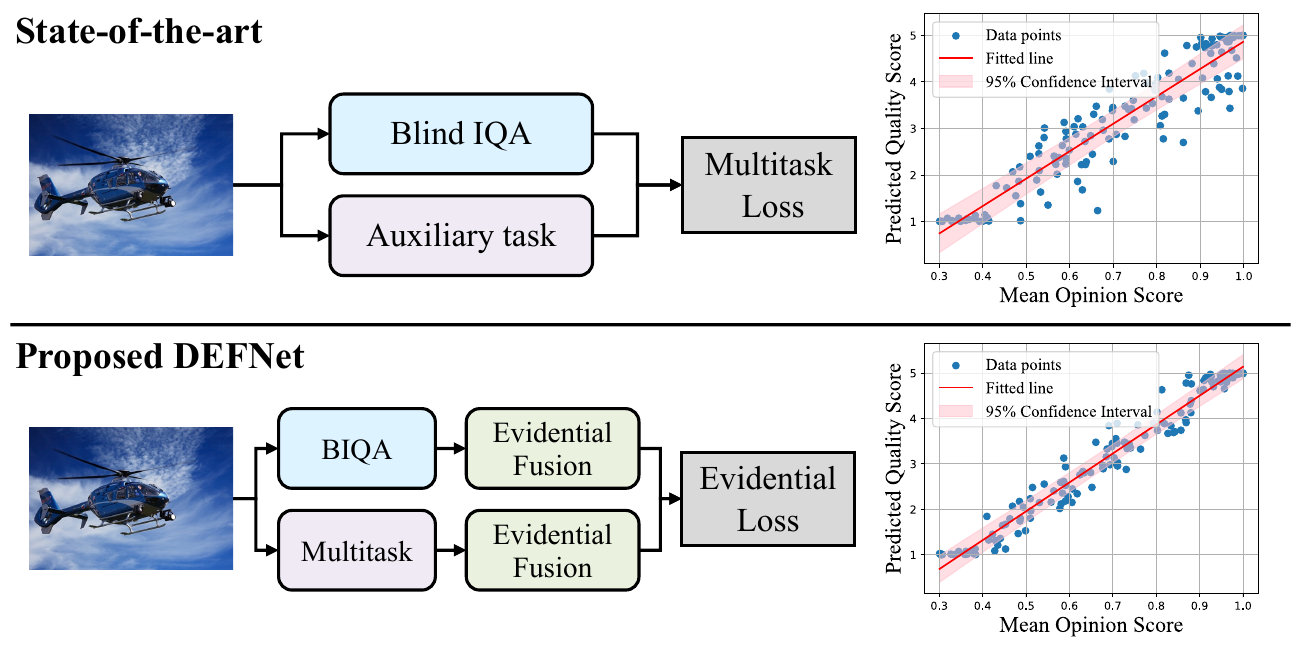} 
    \caption{Comparison between the proposed DEFNet and state-of-the-art methods that utilize auxiliary tasks to assist in BIQA. We propose to include evidential fusion for each task for higher performance and lower uncertainty.}
    \label{fig:intro} 
    \vspace{-0.3cm}
\end{figure}

Over time, BIQA approaches have undergone a significant evolution from early techniques based on handcrafted feature extraction and manual characterization \cite{6353522,mittal2012no} to more sophisticated data-driven and deep learning-based approaches \cite{9796010,Saha_2023_CVPR,lou2023refining,Golestaneh_2022_WACV,Su_2020_CVPR,10440553}. Nonetheless, these methods primarily focus on the assessment of image quality, which limits the assistance of auxiliary tasks and information. Inspired by this, efforts have been made to incorporate auxiliary tasks and information in a multitask learning manner, as shown in Figure \ref{fig:intro}(a).
For instance, scene statistics \cite{8666733} and image content \cite{LI2022307} offer valuable contextual information that can significantly influence the quality perception. Besides, distortion type and degree classification \cite{9796010} and spatial angular estimation \cite{9505016} provide inspiration as auxiliary tasks that enable more accurate assessment of image quality.
These methods typically utilize image content (scene information) and artifact categories (distortion information) to provide complementary insights and knowledge.

Despite these advances, existing methods still face challenges in two major aspects.
\textbf{(i) In-depth information fusion.} 
On one hand, this requires \ding{182} \textit{inter-task information integration}. Some existing approaches treat auxiliary tasks as independent modules, leading to information fragmentation and a lack of in-depth mining of potential inter-task correlations. 
On the other hand, it necessitates \ding{183} \textit{multilevel and cross-region feature fusion}, which involves full considering of the complex interactions between features and exploring diverse sub-regions that may contain different distortion patterns and visual characteristics.
\textbf{(ii) Comprehensive uncertainty estimation.} Though significant progress \cite{9369977,7937920,huang2019convolutional} has been made in uncertainty estimation for BIQA, it is still difficult to provide a \ding{184} \textit{flexible and robust uncertainty representation}.
A key limitation is the inability to simultaneously model both aleatoric and epistemic uncertainty, which often results in overconfident predictions even when the predictions are not correct.

To address these challenges, we propose a multitasks-based \textbf{D}eep \textbf{E}vidential \textbf{F}usion \textbf{Net}work (DEFNet) in this paper.
Our framework integrates three core tasks: BIQA, scene classification, and distortion type classification. It starts by utilizing contrastive language-image pre-training \cite{pmlr-v139-radford21a} to extract both local and global image features across the three different tasks, followed by a simultaneous multitask optimization to tackle challenge \ding{182}.
To further enhance feature fusion, we introduce a trustworthy information fusion strategy operating at two levels: cross sub-region and local-global. The cross sub-region fusion aggregates diverse features and patterns from different image sub-regions, thereby enhancing the information richness and ensuring accurate capture of regional quality. Meanwhile, the local-global fusion combines insights from both fine-grained details and coarse-grained context, providing a holistic understanding of image quality. This multilevel strategy facilitates in-depth information fusion and cross-region interactions, which serves as a solution to challenge \ding{183}. 
Furthermore, to address challenge \ding{184}, DEFNet incorporates a robust uncertainty estimation mechanism inspired by evidence theory \cite{NEURIPS2020_aab08546}. By utilizing the four dimensions of the data distribution and the mixture of normal-inverse gamma distribution, this approach simultaneously captures both aleatoric and epistemic uncertainty, enabling the model to identify the predictive fluctuations. As a result, the proposed DEFNet achieves high adaptability and generalization capabilities in various experimental settings. 

The main contributions of this paper are summarized as follows:
\begin{itemize}
    \item We propose a novel multitask-based deep evidential fusion network for BIQA, which integrates scene classification and distortion type classification to enhance inter-task information fusion.
    \item We propose a two-level trustworthy information fusion strategy, including cross sub-region and local-global information fusion, which integrate cross-region and cross-grained features, respectively.
    \item We develop a robust uncertainty estimation mechanism based on evidential learning and normal-inverse gamma distribution mixture, thereby improving the model's performance and adaptability.
    \item Extensive experiments on both synthetic and authentic distortion datasets are carried out to demonstrate that DEFNet achieves state-of-the-art performance, as well as strong generalization ability.
\end{itemize}

%% file: sec/3_preliminary.tex
\begin{figure*}[htbp] 
\centering
\includegraphics[width=\linewidth]{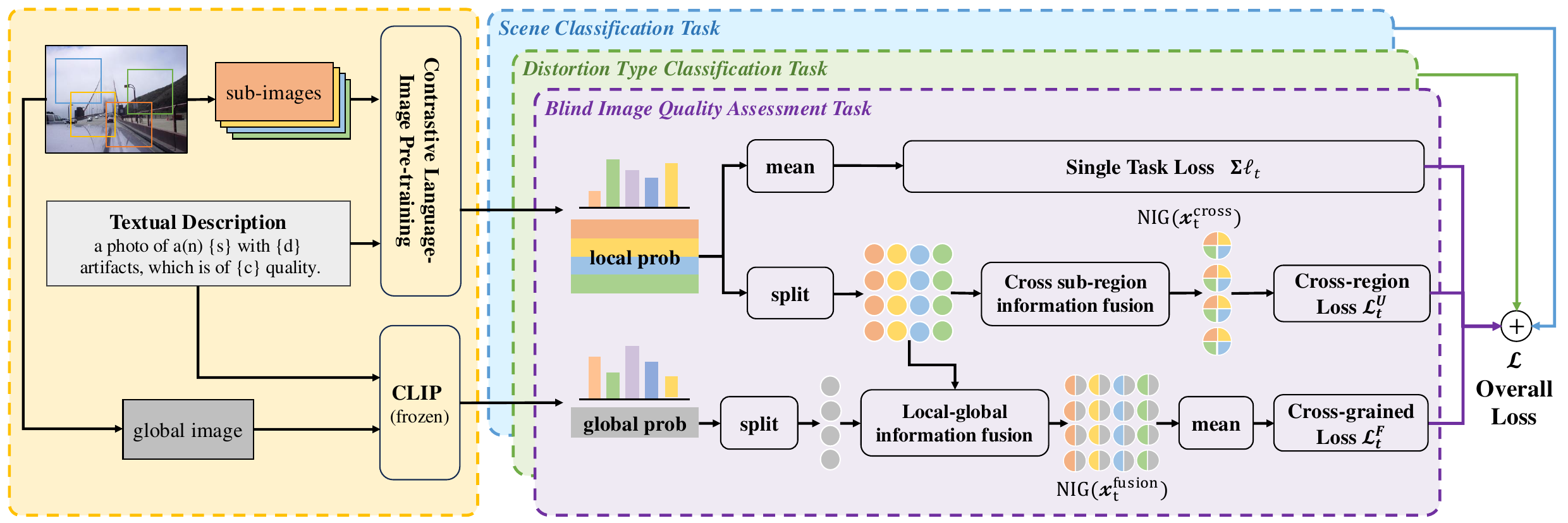} 
\caption{Overview of the proposed DEFNet framework.}
\label{fig:framework} 
\end{figure*}

\section{Problem Statement and Preliminaries}\label{sec:preliminaries}

To formalize the problem of blind image quality assessment, denote $\boldsymbol{x}\in \mathbb{R}^{C\times H\times W}$ as a pristine or distorted image, where $C, H, W$ are the channel number, height, and width, respectively. The goal of BIQA is to train a function $f: \mathbb{R}^{H\times W\times C} \rightarrow \mathbb{R}$ and estimate a quality score $q\in \mathbb{R}$ for the image $\boldsymbol{x}$ that reflects its perceptural quality, ideally aligning with human subjective evaluations.

Viewing the field of BIQA from the perspective of evidential learning, assume that the quality score $q$ of each image is subject to a normal distribution $q \sim \mathcal{N}(\mu, \sigma^2)$ where $\mu$ and $\sigma^2$ are the unknown mean and variance. A detailed justification of this assumption is provided in Supplementary \ref{sup:norm}.
The posterior distribution $p(\mu,\sigma|q(\boldsymbol{x}_{1\dots N}))$ is assumed to follow a normal-inverse gamma (NIG) distribution $(\mu, \sigma) \sim \operatorname{NIG}(\delta, v, \alpha, \beta)$, that is $\mu \sim \mathcal{N}(\delta, \sigma^2 v^{-1})$ and $\sigma^2 \sim \Gamma^{-1}(\alpha, \beta)$, where $\Gamma(\cdot)$ is gamma function, $\mathbf{m}=(\delta,v,\alpha,\beta)$ are distribution parameters with constraints $\delta\in\mathbb{R}, v>0, \alpha>1, \beta>0$. 
To increase the model evidence, denote $\Omega=2\beta(1+v)$, the negative logarithm of model evidence is denoted as:
\begin{equation}
\begin{aligned}
    \ell^{NLL}(\boldsymbol{x},\boldsymbol{y},\theta) =  \dfrac{1}{2} \log\left(\dfrac{\pi}{v}\right) + \log\left(\dfrac{\Gamma(\alpha)}{\Gamma(\alpha+\frac{1}{2})}\right)  \\
    - \alpha_t \log(\Omega) + (\alpha+\dfrac{1}{2}) \log\left((\boldsymbol{y}-\delta)^2 v + \Omega \right),
\end{aligned}
\end{equation}
where $\boldsymbol{x},\boldsymbol{y}$ are the input data and the ground-truth label.
To realign confidence in the predictions by reducing the evidence weight for predictions that deviate from expected values, the regression loss is defined as:
\begin{equation}
    \ell^R(\boldsymbol{x},\boldsymbol{y},\theta) = |\boldsymbol{y}-\mathbb{E}(\mu)| \cdot \phi,
\end{equation}
where $\phi=2v+\alpha$ is the total evidence \cite{NEURIPS2020_aab08546}. This realignment helps to improve the predictive acumen of the model, creating a more rigorous and robust framework for estimating the reasonableness of regression.
The total evidential loss aims to combine the term maximizing the model fit and the term minimizing evidence on errors:
\begin{equation}\label{eq:lu}
    \ell^U(\boldsymbol{x}, \boldsymbol{y}, \theta) = \ell^{NLL}(\boldsymbol{x},\boldsymbol{y},\theta) + \tau \ell^R(\boldsymbol{x},\boldsymbol{y},\theta),
\end{equation}
where $\tau$ is the weights keeping the balance between model fitting and uncertainty inflation.

%% file: sec/4_method.tex
\section{Methodology}\label{sec:method}
This section introduces the proposed DEFNet framework based on multitasks for BIQA. As shown in Figure~\ref{fig:framework}, the proposed framework initiates by extracting feature embeddings and probability scores from both local and global images contexts, and then performs single task optimization, as well as two levels (cross sub-region and local-global) of evidential fusion across all the three tasks. A complete algorithm description is shown in Supplementary \ref{sup:alg}.

\subsection{Local and Global Probability Scores}\label{sec:prob}
In the proposed DEFNet framework, we employ contrastive language-image pre-training (CLIP) \cite{pmlr-v139-radford21a} to extract the feature embeddings and compute both local and global probability scores. Specifically, the CLIP architecture consists of separate image and text encoders, which are trained by feeding multiple images and corresponding textual description (\textit{``a photo of a(n) \{s\} with \{d\} artifacts, which is of \{c\} quality."}), respectively. Detailed information on scenes ($s\in\mathcal{S}$), distortion types ($d\in\mathcal{D}$) and quality levels ($c\in\mathcal{C}$) in the text is given in the Supplementary \ref{sup:imp}.

Considering the prerequisite of the image encoder for inputs of a consistent size, local sub-images are obtained through cropping operation, while the global image is acquired after downsampling operation. This image segmentation approach allows to balance detail-oriented local analysis with a broader global perspective. It is worth mentioning that the training process of CLIP in the proposed DEFNet framework only consists of vision-language information pairs for local sub-images. The global feature embeddings are derived using the CLIP model pre-trained on these sub-images, with its parameters frozen to ensure stability and consistency in feature representation. From this, we have the correspondence score $\operatorname{logit}(c,s,d|\boldsymbol{x})$.
Subsequently, DEFNet performs softmax activation to derive the joint probability
\begin{equation}\label{eq:prob}
    \hat{p}(c,s,d)(\boldsymbol{x}) = \dfrac{\operatorname{exp}(\operatorname{logit}(c,s,d|\boldsymbol{x}) / \kappa)}{\sum_{c,s,d}\operatorname{exp}(\operatorname{logit}(c,s,d|\boldsymbol{x}) / \kappa)},
\end{equation}
where $\kappa$ is a learnable parameter, $c,s,d$ indicate the quality class, scene and distortion type, respectively. 
After that, the local probability scores $\hat{p}(c,s,d|\boldsymbol{x}^{\operatorname{local}})$ and global scores $\hat{p}(c,s,d|\boldsymbol{x}^{\operatorname{global}})$ are derived. 

With the assistant of the local probability scores, the quality score of an image is further estimated as:
\begin{equation}\label{eq:qpred}
    \hat{q}(\boldsymbol{x}) = \sum_{c=1}^C \hat{p}(c|\boldsymbol{x}) \times c,
\end{equation}
where $C=5$ and $c\in \mathcal{C}=\{1,2,3,4,5\}$ indicates the quality level from bad to perfect, and the estimated probability of the quality level is calculated by aggregating all the local scores
\begin{equation}
    \hat{p}(c|\boldsymbol{x}) = \operatorname{AVG}_{i=1}^N\left( \sum_{s\in\mathcal{S},d\in\mathcal{D}} \hat{p}(c,s,d)(\boldsymbol{x}_i^{\operatorname{local}})\right), 
\end{equation}
where $N$ is the number of sub-images, $\operatorname{AVG}(\cdot)$ is averaging operation for the local scores.

\subsection{Multitask Optimization}\label{sec:multitask}

In the multitask optimization framework, BIQA is the primary task represented by the loss component $\ell_q$, while components $\ell_s$ and $\ell_d$ correspond to the auxiliary tasks of scene and distortion type classification, respectively. Each loss component, with specific definition as follows, contributes uniquely to the overall multitask loss.

By adopting the fidelity loss \cite{tsai2007frank}, the BIQA loss for image pair $(\boldsymbol{x}_1, \boldsymbol{x}_2)$ is defined as:
\begin{equation}\label{eq:ellq}
\begin{aligned}
    \ell_q(\boldsymbol{x}_1, \boldsymbol{x}_2; \theta) = & 1 - \sqrt{p(\boldsymbol{x}_1, \boldsymbol{x}_2) \hat{p}(\boldsymbol{x}_1, \boldsymbol{x}_2)}  \\
    & - \sqrt{(1 - p(\boldsymbol{x}_1, \boldsymbol{x}_2))(1-\hat{p}(\boldsymbol{x}_1, \boldsymbol{x}_2))},
\end{aligned}
\end{equation}
where 
\begin{equation}
    \hat{p}(\boldsymbol{x}_1, \boldsymbol{x}_2) = \Phi \left( \dfrac{\hat{q}(\boldsymbol{x}_1) - \hat{q}(\boldsymbol{x}_2)}{\sqrt{2}} \right)
\end{equation}
quantifies the likelihood that $\boldsymbol{x}_1$ is of higher predicted quality than $\boldsymbol{x}_2$ using standard Normal cumulative distribution function $\Phi(\cdot)$ under the Thurstone’s model \cite{thurstone2017law}, and $p(\boldsymbol{x}_1, \boldsymbol{x}_2)$ is a binary label indicating whether the ground-truth MOS $q(\boldsymbol{x}_1) \geq q(\boldsymbol{x}_2)$.

In the settings of DEFNet, an image can be associated with multiple scene categories. Given an image $\boldsymbol{x}$, the estimated probability of a scene $s$  is calculated by aggregating the joint probabilities across all possible quality and distortion combinations:
\begin{equation}
    \hat{p}(s|\boldsymbol{x}) = \sum_{c,d} \hat{p}(c,s,d|\boldsymbol{x}),
\end{equation}
where $\hat{p}(c,s,d|\boldsymbol{x})$ is the joint probability derive in Equation (4). Based on this, the scene classification loss component is defined as:
\begin{equation}\label{eq:scene}
\begin{aligned}
    \ell_s(\boldsymbol{x}; \theta) = \frac{1}{|\mathcal{S}|} \sum_{s \in \mathcal{S}} \Big( & 1 - \sqrt{p(s|\boldsymbol{x}) \hat{p}(s|\boldsymbol{x})} \\
    & - \sqrt{(1-p(s|\boldsymbol{x})) (1-\hat{p}(s|\boldsymbol{x}))} \Big),
\end{aligned}
\end{equation}
where $\mathcal{S}$ is the set of all possible scene categories, $p(s|\boldsymbol{x})$ is a binary label indicating whether the image $\boldsymbol{x}$ falls in ground-truth scene category $s$.

Similar but different, we assume each image only belongs to one dominant distortion type, and we have the predicted probability for specific distortion type $d$ as:
\begin{equation}
    \hat{p}(d|\boldsymbol{x}) = \sum_{c,s} \hat{p}(c,s,d|\boldsymbol{x}),
\end{equation}
and further define the distortion type classification loss as:
\begin{equation}\label{eq:dist}
    \ell_d(\boldsymbol{x}; \theta) = 1 - \sum_{d\in\mathcal{D}} \sqrt{p(d|\boldsymbol{x}) \hat{p}(d|\boldsymbol{x})},
\end{equation}
where $\mathcal{D}$ is the set of all possible distortion types, $p(d|\boldsymbol{x})$ is the binary ground-truth label, and $\hat{p}(d|\boldsymbol{x})$ is the predicted probability for image $\boldsymbol{x}$ belongs to type $d$. 

Utilizing auxiliary tasks, DEFNet optimizes losses of the three separate tasks, integrating them into the multitask loss \cite{Zhang_2023_CVPR}, which in a mini-batch $\mathcal{B}$ is defined as
\begin{equation}\label{eq:lm}
\begin{aligned}
    \mathcal{L}^M (\theta) = & \dfrac{1}{|\mathcal{P}|} \sum_{(\boldsymbol{x}_1,\boldsymbol{x}_2) \in \mathcal{P}} \lambda_q \ell_q(\boldsymbol{x}_1,\boldsymbol{x}_2; \theta)  \\
    & + \dfrac{1}{|\mathcal{B}|} \sum_{\boldsymbol{x} \in \mathcal{B}} \big[ \lambda_s \ell_s(\boldsymbol{x}) + \lambda_d \ell_d(\boldsymbol{x}) \big],
\end{aligned}
\end{equation}
where $\theta$ is the model parameter, $\mathcal{P}$ denotes the set of all possible image pairs with ground-truth quality label, $\lambda_q, \lambda_s, \lambda_d$ are weights updated with the relative descending rate \cite{Liu_2019_CVPR}.

\begin{figure}[t] 
    \centering
    \includegraphics[width=\linewidth]{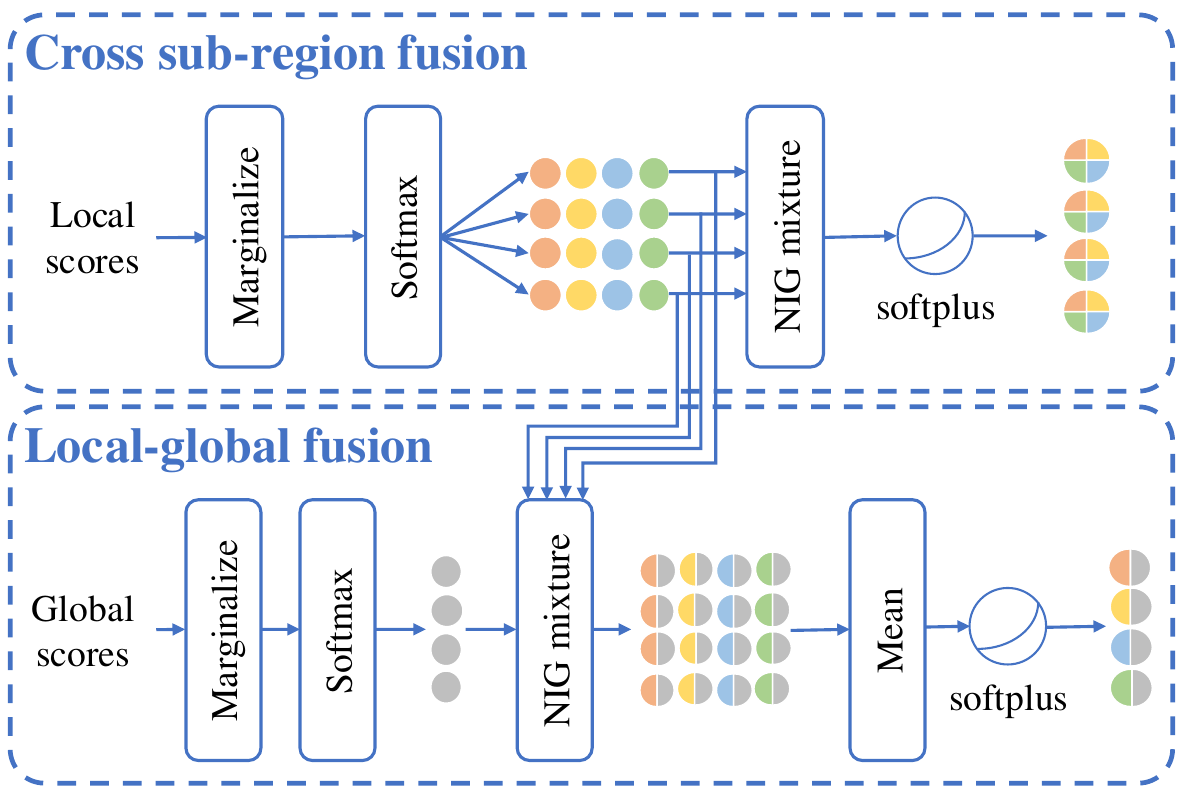} 
    \caption{Overview of the cross sub-region information fusion (top) and local-global information fusion (bottom).}
    \label{fig:fusion} 
\end{figure}

\subsection{Cross Sub-region Information Fusion}\label{sec:cross}
In this section, we introduce the technique of cross sub-region evidential information fusion. As shown in Figure~\ref{fig:fusion}, it integrates fragmented information from different sub-regions, allowing the model to make predictions in a more comprehensive way and decrease aleatoric and epistemic uncertainty. Through fusion across sub-regions, the DEFNet framework integrates diverse features and patterns from different regions effectively, which is critical in dealing with complex images and diverse content. This helps to capture the differences in quality across sub-regions in an image more accurately, thus improving the accuracy of the assessment at a detailed level.

To estimate the parameters of NIG distribution, we randomly sample probability scores $\hat{p}(c,s,d|\boldsymbol{x}_{i}^{\operatorname{local}})$ of four sub-images $i\in\{1,2,3,4\}$.
Subsequently, we marginalize the probability scores for the three specific tasks ($q$ for BIQA task, $s$ for scene classification task, and $d$ for distortion type classification task):
\begin{equation}\label{eq:xqilocal}
    \boldsymbol{x}^{\operatorname{local}}_{q,i} = \operatorname{softplus} \Big(  \sum_{c=1}^{C}  \Big[\sum_{s,d} \hat{p}(c,s,d|\boldsymbol{x}^{\operatorname{local}}_{i}) \times c \Big]\Big),
\end{equation}
\begin{equation}\label{eq:xsilocal}
    \boldsymbol{x}^{\operatorname{local}}_{s,i} = \operatorname{softplus} \Big( \sum_{q,d} \hat{p}(c,s,d|\boldsymbol{x}^{\operatorname{local}}_{i}) \Big),
\end{equation}
\begin{equation}\label{eq:xdilocal}
    \boldsymbol{x}^{\operatorname{local}}_{d,i} = \operatorname{softplus} \Big( \sum_{q,s} \hat{p}(c,s,d|\boldsymbol{x}^{\operatorname{local}}_{i}) \Big),
\end{equation}
where $\operatorname{softplus}$ is the activation function to satisfy parameter constraints, $C=5$ is the number of quality levels. 
The distribution parameters can be computed as follows:
\begin{equation}
\begin{aligned}
    \boldsymbol{m}^{\operatorname{local}}_{t,i} 
    & = (\boldsymbol{x}^{\operatorname{local}}_{t,i})_{\delta}, (\boldsymbol{x}^{\operatorname{local}}_{t,i})_{v}, (\boldsymbol{x}^{\operatorname{local}}_{t,i})_{\alpha}, (\boldsymbol{x}^{\operatorname{local}}_{t,i})_{\beta}  \\
    & = \operatorname{split}( \boldsymbol{x}^{\operatorname{local}}_{t,i} ) ,
\end{aligned}
\end{equation}
where $t\in\{q,s,d\}$ denote the task domain.
Then, we adopt the fusion strategy to fuse multiple NIG distribution and to integrate the inter sub-region information extracted from the four sub-images:
\begin{equation}\label{eq:nigadd}
\begin{aligned}
    \operatorname{NIG}(\boldsymbol{x}_{t}^{\operatorname{cross}}) = & \operatorname{NIG}(\boldsymbol{m}_{t,1}^{\operatorname{local}}) \oplus \operatorname{NIG}(\boldsymbol{m}_{t,2}^{\operatorname{local}}) \\
    & \oplus \operatorname{NIG}(\boldsymbol{m}_{t,3}^{\operatorname{local}}) \oplus \operatorname{NIG}(\boldsymbol{m}_{t,4}^{\operatorname{local}}) ,
\end{aligned}
\end{equation}
where $\boldsymbol{x}^{\operatorname{cross}}_{t}$ is the mixture NIG distribution parameters, $\oplus$ is the summation operation for two NIG distributions \cite{NEURIPS2021_371bce7d}, which is defined as:
\begin{equation}
    \operatorname{NIG}(\delta,v,\alpha,\beta) \triangleq \operatorname{NIG}(\delta_1,v_1,\alpha_1,\beta_1) \oplus \operatorname{NIG}(\delta_2,v_2,\alpha_2,\beta_2)
\end{equation}
where
\begin{equation}
\begin{aligned}
    \delta &= (v_1+v_2)^{-1} (v_1\delta_1 + v_2\delta_2), \\
    v &= v_1 + v_2, \;\; \alpha = \alpha_1 + \alpha_2 + \dfrac{1}{2}, \\
    \beta &= \beta_1 + \beta_2 + \dfrac{1}{2}v_1(\delta_1-\delta)^2 + \dfrac{1}{2}v_2(\delta_2-\delta)^2.
\end{aligned}
\end{equation}

Then, we compute the evidential loss on local outputs for single task $t$:
\begin{equation}\label{eq:lut}
    \mathcal{L}^U_t(\theta) = \dfrac{1}{|\mathcal{B}|} \sum_{\boldsymbol{x}\in\mathcal{B}}  \ell^U\left(\operatorname{softplus}( \boldsymbol{x}^{\operatorname{cross}}_t), \boldsymbol{y}_t, \theta\right),
\end{equation}
where $\boldsymbol{y}_q = q(\boldsymbol{x})$ is the ground-truth MOS, $\boldsymbol{y}_s = p(s|\boldsymbol{x})$ and $\boldsymbol{y}_d = p(d|\boldsymbol{x})$ are binary labels indicating whether the image $\boldsymbol{x}$ falls in ground-truth scene category $s$ and distortion type category $d$, respectively.
Then, the overall cross-region loss is defined as the sum of evidential loss for the three tasks:
\begin{equation}\label{eq:LU}
    \mathcal{L}^U(\theta) = \mathcal{L}^U_q(\theta) + \mathcal{L}^U_s(\theta) + \mathcal{L}^U_d(\theta).
\end{equation}

\subsection{Local-global Information Fusion}\label{sec:lgif}
In this section, we describe the evidential fusion between local and global information, the overall framework is shown in Figure~\ref{fig:fusion}. Local information focuses on fine-grained details within sub-images, while global information provides a coarse-grained perspective of the entire image. The local-global fusion allows them to complement each other effectively and enables DEFNet to combine the local view at a detailed level with a broader global view, providing a comprehensive assessment of image quality. This fusion strategy balances the fine-grained details with the coarse-grained whole, ensuring that DEFNet is neither overly focused on micro-details nor ignoring global perspectives.

To combine information from local sub-images and global downsampled image, we first marginalize the global probability scores $\boldsymbol{x}^{\operatorname{global}}_t$ for tasks $t\in\{q,s,d\}$, following the same approach as in Eq. (\ref{eq:xqilocal}), (\ref{eq:xsilocal}) and (\ref{eq:xdilocal}). The parameters of the global image distribution are computed as:
\begin{equation}
\begin{aligned}
    \boldsymbol{m}^{\operatorname{global}}_t
    & = (\boldsymbol{x}^{\operatorname{global}}_{t})_{\delta}, (\boldsymbol{x}^{\operatorname{global}}_{t})_{v},(\boldsymbol{x}^{\operatorname{global}}_{t})_{\alpha}, (\boldsymbol{x}^{\operatorname{global}}_{t})_{\beta},  \\
    & = \operatorname{split}( \boldsymbol{x}^{\operatorname{global}}_t ).
\end{aligned}
\end{equation}
Then, we employ the fusion strategy to merge local NIG distributions derived from each sub-images with global one:
\begin{equation}
\begin{aligned}
    & \operatorname{NIG}(\boldsymbol{m}_{t,i}^{\operatorname{fusion}}) = \operatorname{NIG}(\boldsymbol{m}_{t,i}^{\operatorname{local}}) \oplus \operatorname{NIG}(\boldsymbol{m}_t^{\operatorname{global}}),
\end{aligned}
\end{equation}
where $\boldsymbol{m}^{\operatorname{fusion}}_{t,i}$ represents the local-global parameters of the mixture NIG distribution between the $i$-th local sub-image and the global image in task $t$. The local-global fusion information is aggregated through averaging:
\begin{equation}\label{eq:xfusiont}
    \boldsymbol{x}^{\operatorname{fusion}}_t = \dfrac{1}{4} \sum_{i} \boldsymbol{m}^{\operatorname{fusion}}_{t,i}.
\end{equation}

Subsequently, we define the evidential loss based on local and global information fusion for single task $t$ as:
\begin{equation}\label{eq:lft}
    \mathcal{L}^F_t(\theta) = \dfrac{1}{|\mathcal{B}|} \sum_{\boldsymbol{x}\in\mathcal{B}}  \ell^U\left(\operatorname{softplus}( \boldsymbol{x}_{t}^{\operatorname{fusion}}),\boldsymbol{y}_t,\theta\right).
\end{equation}
The overall cross-grained loss based on local-global information fusion for multitasks is the sum of these evidential fusion losses for three tasks:
\begin{equation}\label{eq:LF}
    \mathcal{L}^{F}(\theta) = \mathcal{L}^F_q(\theta) + \mathcal{L}^F_s(\theta) + \mathcal{L}^F_d(\theta).
\end{equation}

\subsection{Overall Loss}\label{sec:overloss}
In the proposed DEFNet framework, the overall loss function is composed of multiple components, each targeting a specific aspect of the model performance. The overall loss is denoted as $\mathcal{L}(\theta)$ and contains the multitask loss $\mathcal{L}^M (\theta)$, the cross-region loss $\mathcal{L}^U (\theta)$ resulting from cross sub-region information fusion, and the cross-grained loss $\mathcal{L}^ F (\theta)$ from local-global information fusion. Formally, we have the optimization objective of the proposed DEFNet:
\begin{equation}\label{eq:overall}
    \mathcal{L}(\theta) = \mathcal{L}^M (\theta) + \lambda_1 \mathcal{L}^U (\theta) + \lambda_2 \mathcal{L}^F (\theta),
\end{equation}
where $\lambda_1$ and $\lambda_2$ are parameters that control the relative contribution of each loss component to the overall loss.

%% file: sec/5_experiments.tex
\begin{table*}[t]
    \centering
    \fontsize{9}{10}\selectfont %{字体尺寸}{行距}
    \caption{Performance comparison of the proposed approach and state-of-the-art methods on datasets with synthetic and authentic distortion. Best and second-best scores are highlighted in bold and underlined, respectively.}
    \resizebox{\linewidth}{!}{
    \begin{tabular}{l|cc|cc|cc|cc|cc|cc}
        \toprule
        \multirow{3}{*}{Method} 
        & \multicolumn{6}{c|}{Synthetic distortion} & \multicolumn{6}{c}{Authentic distortion}  \\
        \cmidrule(lr){2-7} \cmidrule(lr){8-13}
        & \multicolumn{2}{c}{LIVE} & \multicolumn{2}{c}{CSIQ} & \multicolumn{2}{c|}{KADID-10k}  & \multicolumn{2}{c}{BID} & \multicolumn{2}{c}{LIVE-C} & \multicolumn{2}{c}{KonIQ-10k}   \\
        & SRCC    & PLCC  & SRCC  & PLCC  & SRCC  & PLCC  & SRCC  & PLCC  & SRCC & PLCC  & SRCC & PLCC  \\
        \midrule
        NIQE \cite{6353522}      & 0.908 & 0.904 & 0.631 & 0.719 & 0.389 & 0.442 & 0.573 & 0.618 & 0.446 & 0.507 & 0.415 & 0.438  \\
        ILNIQE \cite{7094273}    & 0.887 & 0.894 & 0.808 & 0.851 & 0.565 & 0.611 & 0.548 & 0.494 & 0.469 & 0.518 & 0.509 & 0.534  \\
        dipIQ \cite{7934456}     & 0.940 & 0.933 & 0.511 & 0.778 & 0.304 & 0.402 & 0.009 & 0.346 & 0.187 & 0.290 & 0.228 & 0.437  \\
        Ma19 \cite{8803390}      & 0.922 & 0.923 & 0.926 & 0.929 & 0.465 & 0.501 & 0.373 & 0.399 & 0.336 & 0.405 & 0.360 & 0.398  \\
        \midrule
        DBCNN \cite{8576582}     & 0.963 & 0.966 & 0.940 & 0.954 & 0.878 & 0.878 & 0.864 & 0.883 & 0.835 & 0.854 & 0.864 & 0.868  \\
        HyperIQA \cite{Su_2020_CVPR}  & 0.966 & 0.968 & 0.934 & 0.946 & 0.872 & 0.869 & 0.848 & 0.868 & 0.855 & 0.878 & 0.900 & 0.915  \\
        UNIQUE \cite{9369977}    & 0.961 & 0.952 & 0.902 & 0.921 & 0.884 & 0.885 & 0.852 & 0.875 & 0.854 & 0.884 & 0.895 & 0.900  \\
        TreS \cite{Golestaneh_2022_WACV}      & 0.965 & 0.963 & 0.902 & 0.923 & 0.881 & 0.879 & 0.853 & 0.871 & 0.846 & 0.877 & 0.907 & 0.924  \\
        LIQE \cite{Zhang_2023_CVPR}      & 0.970 & 0.951 & 0.936 & 0.939 & 0.930 & 0.931 & \underline{0.875} & \underline{0.900} & \underline{0.904} & \textbf{0.910} & \underline{0.919} & 0.908 \\
        CONTRIQUE \cite{9796010} & 0.960 & 0.961 & 0.942 & 0.955 & 0.934 & 0.937 &  -    &   -   & 0.845 & 0.857 & 0.894 & 0.906  \\
        VCRNet \cite{pan2022vcrnet}   &  0.973  & \underline{0.974}  &  0.943  & 0.955 & 0.853 & 0.849 & -  &  -  &  0.856 & 0.865 & 0.894 & 0.909    \\
        Re-IQA \cite{Saha_2023_CVPR}    & 0.970 & 0.971 & 0.947 & 0.960 & 0.872 & 0.885 &  -    &   -   & 0.840 & 0.854 & 0.914 & 0.923  \\
        DPNet \cite{wang2023learning}  &  0.971 & 0.971 & 0.942 & 0.952 & 0.923 & 0.924 & - & - & 0.849 & 0.864 & - & -   \\
        QAL-IQA \cite{zhou2025blind} & 0.971 & 0.973 & \underline{0.963} & \textbf{0.970} & 0.908 & 0.910 & - & - & 0.859 & 0.875 & 0.917 & \textbf{0.928} \\
        CDINet \cite{10440553}        &  \underline{0.977}  &  \textbf{0.975}  &  0.952  & 0.960  &  0.920  &  0.919  &  0.874  & 0.899 & 0.865  & 0.880 & 0.916 & \textbf{0.928}  \\
        TOPIQ-FR \cite{chen2024topiq} & 0.887 & 0.882 & 0.894 & 0.894 & 0.895 & 0.896  & / & / & / & / & / & / \\
        KGANet \cite{zhou2024multitask} & 0.963 & 0.966 & 0.954 & 0.963 & \underline{0.940} & \underline{0.943} & / & / & / & / & / & /   \\
        CausalQuality-VGG \cite{shen2025image} &  0.932 & 0.929 & 0.952 & 0.949 & 0.899 & 0.898 & / & / & / & / & / & /   \\
        CausalQuality-EffNet \cite{shen2025image} &  0.932 & 0.927 & 0.938 & 0.933 & 0.907 & 0.905 & / & / & / & / & / & /   \\
        DPSF \cite{xia2025blind} & / & / & / & / & / & / & 0.872 & 0.883 & 0.865 & 0.882 & 0.912 & \underline{0.925}  \\
        \midrule
        % \rowcolor{tabhighlight}
        DEFNet      & \textbf{0.978} & 0.960 & \textbf{0.967} & \underline{0.964} & \textbf{0.942} & \textbf{0.944} & \textbf{0.910} & \textbf{0.909} & \textbf{0.918} & \underline{0.897} & \textbf{0.920} & 0.901  \\
        \bottomrule
    \end{tabular}}
    \label{tab:main_results}
\end{table*}

\section{Experiments}\label{sec:exp}
\subsection{Experimental Setups}
We conduct evaluation on both synthetic and authentic distorted datasets. The former includes LIVE \cite{1709988}, CSIQ \cite{10.1117/1.3267105} and KADID-10k \cite{8743252}, while the latter consists of BID \cite{5492198}, LIVE-C \cite{7327186} and KonIQ-10k \cite{8968750}.
Additional experiments are conducted in the TID2013 \cite{PONOMARENKO201557}, SPAQ \cite{Fang_2020_CVPR}, PIPAL \cite{jinjin2020pipal}, and Waterloo exploration database (WED) \cite{7752930}.
Each dataset is randomly divided into training, validation and test sets in the ratio of 70\%, 10\%, 20\% across ten sessions. 
The performance is evaluated using Spearman’s rank order correlation coefficient (SRCC) and Pearson’s linear correlation coefficient (PLCC).

\subsection{Implementation Details}
Within the CLIP, we employ ViT-B/32 \cite{pmlr-v139-radford21a} as the visual encoder and GPT-2 base model \cite{radford2019language} as the text encoder. We train the uncertainty-based evidential loss in Eq. (\ref{eq:lu}) with weights $\tau=0.05$. For the training phase, we initialize the learning rate to $5e-6$ and train the model for a total of 80 epochs. The mini-batch size is set to 48, with 4 samples from each of the LIVE, CSIQ, BID, and LIVE-C datasets, and 16 samples from both the KADID-10k and KonIQ-10k datasets. Throughout the training and inference processes, we perform random cropping to obtain 4 and 15 sub-images from the raw input images, respectively. Each sub-image is with a fixed size of $3\times 224\times 224$. All experiments are conducted with one NVIDIA RTX 4090 GPU.

\subsection{Model Performance}
To evaluate the effectiveness of the proposed DEFNet framework, we compare it to four knowledge-driven MOS-free BIQA models including NIQE \cite{6353522}, ILNIQE \cite{7094273}, dipIQ \cite{7934456} and Ma19 \cite{8803390}, as well as neural network-based methods, including DBCNN \cite{8576582}, HyperIQA \cite{Su_2020_CVPR}, UNIQUE \cite{9369977}, TreS \cite{Golestaneh_2022_WACV}, LIQE \cite{Zhang_2023_CVPR}, CONTRIQUE \cite{9796010}, VCRNet \cite{pan2022vcrnet}, Re-IQA \cite{Saha_2023_CVPR}, DPNet \cite{wang2023learning}, QAL-IQA \cite{zhou2025blind}, CDINet \cite{10440553}, TOPIQ-FR \cite{chen2024topiq}, KGANet \cite{zhou2024multitask}, CausalQuality \cite{shen2025image} and DPSF \cite{xia2025blind}.
The experimental results in terms of SRCC and PLCC are shown in Table~\ref{tab:main_results}, where we draw several conclusions.
First, DEFNet exhibits outstanding performance on both synthetic and authentic distortion datasets compared to existing methods. The superior performance can be attributed to the multilevel information fusion strategy, which effectively integrates quality features and patterns across sub-regions while maintaining a balance between detailed and global perspectives.
Second, LIQE \cite{Zhang_2023_CVPR} and CDINet \cite{10440553} demonstrate promising performance, especially on datasets that are relatively small and with basic distortions.
Third, assessing the image quality of authentic distorted scenarios is more difficult than for synthetic distorted scenarios. This holds for most methods and is the general agreement in the field of BIQA.

\begin{figure}[t] 
    \centering
    \includegraphics[width=\linewidth]{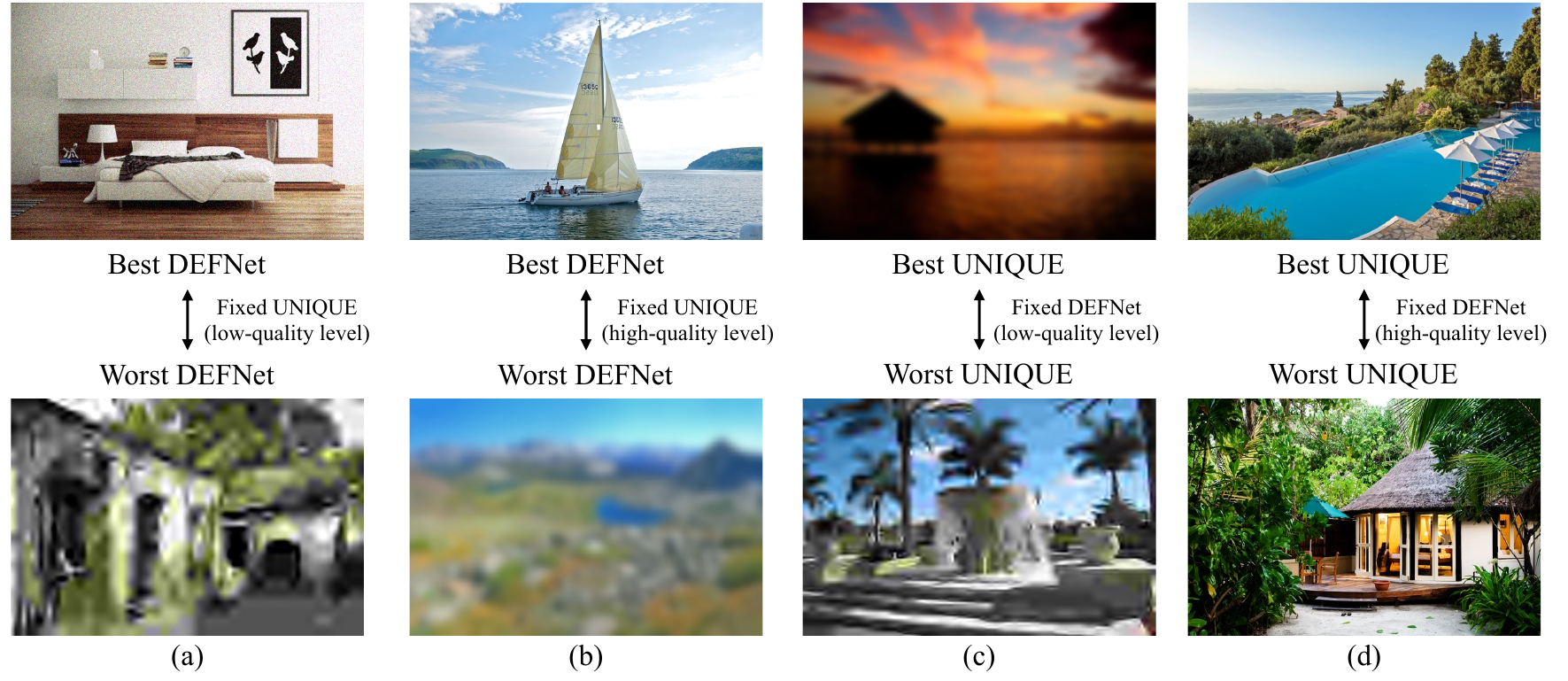} 
    \caption{gMAD competition between UNIQUE and DEFNet in WED. 
    (a) Fixed UNIQUE at low-quality level. (b) Fixed UNIQUE at high-quality level. (c) Fixed DEFNet at low-quality level. (d) Fixed DEFNet at high-quality level.}
    \label{fig:gmad_wed} 
\end{figure}

\subsection{gMAD Competition}\label{sec:gmad}

To illustrate DEFNet's ability to effectively reintegrate information from different IQA datasets into a shared perceptual scale, gMAD competition \cite{8590800} are carried out in WED \cite{7752930}, which is representative of synthetic distortion dataset. The gMAD framework presents qualitative differences by fixing one method's prediction while varying the quality identified by another. This setup highlights scenarios where each model performs well or poorly, providing a deeper view of their strengths and weaknesses.
As shown in Figure \ref{fig:gmad_wed}, the highest preditions of DEFNet align well with high-quality images, while the worst output appropriately reflects severe distortion. In contrast, UNIQUE shows a tendency to misclassify certain low-quality and high-quality images. DEFNet demonstrates superior consistency in ranking high-quality and low-quality images.
This indicates that DEFNet not only consistently produces more reliable quality assessment, but also offers strong generalization capability.
Further results and analysis in SPAQ (representative of authentic distortion dataset) are given in Supplementary \ref{sup:gmad}.

\begin{table}[t]
    \centering
    \fontsize{9}{10}\selectfont %{字体尺寸}{行距}
    \caption{SRCC performance in cross-dataset evaluation under zero-shot setting. The subscripts ``d'' and ``q'' represent that the model is trained on KADID-10k and KonIQ-10k, respectively.}
    \begin{tabular}{c|ccc}
        \toprule
        Training   & TID2013  &   SPAQ    &   PIPAL  \\
        \midrule
        NIQE \cite{6353522}            &   0.314  &   0.578   &   0.153  \\
        DBCNN$_d$ \cite{8576582}       &   0.471  &   0.801   &   0.413  \\
        DBCNN$_q$ \cite{8576582}       &   0.686  &   0.412   &   0.321  \\
        PaQ2PiQ \cite{Ying_2020_CVPR}  &   0.423  &   0.823   &   0.400  \\
        MUSIQ \cite{Ke_2021_ICCV}      &   0.584  &   0.853   &   0.450  \\
        UNIQUE \cite{9369977}          &   0.768  &   0.838   &   0.444  \\
        % LIQE       &   0.811  &   0.881   &   0.478  \\
        \midrule
        DEFNet     &   \textbf{0.828}  &   \textbf{0.868}   &  \textbf{0.464} \\
        \bottomrule
    \end{tabular}
    \label{tab:cross}
\end{table}

\subsection{Cross-Dataset Evaluation}
To evaluate the generalization capability, we conduct cross-dataset evaluation in a zero-shot setting, following the approach outlined in previous works \cite{Zhang_2023_CVPR,9369977}. The experiments are performed on the TID2013 \cite{PONOMARENKO201557}, SPAQ \cite{Fang_2020_CVPR} and PIPAL training set \cite{jinjin2020pipal} datasets. As shown in Table~\ref{tab:cross}, DEFNet demonstrates high robustness in TID2013 and SPAQ, achieving SRCC values of 0.828 and 0.868, respectively. These results highlight the model's strong ability to generalize effectively to unseen datasets with both synthetic and authentic distortions, outperforming existing methods. 
However, the model's performance on PIPAL, while competitive, is comparatively lower with an SRCC of 0.464. This indicates that its generalization to highly diverse and novel distortions remains a challenge.

\begin{table}[t]
    \centering
    \fontsize{9}{10}\selectfont %{字体尺寸}{行距}
    \caption{SRCC performance on diverse distortion types of CSIQ.}
    \resizebox{\linewidth}{!}{
    \begin{tabular}{c|cccccc}
        \toprule
        Distortion & WN & GB & PN & CD & JPEG & JP2K \\
        \midrule
        BRISQUE \cite{mittal2012no}           & 0.682 & 0.808 & 0.743 & 0.396 & 0.846 & 0.817   \\
        ILNIQE \cite{7094273}                 & 0.850 & 0.858 & 0.874 & 0.501 & 0.899 & 0.906   \\
        deepIQA \cite{bosse2016deep}          & 0.944 & 0.901 & 0.867 & 0.847 & 0.922 & 0.934   \\
        DBCNN \cite{8576582}                  & 0.948 & 0.947 & 0.941 & 0.872 & 0.940 & 0.953 \\
        HyperIQA \cite{Su_2020_CVPR}          & 0.927 & 0.915 & 0.931 & 0.874 & 0.934 & 0.960 \\
        DCNet \cite{10.1145/3503161.3547982}  & \underline{0.964} & \underline{0.968} & \underline{0.958} & \underline{0.931} & \textbf{0.972} & \underline{0.966} \\
        OLNet \cite{9718583}                  & 0.945 & 0.965 & 0.953 & 0.925 & \underline{0.968} & 0.945 \\
        VCRNet \cite{pan2022vcrnet}           & 0.939 & 0.950 & 0.899 & 0.919 & 0.956 & 0.962   \\
        \midrule
        \textbf{DEFNet}                       & \textbf{0.969} & \textbf{0.971} & \textbf{0.974} & \textbf{0.941} & 0.967 &  \textbf{0.971} \\
        \bottomrule
    \end{tabular}}
    \label{tab:distortion_type_csiq}
\end{table}

\begin{table}[t]
    \centering
    \caption{Mean correlation coefficients (SRCC and PLCC) and mean accuracy (ACC) on the six datasets. The subscripts ``$s$'' and ``$d$'' stands for accuracy of the scene and distortion type classification tasks, respectively. Best scores are highlighted in bold.}
    \label{tab:ablation}
    \resizebox{\linewidth}{!}{
    \begin{tabular}{c|ccc|cccc}
        \toprule
        \multirow{2}{*}{Task} & \multicolumn{3}{c|}{Loss component} & \multirow{2}{*}{SRCC} & \multirow{2}{*}{PLCC} & \multirow{2}{*}{$\operatorname{ACC}_s$} & \multirow{2}{*}{$\operatorname{ACC}_d$} \\
        &  $\mathcal{L}^M$  &  $\mathcal{L}^U$  &  $\mathcal{L}^F$   &  \\
        \midrule
        \multirow{4}{*}{$q$} 
        & \checkmark &            &            & 0.910 & 0.898 & - & - \\
        & \checkmark & \checkmark &            & 0.914 & 0.905 & - & - \\
        & \checkmark &            & \checkmark & 0.916 & 0.905 & - & - \\
        & \checkmark & \checkmark & \checkmark & 0.922 & 0.908 & - & - \\
        \midrule
        \multirow{4}{*}{$q+s$} 
        & \checkmark &            &            & 0.915 & 0.904 & 0.873 & - \\
        & \checkmark & \checkmark &            & 0.920 & 0.915 & 0.878 & - \\
        & \checkmark &            & \checkmark & 0.921 & 0.910 & 0.878 & - \\
        & \checkmark & \checkmark & \checkmark & 0.923 & 0.921 & \textbf{0.882} & - \\
        \midrule
        \multirow{4}{*}{$q+d$} 
        & \checkmark &            &            & 0.913 & 0.906 & - & 0.837 \\
        & \checkmark & \checkmark &            & 0.924 & 0.915 & - & 0.840 \\
        & \checkmark &            & \checkmark & 0.925 & 0.913 & - & 0.838 \\
        & \checkmark & \checkmark & \checkmark & 0.933 & 0.921 & - & 0.838 \\
        \midrule
        \multirow{4}{*}{$q+s+d$} 
        & \checkmark &            &            & 0.916 & 0.906 & 0.870 & \textbf{0.851}  \\
        & \checkmark & \checkmark &            & 0.925 & 0.921 & 0.864 & 0.830 \\
        & \checkmark &            & \checkmark & 0.926 & 0.921 & 0.873 & 0.838 \\
        & \checkmark & \checkmark & \checkmark & \textbf{0.939} & \textbf{0.929} & 0.879 & 0.847 \\
        \bottomrule
    \end{tabular}}
\end{table}

\subsection{Comparison on Distortion Types}\label{sec:dist}
In this section, we present the SRCC performance of the proposed DEFNet compared to several state-of-the-art methods across diverse distortion types in the CSIQ datasets. The distortion types CSIQ include white noise (WN), Gaussian blur (GB), pink Gaussian noise (PN), contrast decrements (CD), JPEG compression (JPEG) and JPEG2000 compression (JP2K).
As shown in Table \ref{tab:distortion_type_csiq}, DEFNet outperforms other methods across nearly all types of distortion. This highlights the robustness and adaptability of DEFNet to various types of synthetic image distortion. 
In addition, this validates the superiority of the multilevel information fusion strategy, which allows to give accurate quality predictions regardless of distortion types.
More comparison results on distortion types in other datasets are shown in Supplementary \ref{sup:dist}.

\begin{table*}[ht]
    \centering
    \fontsize{9}{10}\selectfont %{字体尺寸}{行距}
    \caption{SRCC and PLCC across the six IQA datasets under different weighting parameters. Best scores are highlighted in bold.}
    \label{tab:lambda}
    % \resizebox{\linewidth}{!}{
    \begin{tabular}{c|c|cc|cc|cc|cc|cc|cc}
        \toprule
        \multicolumn{2}{c|}{Parameter} & \multicolumn{6}{c|}{Synthetic Distortion} & \multicolumn{6}{c}{Authentic Distortion} \\
        \cmidrule(lr){1-2} \cmidrule(lr){3-8} \cmidrule(lr){9-14} 
        \multirow{2}{*}{$\lambda_1$} & \multirow{2}{*}{$\lambda_2$} & \multicolumn{2}{c|}{LIVE} & \multicolumn{2}{c|}{CSIQ} & \multicolumn{2}{c|}{KADID-10k} & \multicolumn{2}{c|}{BID} & \multicolumn{2}{c|}{LIVE-C} & \multicolumn{2}{c}{KonIQ-10k}  \\
        & & SRCC  & PLCC  & SRCC  & PLCC  & SRCC  & PLCC & SRCC  & PLCC  & SRCC  & PLCC  & SRCC  & PLCC   \\
        \midrule
        0.1 & 0.1 & 0.978 & \textbf{0.960} & 0.967 & 0.964 & 0.942 & 0.944 & 0.910 & 0.909 & 0.918 & \textbf{0.897} & 0.920 & \textbf{0.901} \\
        0.1 & 0.2 & 0.979  & 0.959  & 0.971  & 0.964   & \textbf{0.947}  & \textbf{0.947}  & 0.906  & 0.911  & \textbf{0.921}  & 0.890  & \textbf{0.921}  & \textbf{0.901}   \\
        0.1 & 0.3 & 0.976  & 0.952  & 0.970  & 0.961   & 0.945  & 0.943  & 0.903  & 0.901  & 0.907  & 0.867  & 0.920  & 0.894   \\
        \midrule
        0.2 & 0.1 & \textbf{0.981}  & 0.952  & \textbf{0.973}  & \textbf{0.965}  & 0.945  & \textbf{0.947}  & \textbf{0.912}  & \textbf{0.921} & 0.910  & 0.869  & 0.919  & 0.898    \\
        0.3 & 0.1 & 0.978  & 0.946  & 0.967  & 0.949  & 0.946  & 0.943  & 0.911  & 0.900 & 0.904  & 0.836 & 0.918  & 0.893  \\
        0.4 & 0.1 & 0.979  & 0.945  & 0.966  & 0.949  & 0.944  & 0.940  & 0.903  & 0.897  & 0.903  & 0.821  & 0.917  & 0.890 \\
        \bottomrule
    \end{tabular}
\end{table*}

\subsection{Ablation Study}\label{sec:ablation}
In this section, we conduct ablation study to validate the contribution of each task assistance and each loss components to the overall performance. A total of four different task combinations are explored, specifically, ablation experiments are conducted on the presence or absence of scene classification and distortion type categorization. Within each task combination, ablation of each component in the overall loss is also performed. The results for the multitask learning are presented in Supplementary \ref{sup:ablation}. All the results are averaged across the six datasets and listed in Table~\ref{tab:ablation}.

A couple of observations can be drawn.
First, utilizing auxiliary tasks can significantly improve the performance of BIQA. 
With both two auxiliary tasks aided, DEFNet archieves the best performance across all 16 settings in the ablation study.
Second, the proposed information fusion strategy, either across sub-regions or between local and global image context, contribute positively to BIQA. 
The inclusion of either cross-region loss or cross-grained loss leads to noticeable improvements in model performance. These loss components enable the model to better capture complementary features. With both loss components aided, DEFNet achieves highest performance in most cases.
Third, the extent to which evidential fusion positively impacts performance surpasses that offered by the auxiliary tasks alone. This underscores the contribution of the proposed DEFNet, in which the multilevel trustworthy evidential fusion leads to a more accurate quality assessment.

\subsection{Hyperparameter Analysis}

In order to discuss the effect of the weighting parameters in Eq. (\ref{eq:overall}), we adjust different combinations of $\lambda_1,\lambda_2$ and list the experimental results in the IQA six datasets in Table~\ref{tab:lambda}. This gives an illustration of the trade-off between the contributions of the cross-region loss and the cross-grained loss to the overall model performance.
As the weighting parameters increase from small values, the performance of both BIQA and the auxiliary tasks improves initially, reaching optimal values at moderate weight settings (e.g., $\lambda_1=0.2$ and $\lambda_2=0.2$). This improvement can be attributed to the gradual integration of evidential learning, which enhances the model's ability to extract and integrate complementary information from the auxiliary tasks. However, when the weights become excessively large (e.g., $\lambda_1=0.4$), the performance begins to degrade. This decline is likely due to the model overemphasizing the evidential loss components, which detracts from the focus on the primary BIQA task.

\begin{table}[t]
  \centering
  \fontsize{9}{10}\selectfont %{字体尺寸}{行距}
  \caption{Mean confidence interval widths.}
  \label{tab:ci}
  \begin{tabular}{l|cc}
      \toprule
      Method & LIQE \cite{Zhang_2023_CVPR}  & DEFNet \\
      \midrule
      CI width ($\downarrow$) & 0.286   &  0.251   \\
      \bottomrule  
  \end{tabular}
  \vspace{-0.3cm}
\end{table}

\subsection{Uncertainty Analysis}\label{sec:uncertainty}

By applying evidence theory to BIQA task and utilizing normal-inverse gamma distribution mixture, we reduce the epistemic uncertainty of the model. Specifically, as shown in Figure \ref{fig:intro} and Supplementary \ref{sup:uncertainty}, DEFNet exhibits better performance and lower uncertainty compared to LIQE (a representative of methods that utilize auxiliary tasks to aid BIQA). 
In addition, the mean confidence interval (CI) widths in the scatter plot are shown in Table \ref{tab:ci}, which quantitatively illustrates that the advanced uncertainty estimation technique is beneficial for decreasing uncertainty.

%% file: sec/6_conclusion.tex
\section{Conclusion}\label{sec:conclusion}
This paper introduced evidential fusion into BIQA with the assistance of auxiliary tasks for in-depth information fusion. To this end, we proposed a trustworthy information fusion strategy at two levels. The cross sub-region fusion effectively captures diverse features and patterns from different regions, while the local-global fusion balances fine-grained local details with a broader global view, providing a comprehensive representation of image quality. Our proposed method serves as a practical solution for BIQA uncertainty estimation and in-depth information fusion.
\textbf{Limitations.} There is still room to improve robustness and generalization for highly diverse and novel distortions and never-before-learned scenarios. In addition, the number of parameters of the model is relatively high and can be further optimized.

%% file: sec/7_supplementary.tex
% \newpage
\clearpage
\appendix

\section{Related Work}\label{sup:related}

\subsection{Blind Image Quality Assessment}
In the field of blind image quality assessment, state-of-the-art methods have evolved from manual feature extraction to deep learning based approaches. Zhang \textit{et al.} \cite{8576582} achieved high quality assessment performance by exploring deep bilinear convolutional neural network for synthetic and authentic distorted datasets. Ke \textit{et al.} \cite{Ke_2021_ICCV} presented a multi-scale image quality Transformer to process native resolution images with different sizes and aspect ratios. Saha \textit{et al.} \cite{Saha_2023_CVPR} proposed an expert hybrid method for automatic perceptual image quality assessment.
In addition to improvements at the model level, some research work combined extracted features from different modules \cite{9718583,lou2023refining}, some work explored how to perform image quality assessment in a self-supervised manner \cite{9796010}, and some trained a unified model jointly on multiple datasets \cite{Zhang_2023_CVPR,9369977}.

\subsection{Evidential Learning}
With the rapid development of deep learning technology, although it has achieved good enough results, its reliability and safety have always been concerned \cite{he2025co,he2024epl}. How to make safe and effective decisions is still a very important research focus \cite{WU2024102014,MENG2023344,he2021conflicting,he2022mmget}. To provide a feasible solution to the problem, some researchers choose to transfer the evidence theory into the field of deep learning \cite{he2024generalized,he2024mutual}. The evidence theory is also called Dempster–Shafer theory, which is firstly proposed by Dempster \cite{Dempster2008} to serve as a general framework for reasoning with uncertainty and further developed by Shafer \cite{SHAFER20167}. The evidnece theory satisfies weaker conditions than Bayesian probability theory and possesses the ability to model uncertainty directly. Recently, a representative work proposed by Amini introduces the concept of uncertainty measures for neural networks for generating more precise predictions \cite{NEURIPS2020_aab08546}. The newly proposed method has two main advantages. First, it can be typically integrated into neural network architectures directly so that the network is able to learn to output parameters of a probability distribution instead of just a single point estimate. Second, when the model makes a prediction, the corresponding uncertainty assessment can be obtained at the same time, and the prediction results can be further modified to achieve better performance.

\begin{figure}[ht]
    \centering
    \begin{minipage}[b]{0.32\linewidth}
        \centering
        \centerline{\includegraphics[width=\linewidth]{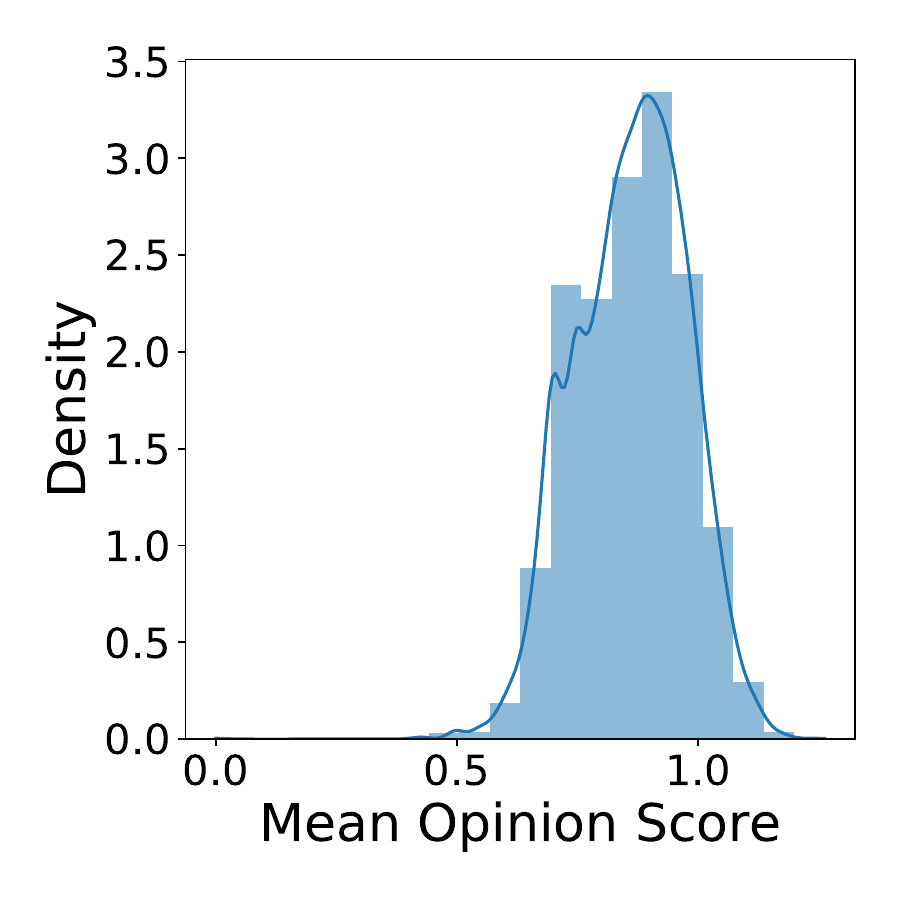}}
        \centerline{(a) KADID-10k}\medskip
    \end{minipage}
    \begin{minipage}[b]{0.32\linewidth}
        \centering
        \centerline{\includegraphics[width=\linewidth]{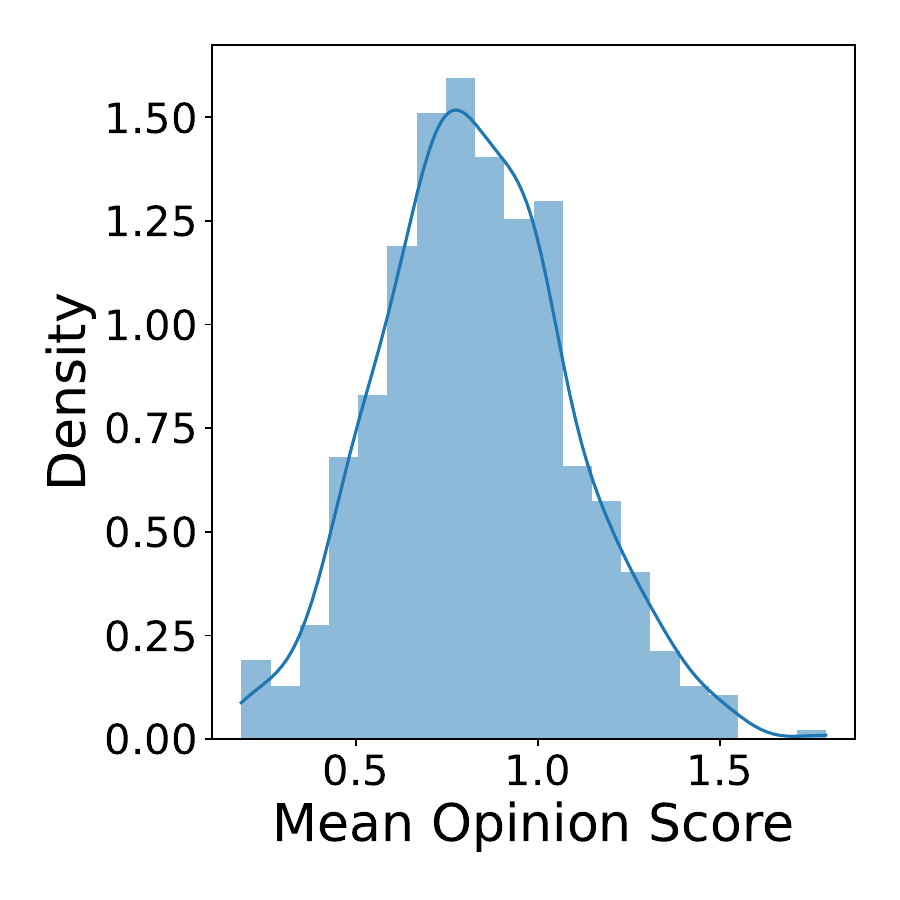}}
        \centerline{(b) BID}\medskip
    \end{minipage}
    \begin{minipage}[b]{0.32\linewidth}
        \centering
        \centerline{\includegraphics[width=\linewidth]{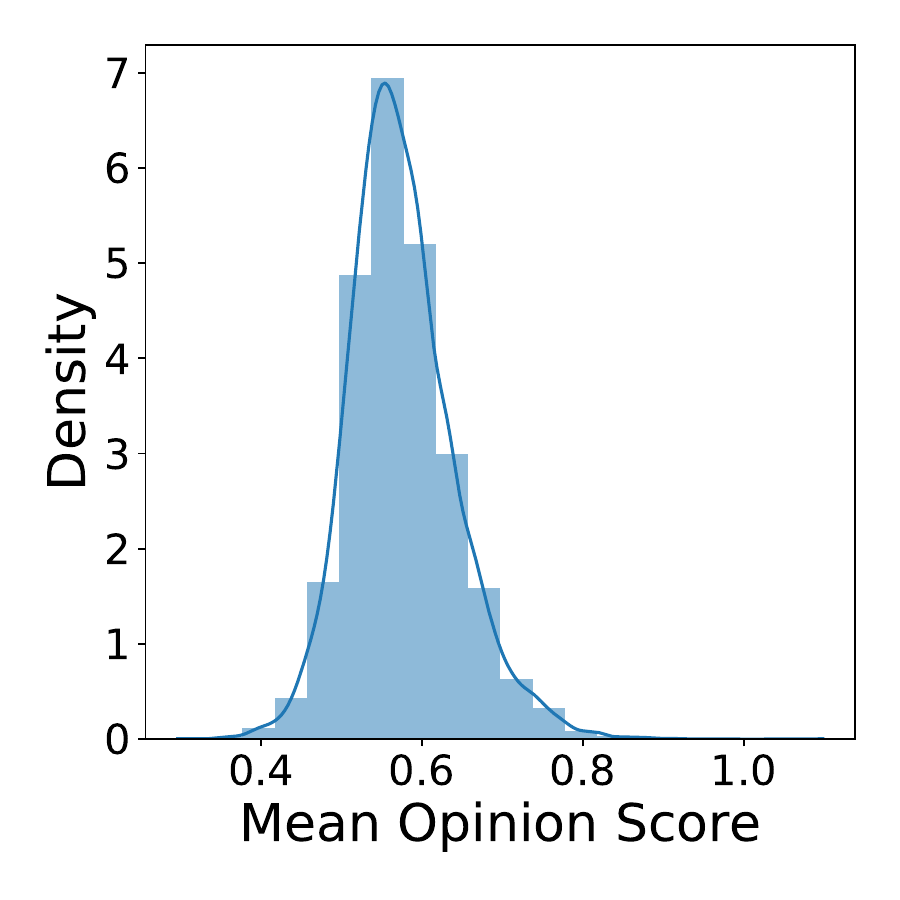}}
        \centerline{(c) KonIQ-10k}\medskip
    \end{minipage}
    \caption{Kernel density estimation plots.}
    \label{fig:kdeplot}
\end{figure}

\begin{figure}[ht]
    \centering
    \begin{minipage}[b]{0.32\linewidth}
        \centering
        \centerline{\includegraphics[width=\linewidth]{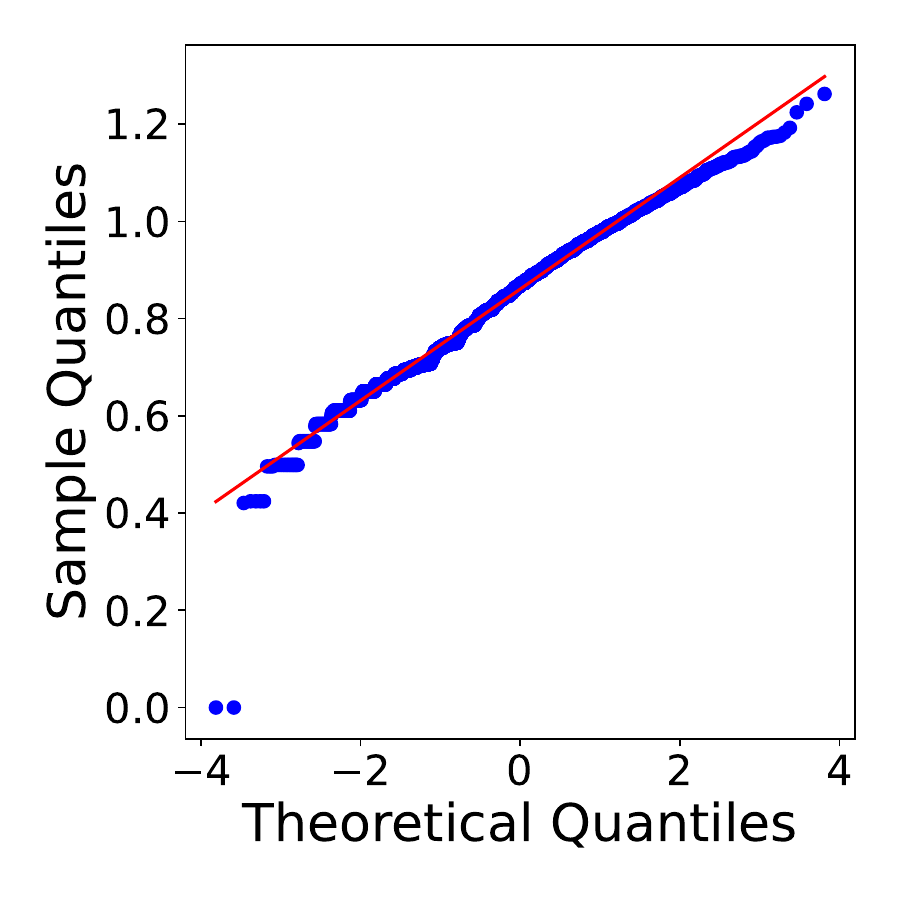}}
        \centerline{(a) KADID-10k}\medskip
    \end{minipage}
    \begin{minipage}[b]{0.32\linewidth}
        \centering
        \centerline{\includegraphics[width=\linewidth]{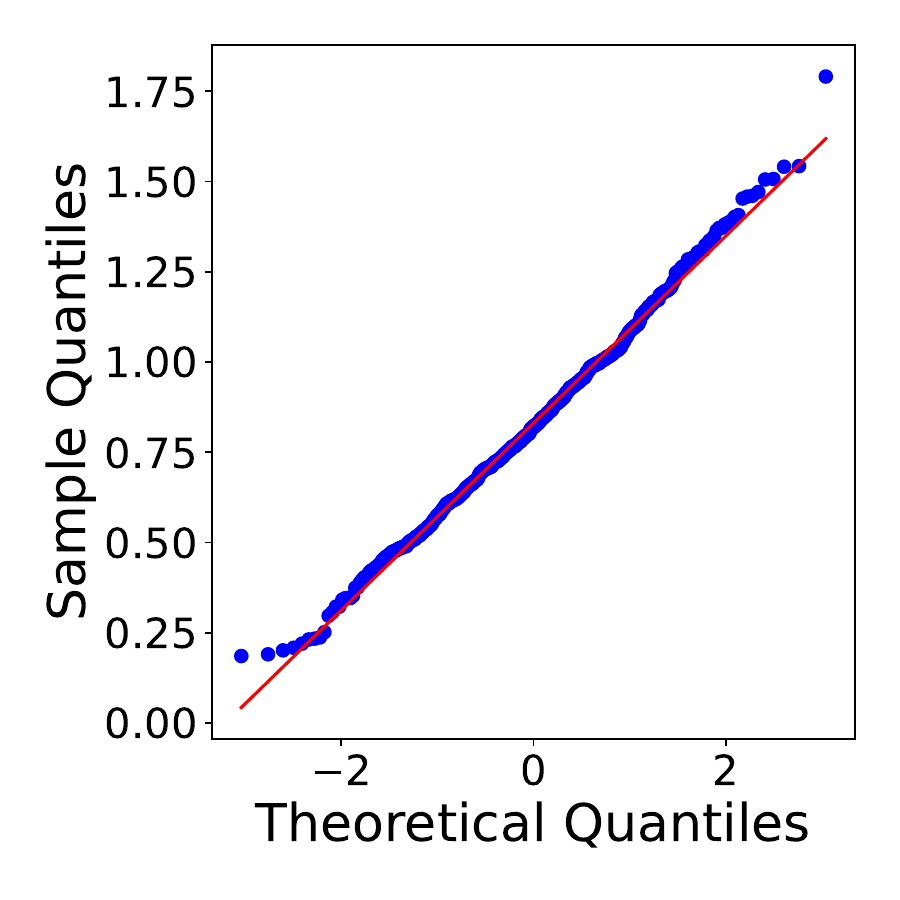}}
        \centerline{(b) BID}\medskip
    \end{minipage}
    \begin{minipage}[b]{0.32\linewidth}
        \centering
        \centerline{\includegraphics[width=\linewidth]{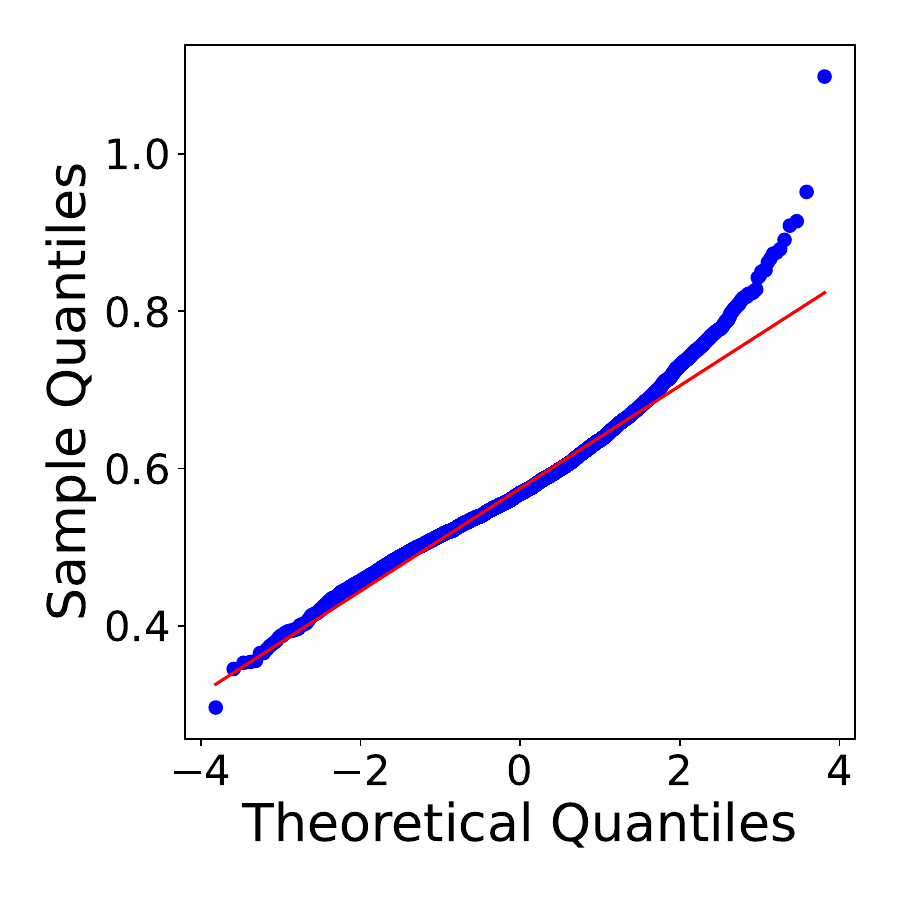}}
        \centerline{(c) KonIQ-10k}\medskip
    \end{minipage}
    \caption{Quantile–quantile plots.}
    \label{fig:qqplot}
\end{figure}

\section{Assumption Justification}\label{sup:norm}
To illustrate the quality score is subject to a normal distribution, theoretical justification and empirical evidence are given as follows.

\textbf{Theoretical Justification.}
The quality score label (typically mean opinion score) always necessitates averaging the scores of multiple human evaluators. According to the central limit theorem, the sum (or average) of these random variables, which can be interpreted as the quality scores provided by various evaluators, tends to approximate a normal distribution. This holds true even when the individual variables do not follow a normal distribution. Thus, theoretically, as the number of evaluators increases and the sample size becomes large enough (it holds true for datasets LIVE \cite{1709988}, CSIQ \cite{10.1117/1.3267105}, KADID-10k \cite{8743252}, BID \cite{5492198}, LIVE-C \cite{7327186} and KonIQ-10k \cite{8968750}), these labels will align with a normal distribution.

\textbf{Empirical Evidence.}
To empirically validate the assumption, we conducted several statistical analysis.
First, we plot kernel density and distribution of the quality scores across datasets (as shown in Figure \ref{fig:kdeplot}) and observe that the histogram forms a bell-shaped curve, indicating it is subject to normal distribution. 
Second, we generate Quantile–quantile plots (as shown in Figure \ref{fig:qqplot}) comparing the quantiles of the quality score distributions against the quantiles of a standard normal distribution. The alignment of the points along the reference line further confirms the normal distribution.

\begin{algorithm*}[t]                       
    \textbf{Input:} Distorted image $\boldsymbol{x} \in \mathbb{R}^{H\times W\times 3}$\\
    \textbf{Output:} Predicted quality scores $\hat{q}(\boldsymbol{x}) \in \mathbb{R}$
    \caption{Deep Evidential Fusion Network}
    \label{alg:framework}
    \begin{algorithmic}[1]
        \STATE \# \textbf{Stage I: Local and Global Probability Scores} (Section \ref{sec:prob})
        \STATE Crop to obtain local sub-images $\{\boldsymbol{x}^{\text{local}}_i\}_{i=1}^N$, downsample $\boldsymbol{x}$ to generate the global image $\boldsymbol{x}^{\text{global}}$
        \STATE Use CLIP to obtain $\operatorname{logit}(c,s,d|\boldsymbol{x}')$, and then derive the joint probability with Eq. (\ref{eq:prob}):
        $$\hat{p}(c,s,d)(\boldsymbol{x}') = \dfrac{\operatorname{exp}(\operatorname{logit}(c,s,d|\boldsymbol{x}') / \kappa)}{\sum_{c,s,d}\operatorname{exp}(\operatorname{logit}(c,s,d|\boldsymbol{x}') / \kappa)},$$
        where $\boldsymbol{x}'$ denotes for either the local sub-image $\boldsymbol{x}^{\text{local}}_i$ or the global image $\boldsymbol{x}^{\text{global}}$
        \\
        \STATE \# \textbf{Stage II: Multitask Optimization} (Section \ref{sec:multitask})
        \STATE Compute the BIQA loss $\ell_q$ for each image pair, as defined in Eq. (\ref{eq:ellq}):
        $$\ell_q(\boldsymbol{x}_1, \boldsymbol{x}_2; \theta) = 1 - \sqrt{p(\boldsymbol{x}_1, \boldsymbol{x}_2) \hat{p}(\boldsymbol{x}_1, \boldsymbol{x}_2)} - \sqrt{(1 - p(\boldsymbol{x}_1, \boldsymbol{x}_2))(1-\hat{p}(\boldsymbol{x}_1, \boldsymbol{x}_2))}$$
        \STATE Compute the scene classification loss $\ell_s$ for each image, as defined in Eq. (\ref{eq:scene}):
        $$\ell_s(\boldsymbol{x}; \theta) = \frac{1}{|\mathcal{S}|} \sum_{s \in \mathcal{S}} \Big( 1 - \sqrt{p(s|\boldsymbol{x}) \hat{p}(s|\boldsymbol{x})} - \sqrt{(1-p(s|\boldsymbol{x})) (1-\hat{p}(s|\boldsymbol{x}))} \Big)$$
        \STATE Compute the distortion type classification loss $\ell_d$ for each image, as defined in Eq. (\ref{eq:dist}):
        $$\ell_d(\boldsymbol{x}; \theta) = 1 - \sum_{d\in\mathcal{D}} \sqrt{p(d|\boldsymbol{x}) \hat{p}(d|\boldsymbol{x})}$$
        \STATE Combine auxiliary tasks to multitask loss, which is specifically defined in Eq. (\ref{eq:lm}):
        $$\mathcal{L}^M (\theta) = \dfrac{1}{|\mathcal{P}|} \sum_{(\boldsymbol{x}_1,\boldsymbol{x}_2) \in \mathcal{P}} \lambda_q \ell_q(\boldsymbol{x}_1,\boldsymbol{x}_2; \theta) + \dfrac{1}{|\mathcal{B}|} \sum_{\boldsymbol{x} \in \mathcal{B}} \big[ \lambda_s \ell_s(\boldsymbol{x}) + \lambda_d \ell_d(\boldsymbol{x}) \big],$$
        \STATE
        \STATE \# \textbf{Stage III: Cross Sub-regions Information Fusion} (Section \ref{sec:cross})
        \STATE For BIQA task ($t=q$), derive the evidential loss on local outputs, as defined in Eq. (\ref{eq:lut}),
        $$\mathcal{L}^U_q(\theta) = \dfrac{1}{|\mathcal{B}|} \sum_{\boldsymbol{x}\in\mathcal{B}}  \ell^U\left(\operatorname{softplus}( \boldsymbol{x}^{\operatorname{cross}}_q), \boldsymbol{y}_q, \theta\right),$$
        where $\boldsymbol{x}^{\operatorname{cross}}_q$ is obtained with Eq. (\ref{eq:nigadd}), and $\ell^U$ is defined in Eq. (\ref{eq:lu})
        \STATE For scene classification task ($t=s$), derive the evidential loss, as defined in Eq. (\ref{eq:lut}):
        $$\mathcal{L}^U_s(\theta) = \dfrac{1}{|\mathcal{B}|} \sum_{\boldsymbol{x}\in\mathcal{B}}  \ell^U\left(\operatorname{softplus}( \boldsymbol{x}^{\operatorname{cross}}_s), \boldsymbol{y}_s, \theta\right)$$
        \STATE For distortion type classification task ($t=d$), derive the evidential loss, as defined in Eq. (\ref{eq:lut})
        $$\mathcal{L}^U_d(\theta) = \dfrac{1}{|\mathcal{B}|} \sum_{\boldsymbol{x}\in\mathcal{B}}  \ell^U\left(\operatorname{softplus}( \boldsymbol{x}^{\operatorname{cross}}_d), \boldsymbol{y}_d, \theta\right)$$
        \STATE Integrate sub-region information to the cross-region loss, as defined in Eq. (\ref{eq:LU}):
        $$\mathcal{L}^U(\theta) = \mathcal{L}^U_q(\theta)+\mathcal{L}^U_s(\theta)+\mathcal{L}^U_d(\theta)$$
        \algstore{myalg}
    \end{algorithmic}
\end{algorithm*}

\begin{algorithm*}[t]                  
    \begin{algorithmic} [1]
        \algrestore{myalg}
        \STATE \# \textbf{Stage IV: Local-global Information Fusion} (Section \ref{sec:lgif})
        \STATE Derive the evidential loss on local-global outputs for BIQA task ($t=q$), as defined in Eq. (\ref{eq:lft}):
        $$\mathcal{L}^F_q(\theta) =\sum \ell^U\left(\operatorname{softplus}( \boldsymbol{x}^{\operatorname{fusion}}_q), \boldsymbol{y}_q, \theta\right),$$
        where $\boldsymbol{x}^{\operatorname{fusion}}_q$ is obtained with Eq. (\ref{eq:xfusiont}), and $\ell^U$ is defined in Eq. (\ref{eq:lu})
        \STATE Derive the evidential loss for scene classification task ($t=s$), as defined in Eq. (\ref{eq:lft}):
        $$\mathcal{L}^F_s(\theta) =\sum \ell^U\left(\operatorname{softplus}( \boldsymbol{x}^{\operatorname{fusion}}_s), \boldsymbol{y}_s, \theta\right)$$
        \STATE Derive the evidential loss $\mathcal{L}^F_d(\theta)$ for distortion type classification task ($t=d$), as defined in Eq. (\ref{eq:lft}):
        $$\mathcal{L}^F_d(\theta) =\sum \ell^U\left(\operatorname{softplus}( \boldsymbol{x}^{\operatorname{fusion}}_d), \boldsymbol{y}_d, \theta\right)$$
        \STATE Integrate local-global information to the cross-grained loss, as defined in Eq. (\ref{eq:LF}):
        $$\mathcal{L}^F(\theta)=\mathcal{L}^F_q(\theta) + \mathcal{L}^F_s(\theta) + \mathcal{L}^F_d(\theta)$$
        \STATE 
        \STATE \# \textbf{Stage V: Overall Loss} (Section \ref{sec:overloss})
        \STATE Calculate the overall loss, as defined in Eq. (\ref{eq:overall}):
        $$\mathcal{L}(\theta) = \mathcal{L}^M (\theta) + \lambda_1 \mathcal{L}^U (\theta) + \lambda_2 \mathcal{L}^F (\theta)$$
        \STATE Update parameters in DEFNet through gradient feedback based on $\mathcal{L}(\theta)$
        \STATE 
        \STATE \# Give predict quality score, as defined in Eq. (\ref{eq:qpred})
        \STATE \textbf{Return} $\hat{q}(\boldsymbol{x}) = \sum_{c=1}^C \hat{p}(c|\boldsymbol{x}) \times c$
    \end{algorithmic}
\end{algorithm*}

\section{Algorithm Overview}\label{sup:alg}

To give a detailed presentation of variable flow  of the method presented in Figure \ref{fig:framework} and Section \ref{sec:method}, we present Algorithm \ref{alg:framework} in association with specific formulas.

\section{More about the Textual Description}\label{sup:imp}
To obtain the logits and probability scores for local and global images, CLIP is utilized. The language encoder process the textual template \textit{``a photo of a(n) \{s\} with \{d\} artifacts, which is of \{c\} quality.``}, where $s$ denotes scene type, $d$ denotes distortion type, $c$ denotes quality levels. These three variables take arbitrary values in their respective sets. Specifically, the scene type $s\in\mathcal{S}=$ \{``animal", ``cityscape", ``human", ``indoor scene", ``landscape", ``night scene", ``plant", ``still-life", and ``others"\}, where there are altogether 9 choices. The distortion type $d\in\mathcal{D}=$ \{``blur", ``color-related", ``contrast", ``JPEG compression", ``JPEG2000 compression", ``noise", ``over-exposure", ``quantization", ``under-exposure", ``spatially-localized", and ``others"\}, where there are altogether 11 choices. It is worth noting that the image belongs to the last category ``others" if no distortion artifact is performed.
The quality level $c\in\mathcal{C}=\{1,2,3,4,5\}$, which corresponds to levels of ``bad", ``poor", ``fair", ``good" and ``perfect", respectively.

\section{More Experimental Results}\label{sup:exp}
In this section, we present some additional experimental results as supplementary to Section \ref{sec:exp}:
\begin{itemize}
    \item Supplementary \ref{sup:gmad} presents the results of the gMAD competition in SPAQ, supplementing Section \ref{sec:gmad}.
    \item Supplementary \ref{sup:dist} presents the comparison on distortion types in LIVE, supplementing Section \ref{sec:dist}.
    \item Supplementary \ref{sup:ablation} presents complete results in ablation study, supplementing Section \ref{sec:ablation}.
    \item Supplementary \ref{sup:uncertainty} presents clear scatter plots and confidence intervals, supplementing Section \ref{sec:uncertainty}.
    \item Supplementary \ref{sup:complexity} presents additional model complexity analysis.
\end{itemize}

\subsection{gMAD Competition in SPAQ}\label{sup:gmad}
\begin{figure*}[t] 
    \centering
    \includegraphics[width=0.95\linewidth]{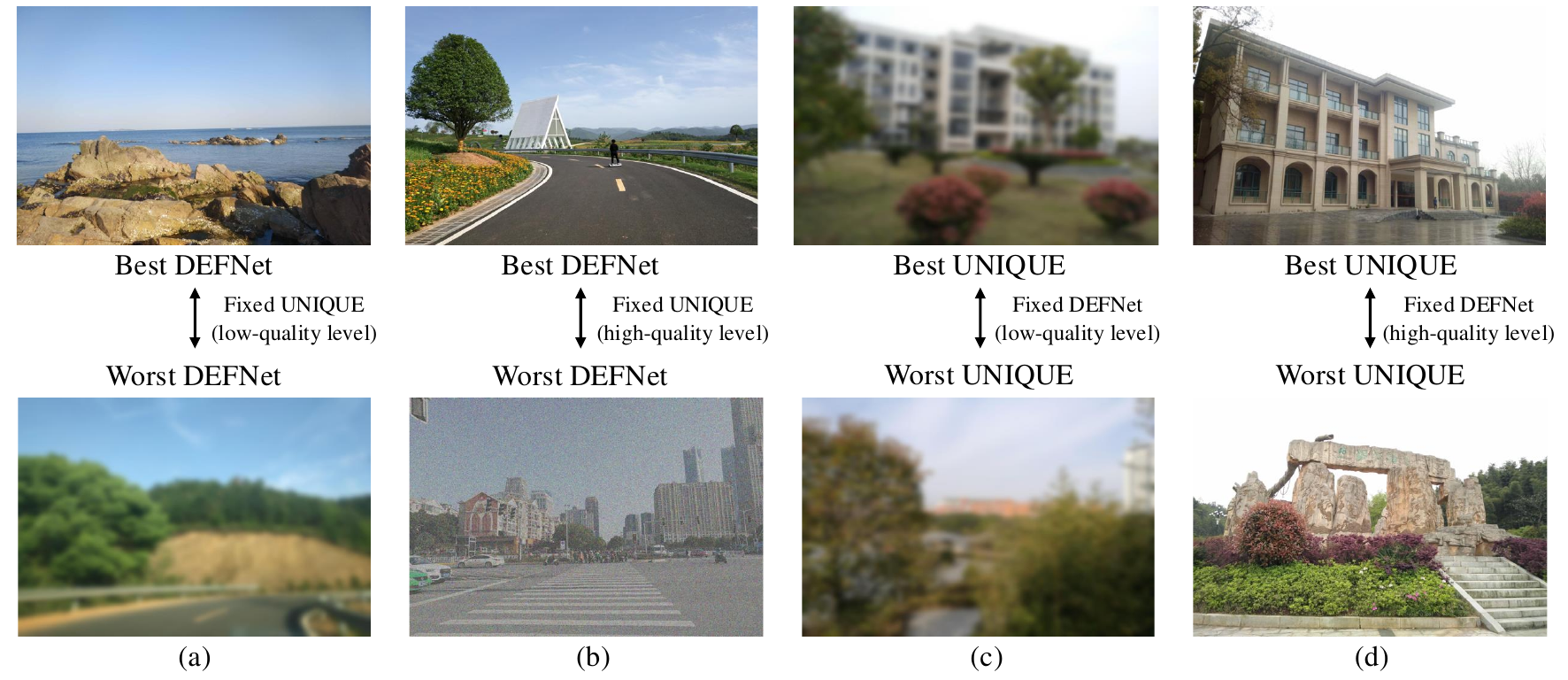} 
    \caption{gMAD competition between UNIQUE and DEFNet in SPAQ. (a) Fixed UNIQUE at low-quality level. (b) Fixed UNIQUE at high-quality level. (c) Fixed DEFNet at low-quality level. (d) Fixed DEFNet at high-quality level.}
    \label{fig:gmad_spaq} 
\end{figure*}

The results of the gMAD competition in SPAQ are shown in Figure \ref{fig:gmad_spaq}. In the experimental settings, SPAQ is a representative of authentic distortion datasets. Similar conclusions can be drawn from the SPAQ's gMAD comparisons as those in the WED in Section \ref{sec:gmad}.

\subsection{Comparison on Distortion Types in LIVE}\label{sup:dist} 

The performance of DEFNet, as well as several state-of-the-art methods, across diverse distortion types in the LIVE dataset is shown in Table~\ref{tab:distortion_type_live}.
The distortion types in LIVE include white noise (WN), fast fading (FF), Gaussian blur (GB), JPEG compression (JPEG) and JPEG2000 compression (JP2K).
In the LIVE dataset, DEFNet exhibits the highest SRCC values across a majority of distortion types.

\begin{table}[t]  % session 8
    \centering
    \fontsize{9}{10}\selectfont %{字体尺寸}{行距}
    \caption{SRCC performance on diverse distortion types of LIVE.} %  WN, FF, GB, JPEG, JP2K stand for white noise, fast fading, Gaussian blur, JPEG and JPEG2000 compression.}
    \resizebox{\linewidth}{!}{
    \begin{tabular}{c|cccccc}
        \toprule
        Distortion & WN & FF & GB & JPEG & JP2K \\
        \midrule
        BRISQUE \cite{mittal2012no}       & 0.977 & 0.877 & 0.951 & 0.965 & 0.914 \\
        ILNIQE \cite{7094273}             & 0.975 & 0.827 & 0.911 & 0.931 & 0.902 \\
        deepIQA \cite{bosse2016deep}      & 0.979 & 0.897 & 0.970 & 0.953 & 0.968 \\
        PQR \cite{zeng2017probabilistic}  & 0.981 & 0.921 & 0.944 & 0.965 & 0.953 \\
        DBCNN \cite{8576582}              & 0.980 & 0.930 & 0.935 & 0.972 & 0.955 \\
        HyperIQA \cite{Su_2020_CVPR}      & 0.982 & 0.934 & 0.926 & 0.961 & 0.949 \\
        OLNet \cite{9718583}              & 0.984 & 0.930 & 0.931 & 0.976 & 0.970 \\
        VCRNet \cite{pan2022vcrnet}       & \underline{0.988} & \textbf{0.962} & \textbf{0.978} & \underline{0.979} & \underline{0.975}   \\
        \midrule
        \textbf{DEFNet}                   & \textbf{0.989} & \underline{0.942} & \underline{0.971} & \textbf{0.985} & \textbf{0.983} \\
        \bottomrule
    \end{tabular}}
    \label{tab:distortion_type_live}
\end{table}

\subsection{Ablation Study Results}\label{sup:ablation}

\begin{figure*}[!ht]
    \centering
    \begin{minipage}[b]{0.245\linewidth}
        \centering
        \centerline{\includegraphics[width=\linewidth]{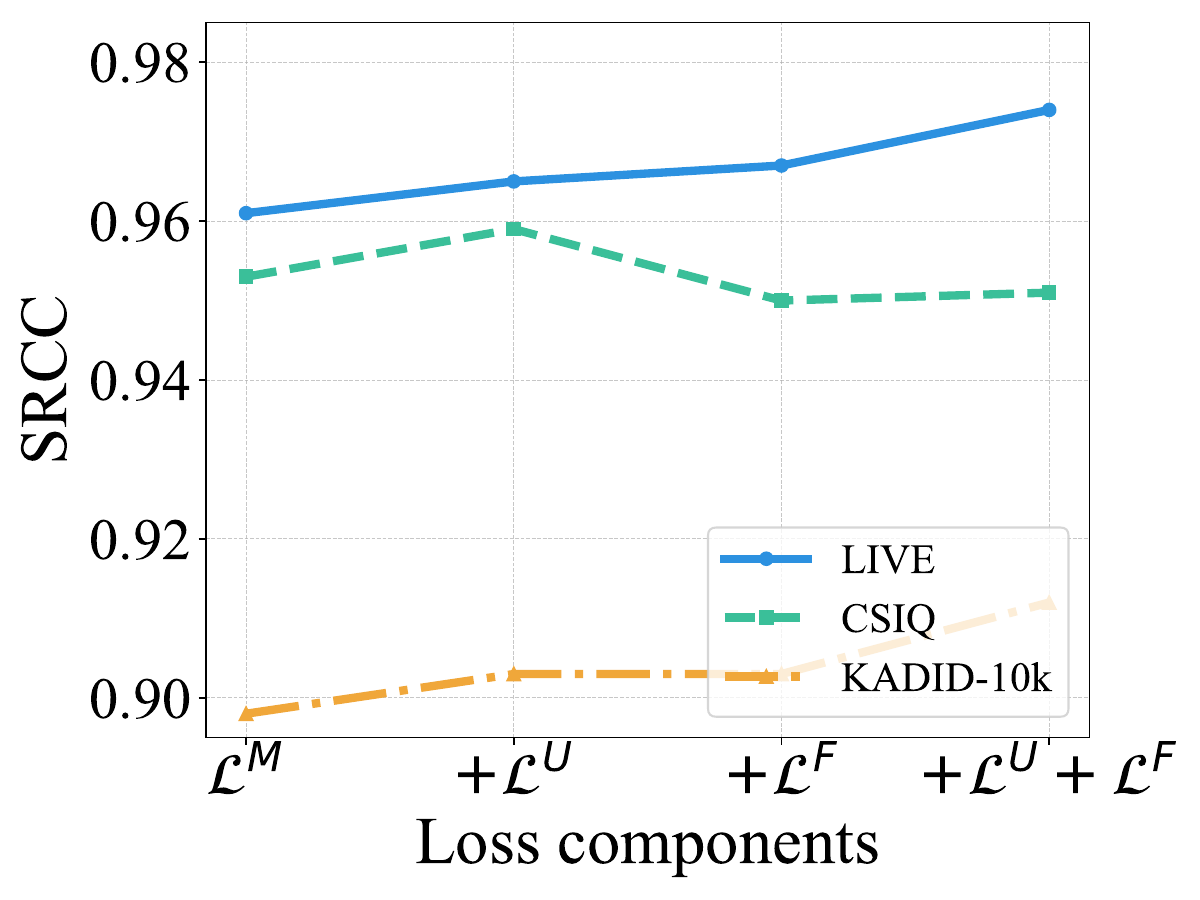}}
        \centerline{(a) IQA task}\medskip
    \end{minipage}
    \begin{minipage}[b]{0.245\linewidth}
        \centering
        \centerline{\includegraphics[width=\linewidth]{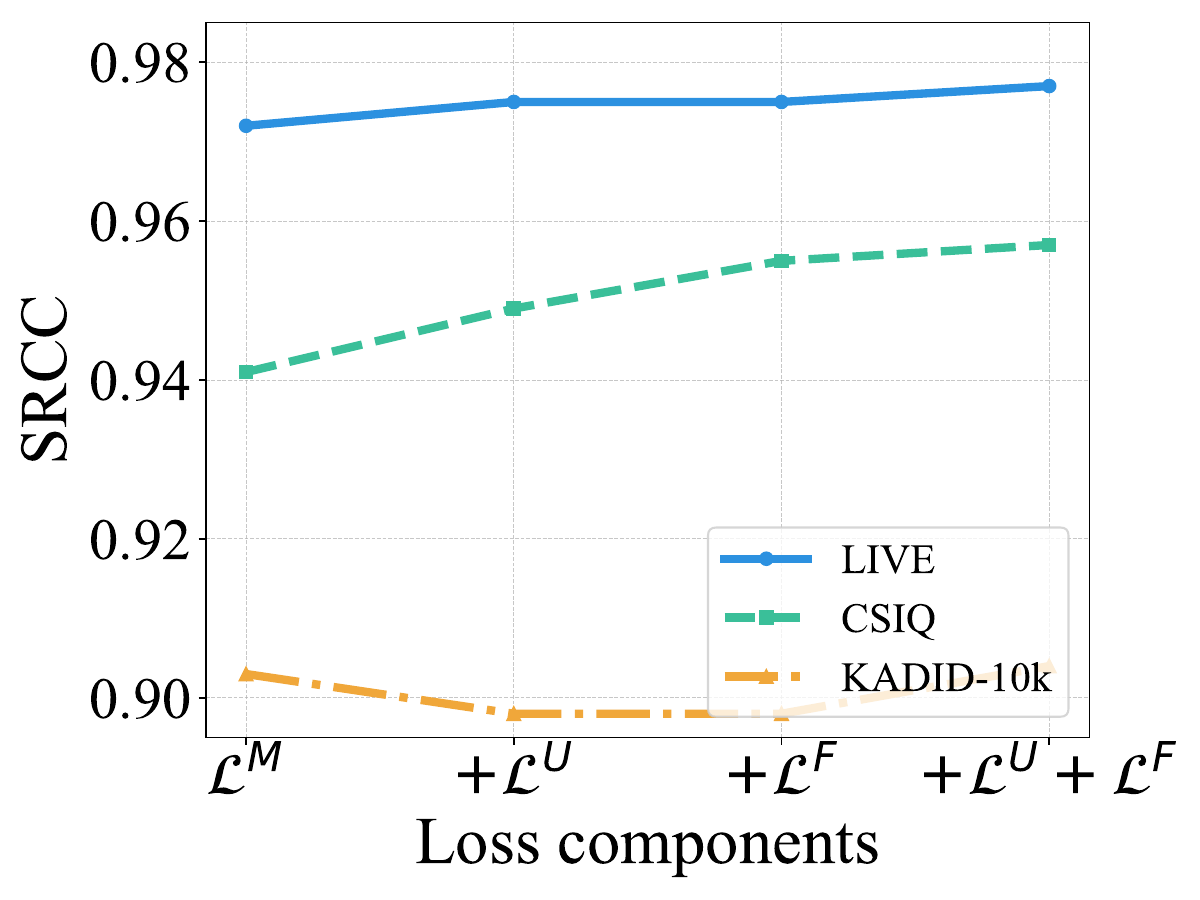}}
        \centerline{(b) Scene aided}\medskip
    \end{minipage}
    \begin{minipage}[b]{0.245\linewidth}
        \centering
        \centerline{\includegraphics[width=\linewidth]{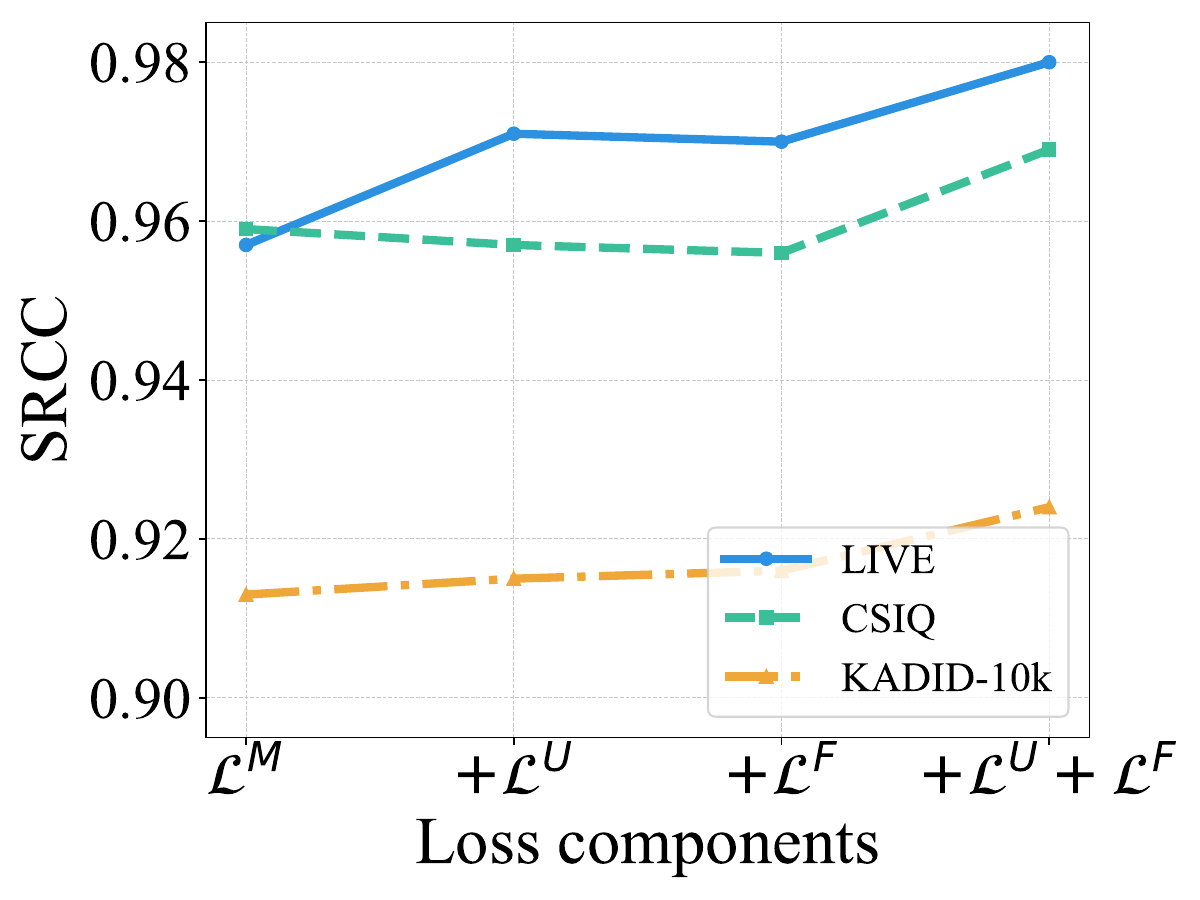}}
        \centerline{(c) Distortion aided}\medskip
    \end{minipage}
    \begin{minipage}[b]{0.245\linewidth}
        \centering
        \centerline{\includegraphics[width=\linewidth]{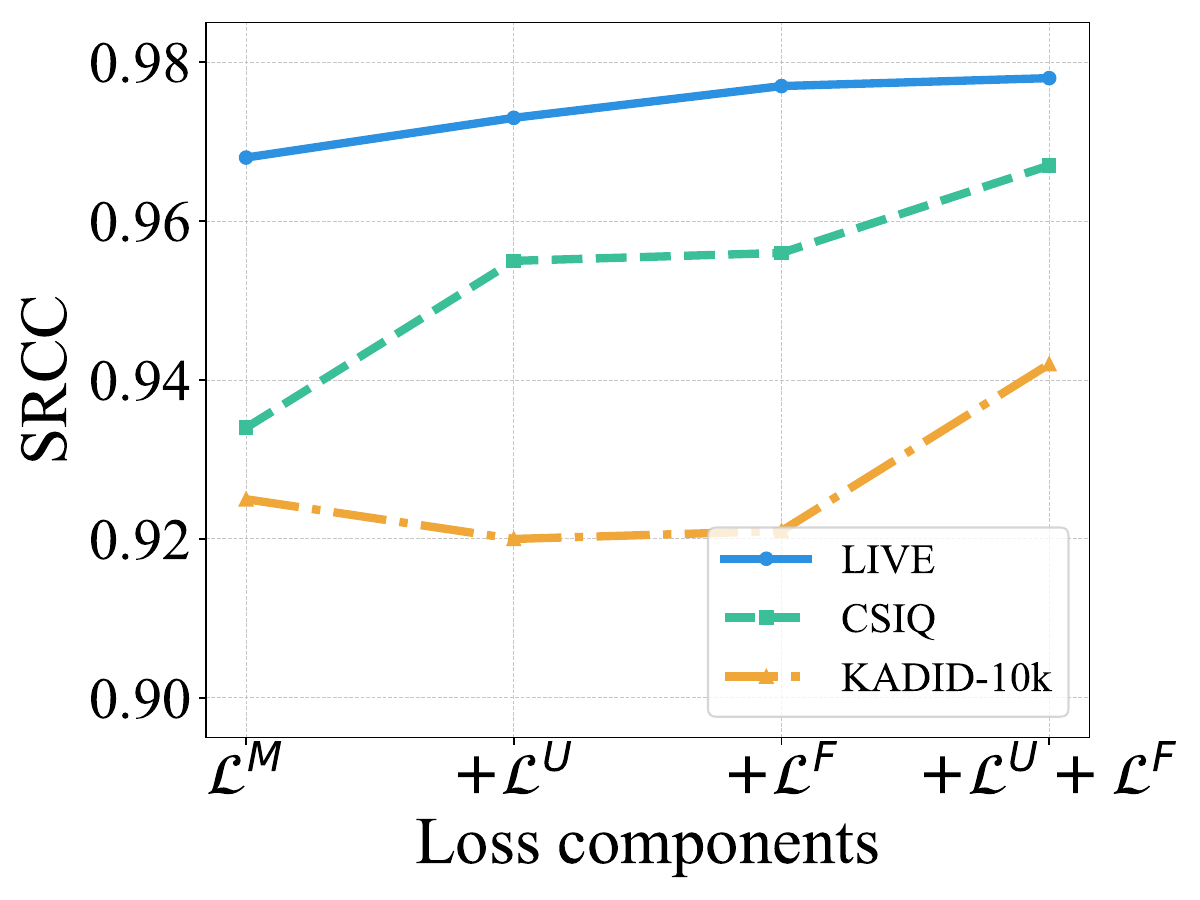}}
        \centerline{(d) Scene and distortion aided}\medskip
    \end{minipage}
    \begin{minipage}[b]{0.245\linewidth}
        \centering
        \centerline{\includegraphics[width=\linewidth]{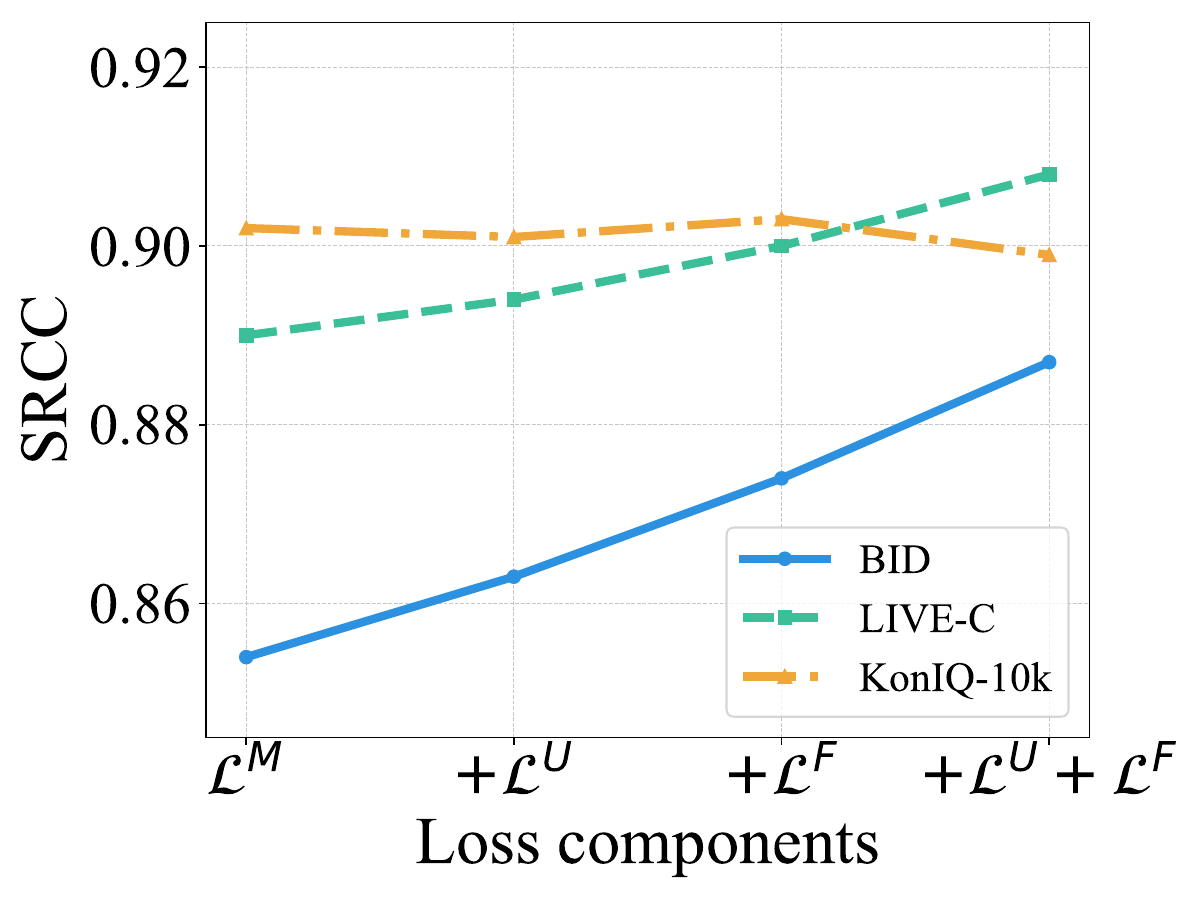}}
        \centerline{(e) IQA task}\medskip
    \end{minipage}
    \begin{minipage}[b]{0.245\linewidth}
        \centering
        \centerline{\includegraphics[width=\linewidth]{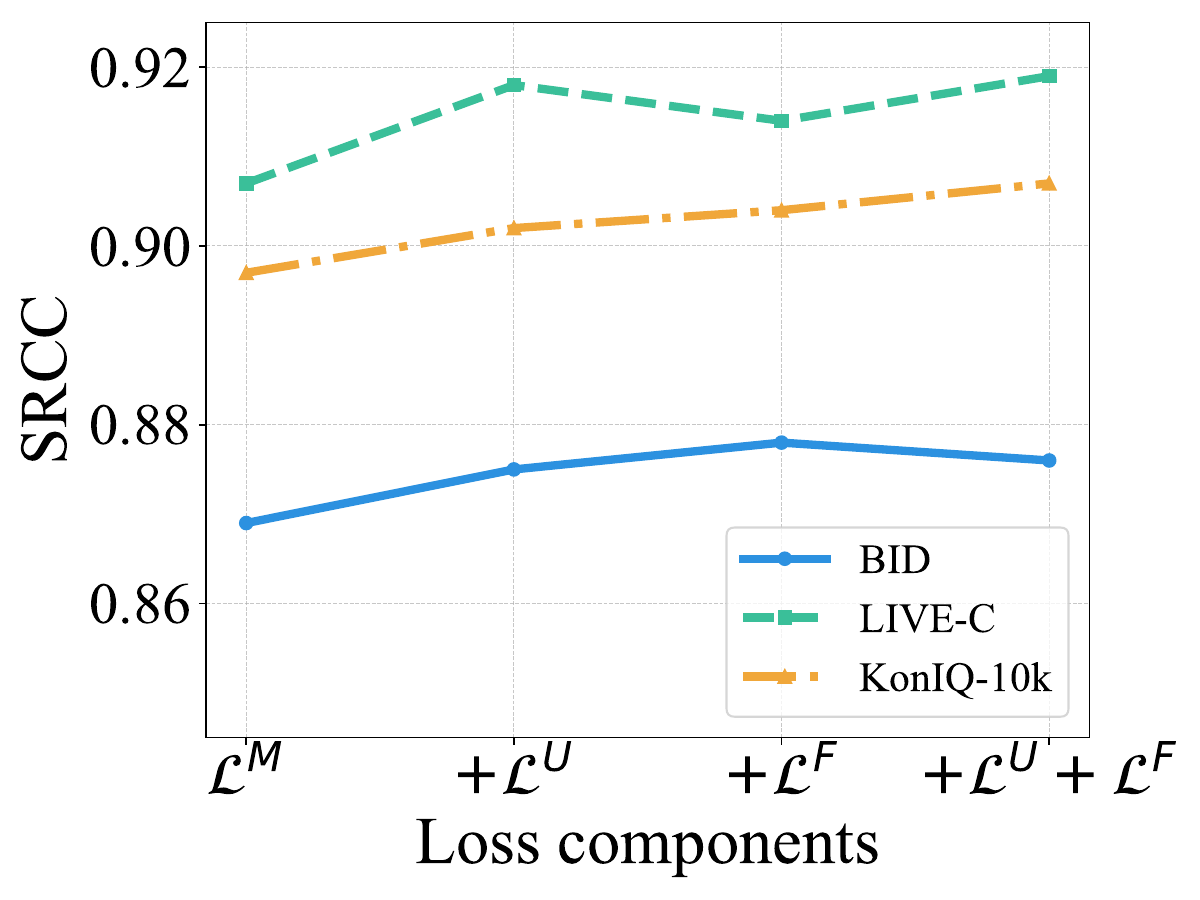}}
        \centerline{(f) Scene aided}\medskip
    \end{minipage}
    \begin{minipage}[b]{0.245\linewidth}
        \centering
        \centerline{\includegraphics[width=\linewidth]{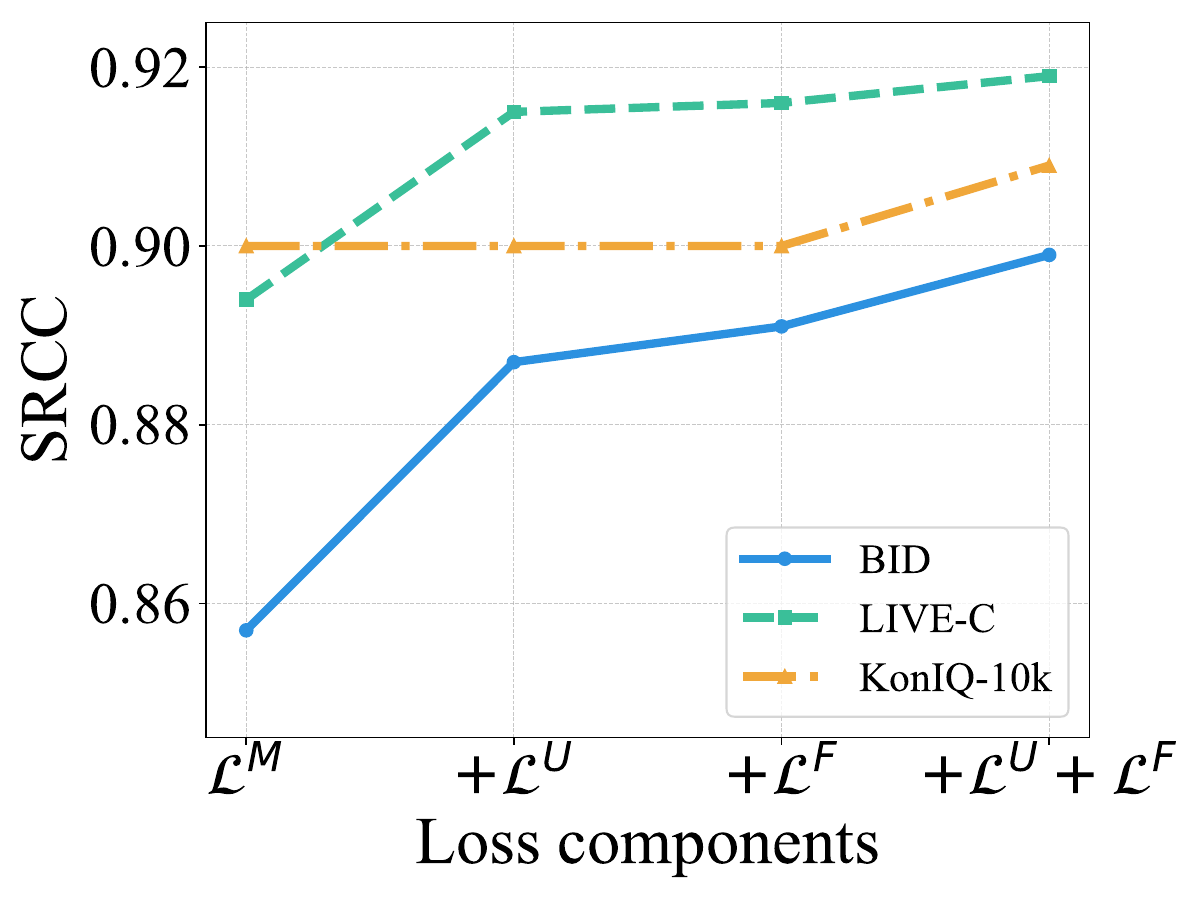}}
        \centerline{(g) Distortion aided}\medskip
    \end{minipage}
    \begin{minipage}[b]{0.245\linewidth}
        \centering
        \centerline{\includegraphics[width=\linewidth]{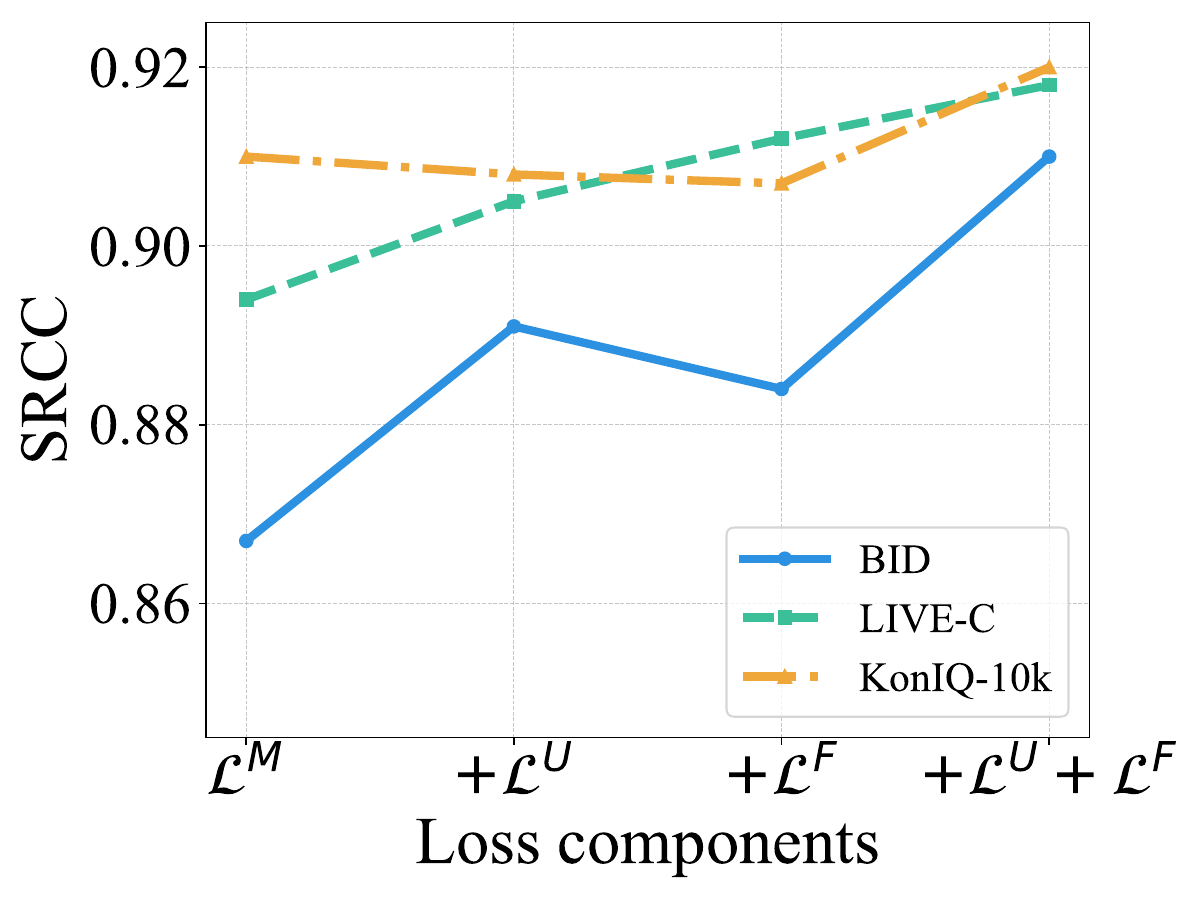}}
        \centerline{(h) Scene and distortion aided}\medskip
    \end{minipage}
    \caption{SRCC performance among different combinations of task assistance for synthetic distortion (a-d) and authentic distortion (e-h).}
    \label{fig:ablation_srcc}
\end{figure*}

\begin{figure*}[htbp]
    \centering
    \begin{minipage}[b]{0.245\linewidth}
        \centering
        \centerline{\includegraphics[width=\linewidth]{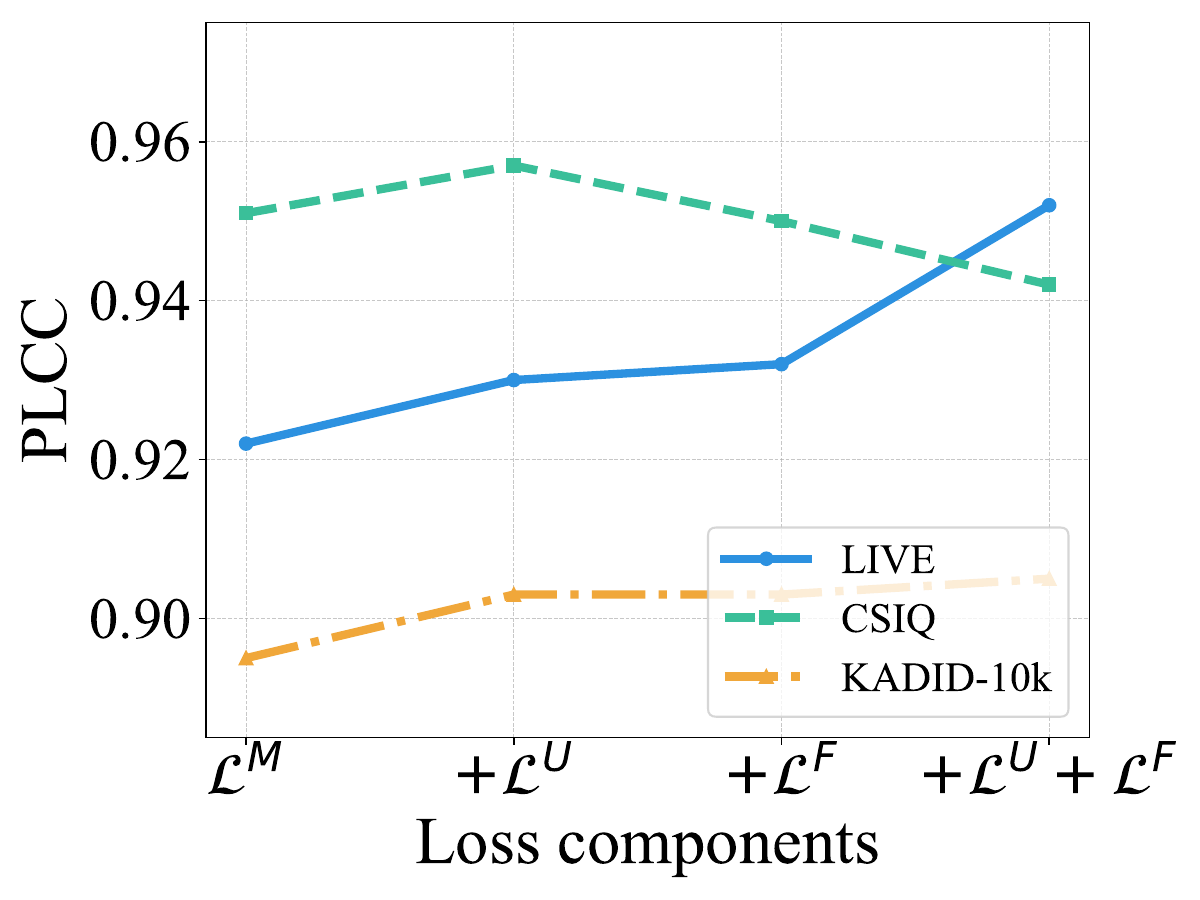}}
        \centerline{(a) IQA task}\medskip
    \end{minipage}
    \begin{minipage}[b]{0.245\linewidth}
        \centering
        \centerline{\includegraphics[width=\linewidth]{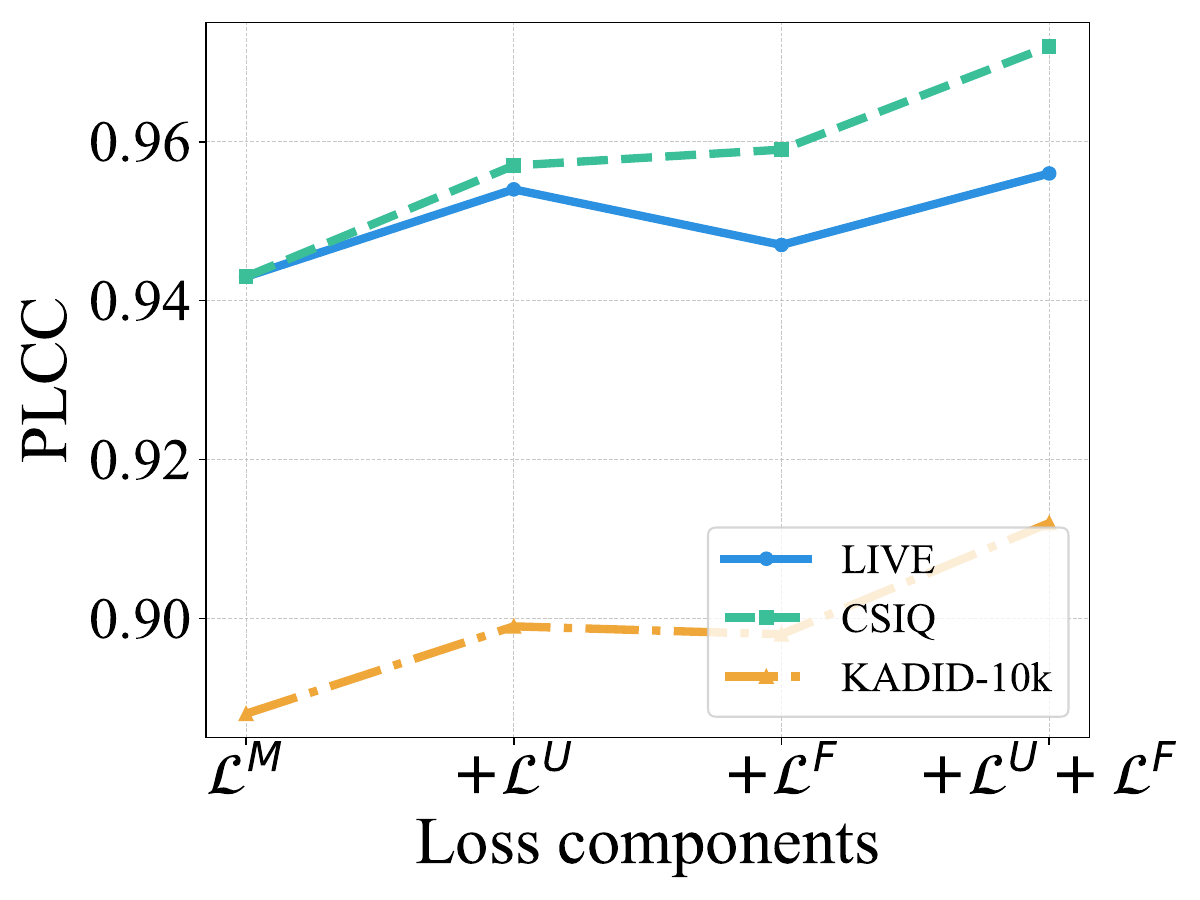}}
        \centerline{(b) Scene aided}\medskip
    \end{minipage}
    \begin{minipage}[b]{0.245\linewidth}
        \centering
        \centerline{\includegraphics[width=\linewidth]{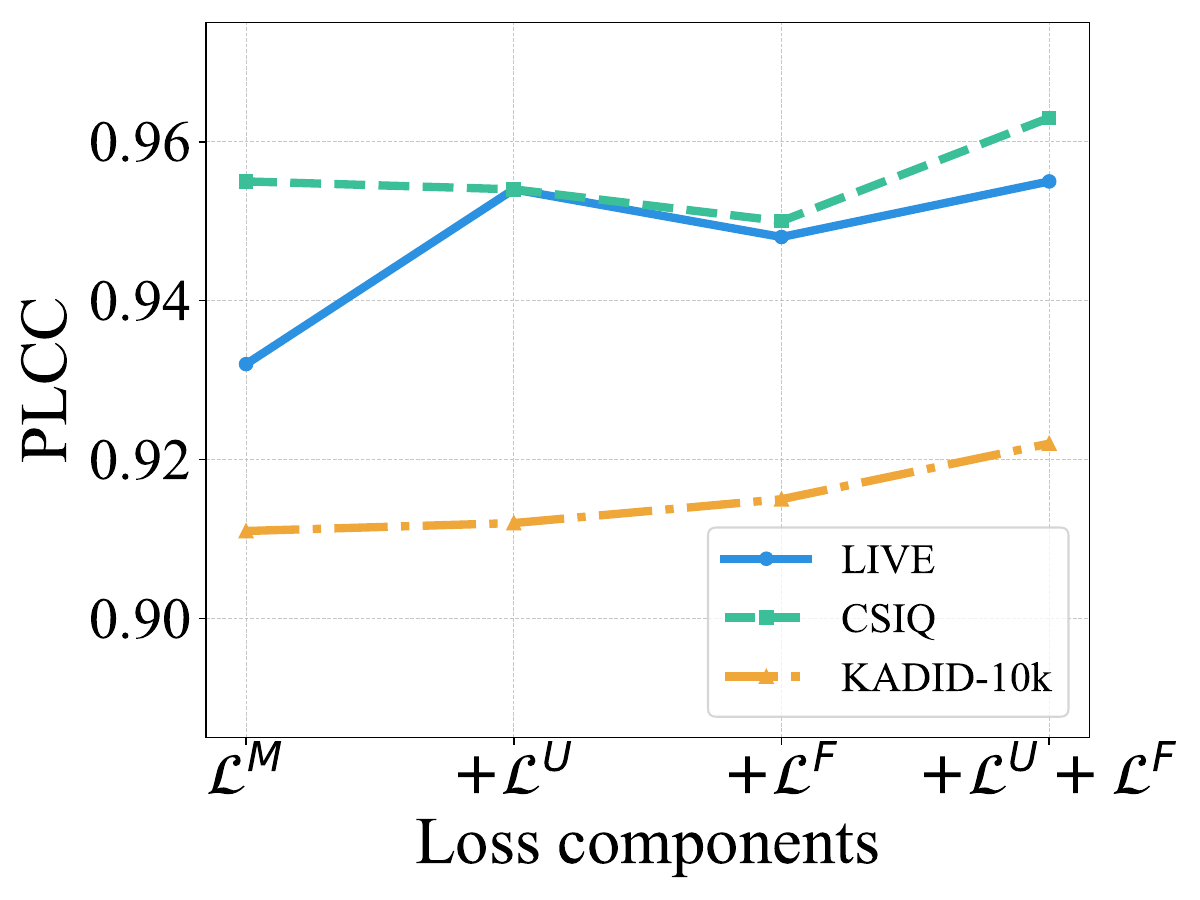}}
        \centerline{(c) Distortion aided}\medskip
    \end{minipage}
    \begin{minipage}[b]{0.245\linewidth}
        \centering
        \centerline{\includegraphics[width=\linewidth]{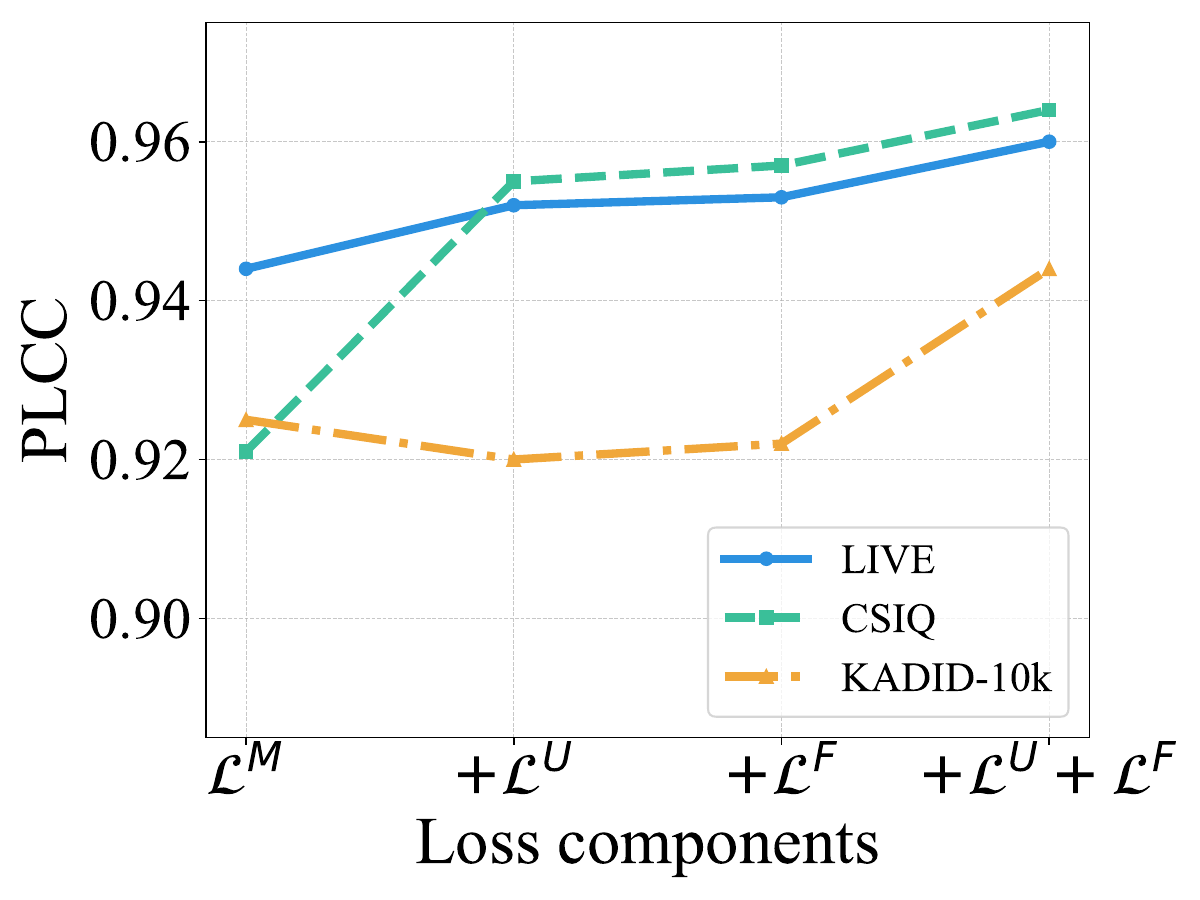}}
        \centerline{(d) Scene and distortion aided}\medskip
    \end{minipage}
    \begin{minipage}[b]{0.245\linewidth}
        \centering
        \centerline{\includegraphics[width=\linewidth]{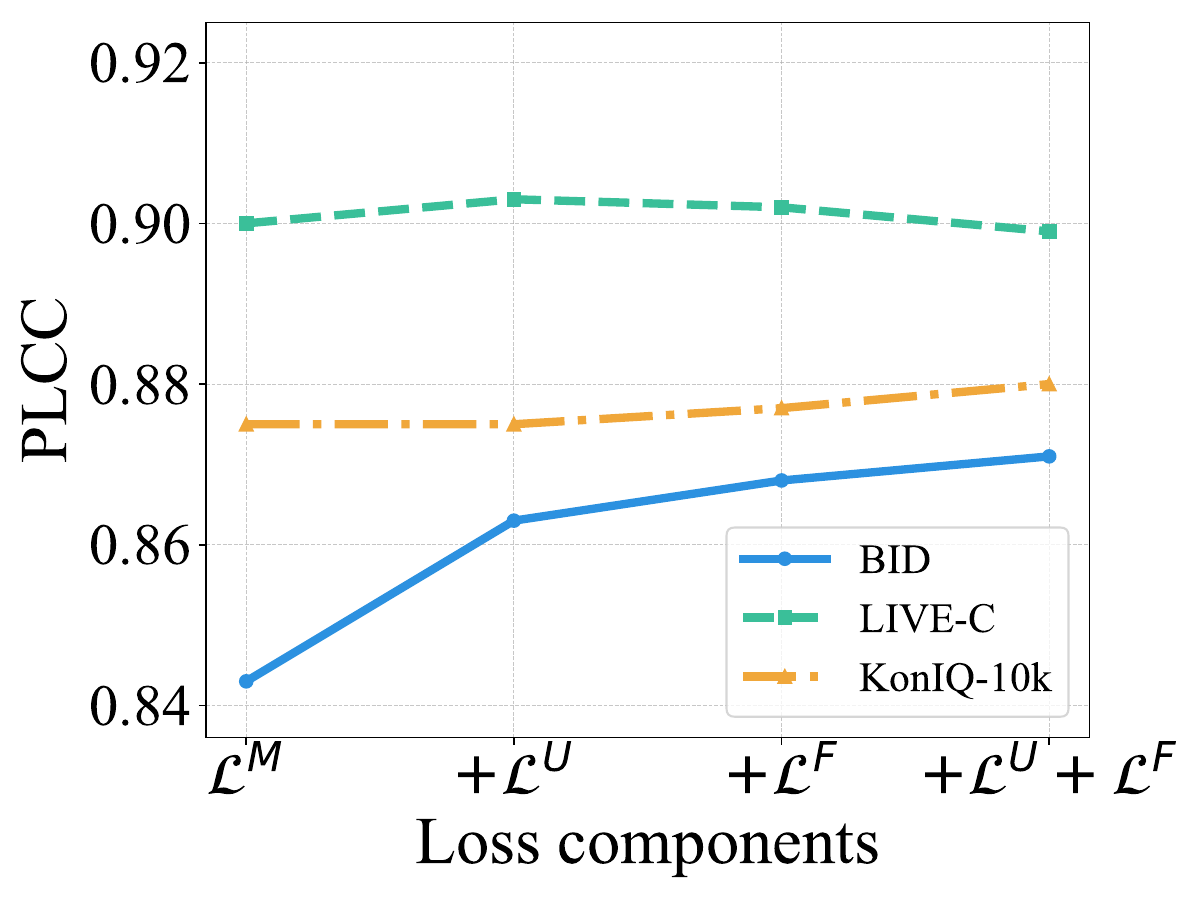}}
        \centerline{(e) IQA task}\medskip
    \end{minipage}
    \begin{minipage}[b]{0.245\linewidth}
        \centering
        \centerline{\includegraphics[width=\linewidth]{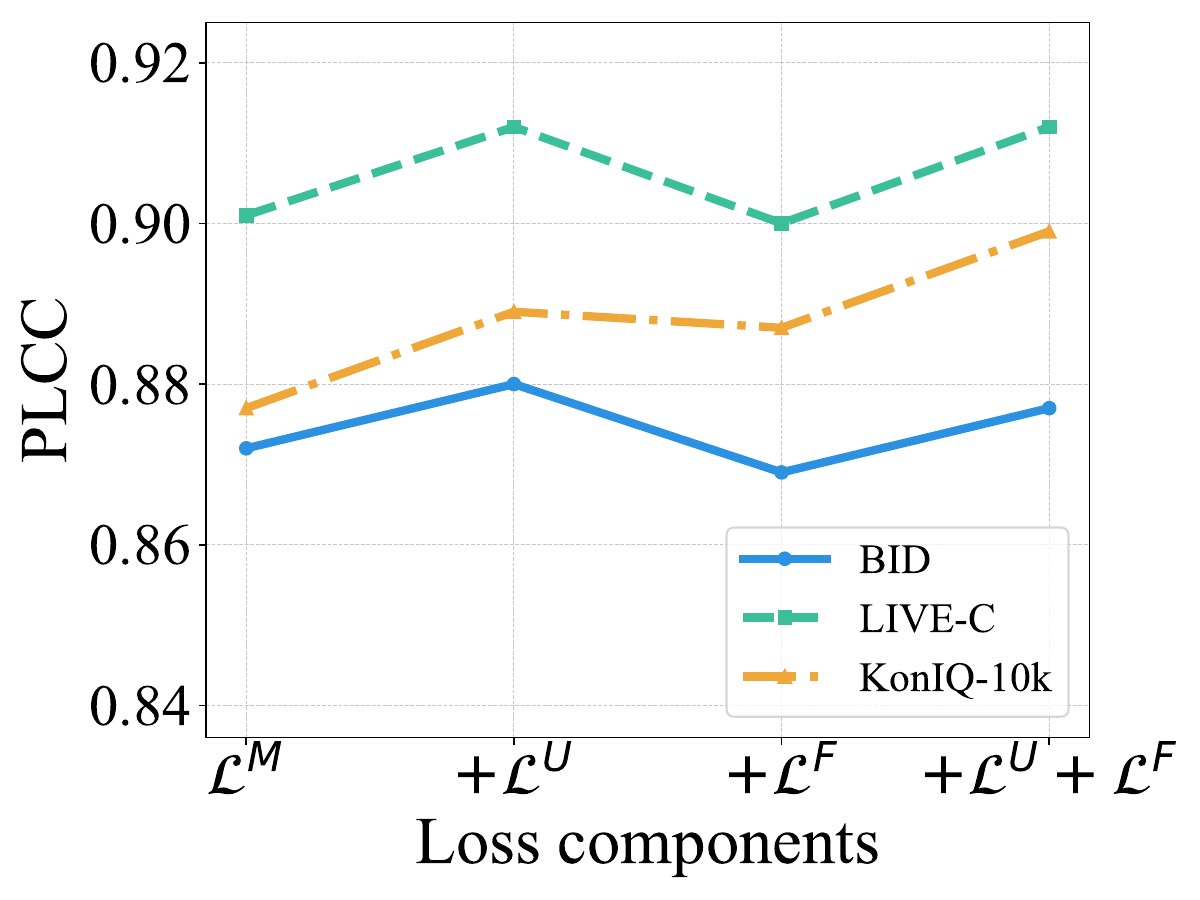}}
        \centerline{(f) Scene aided}\medskip
    \end{minipage}
    \begin{minipage}[b]{0.245\linewidth}
        \centering
        \centerline{\includegraphics[width=\linewidth]{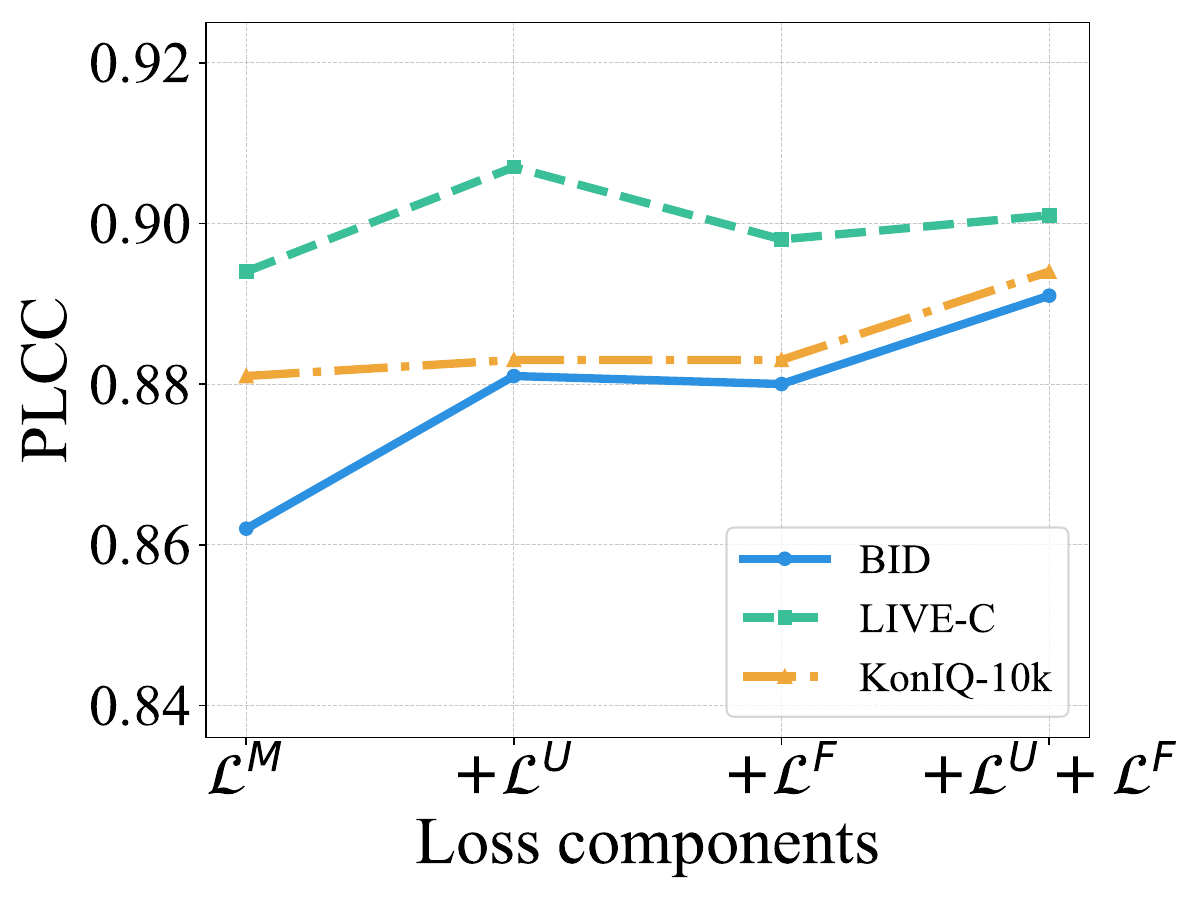}}
        \centerline{(g) Distortion aided}\medskip
    \end{minipage}
    \begin{minipage}[b]{0.245\linewidth}
        \centering
        \centerline{\includegraphics[width=\linewidth]{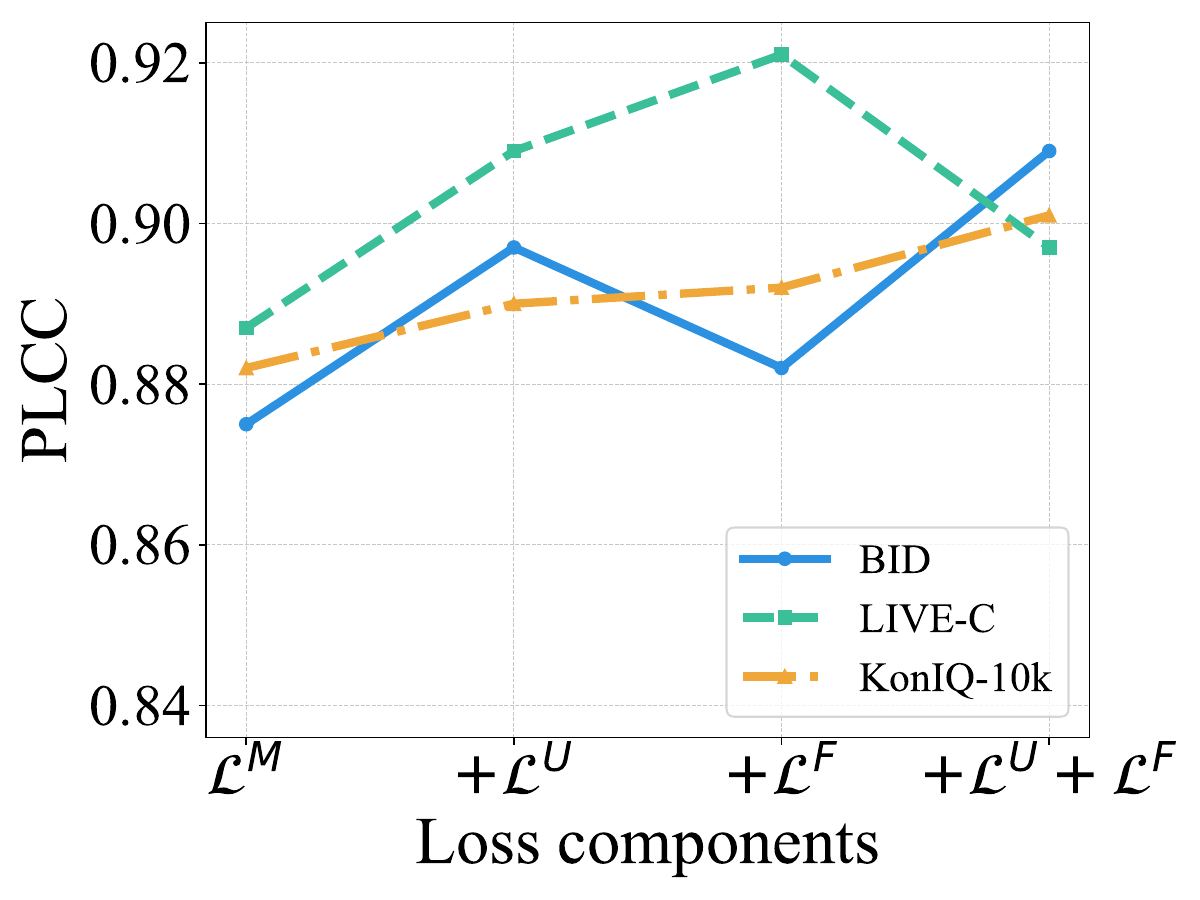}}
        \centerline{(h) Scene and distortion aided}\medskip
    \end{minipage}
    \caption{PLCC performance among different combinations of task assistance for synthetic distortion (a-d) and authentic distortion (e-h).}
    \label{fig:ablation_plcc}
\end{figure*}

\begin{figure*}[htbp]
    \centering
    \begin{minipage}[b]{0.245\linewidth}
        \centering
        \centerline{\includegraphics[width=\linewidth]{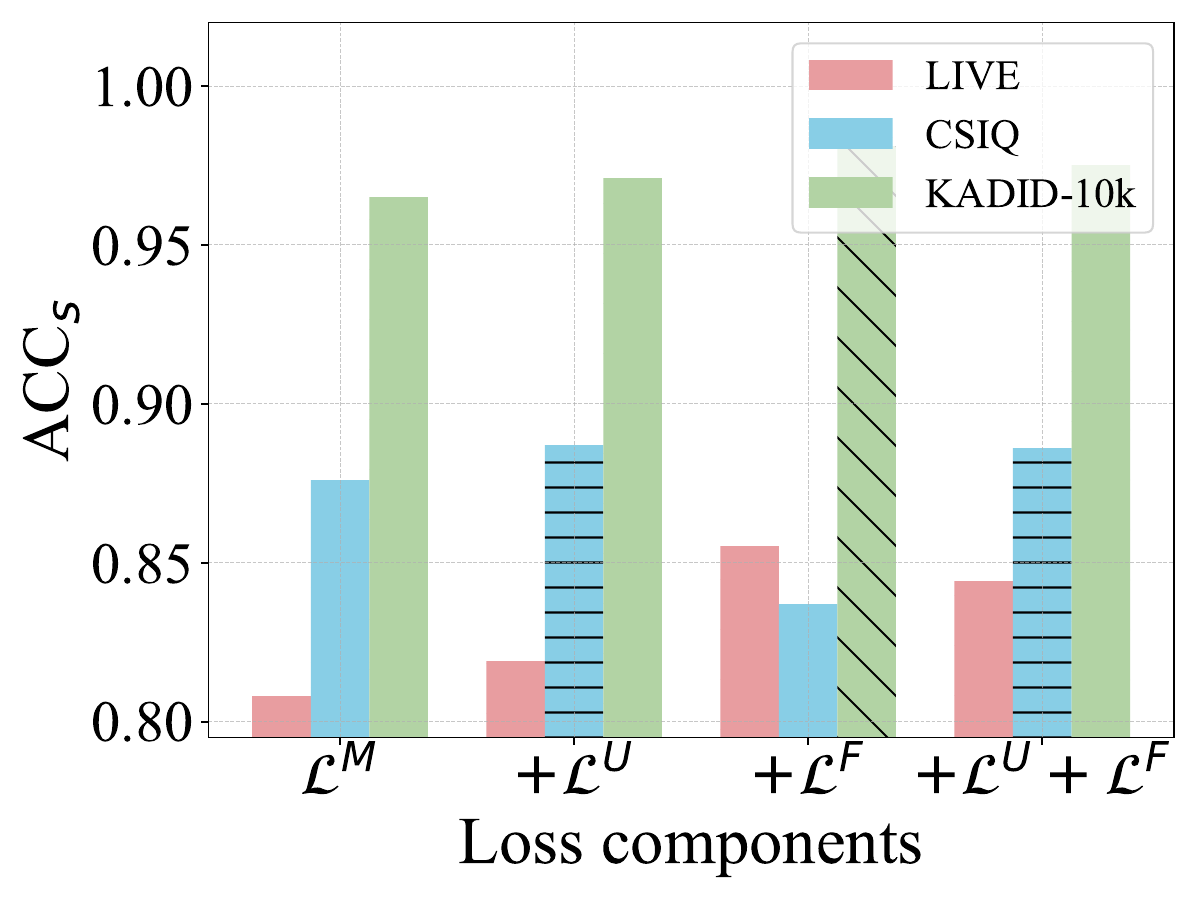}}
        \centerline{(a) Scene aided}\medskip
    \end{minipage}
    \begin{minipage}[b]{0.245\linewidth}
        \centering
        \centerline{\includegraphics[width=\linewidth]{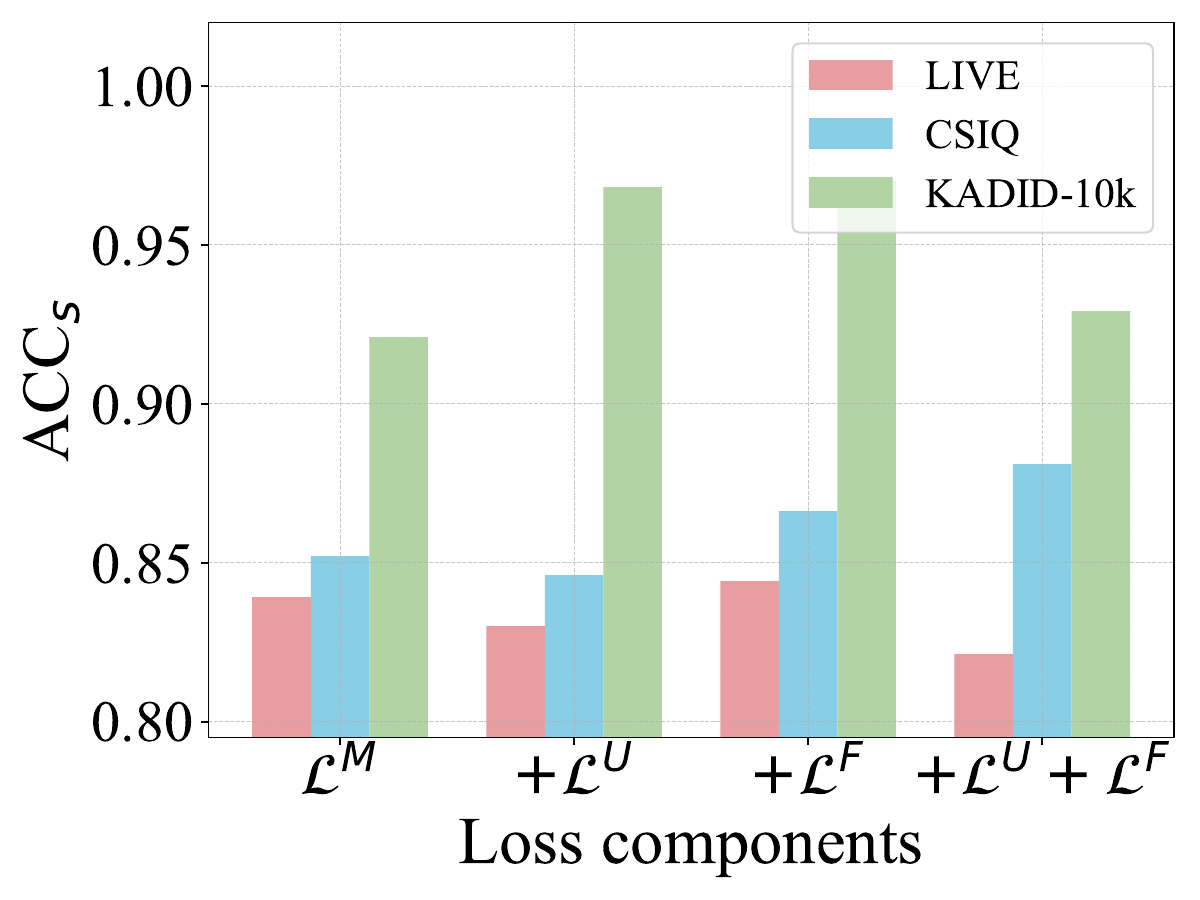}}
        \centerline{(b) Scene and distortion aided}\medskip
    \end{minipage}
    \begin{minipage}[b]{0.245\linewidth}
        \centering
        \centerline{\includegraphics[width=\linewidth]{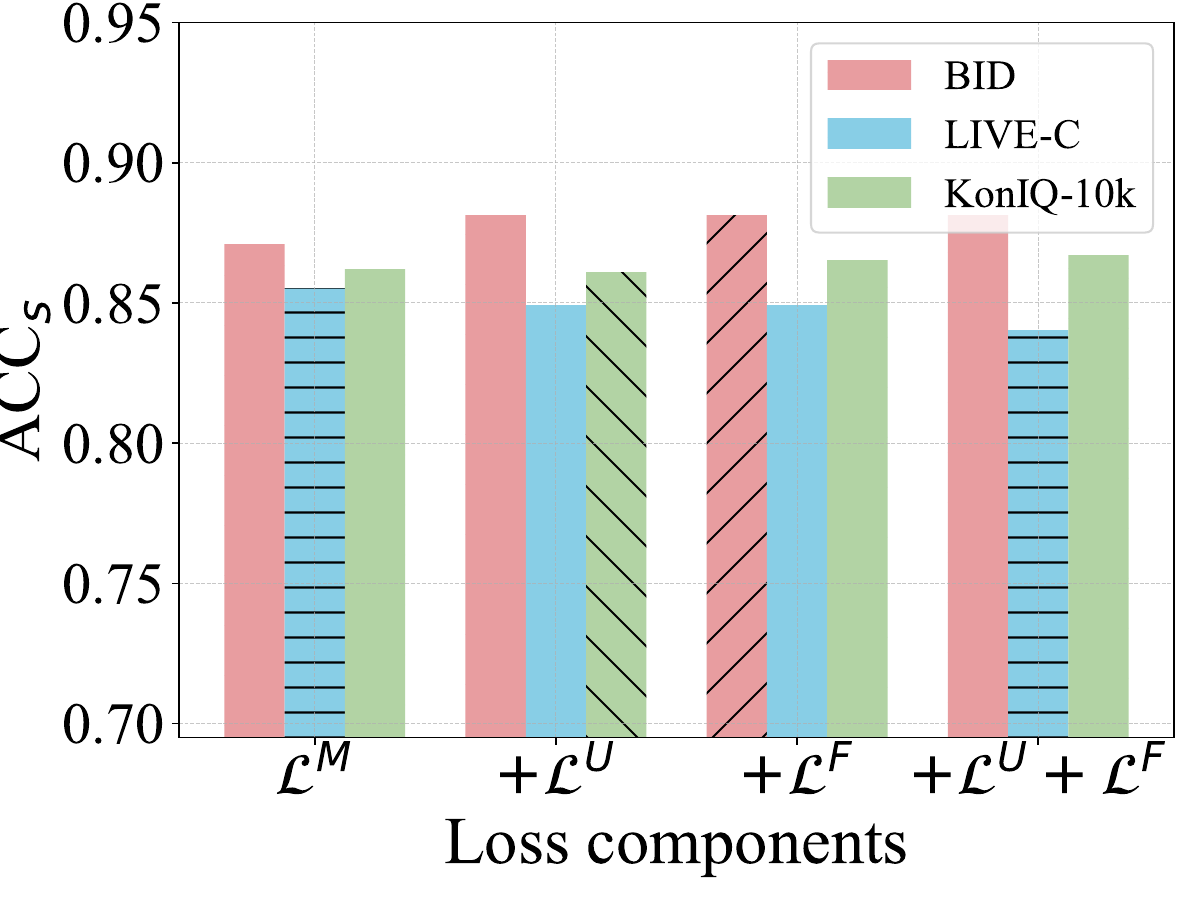}}
        \centerline{(c) Scene aided}\medskip
    \end{minipage}
    \begin{minipage}[b]{0.245\linewidth}
        \centering
        \centerline{\includegraphics[width=\linewidth]{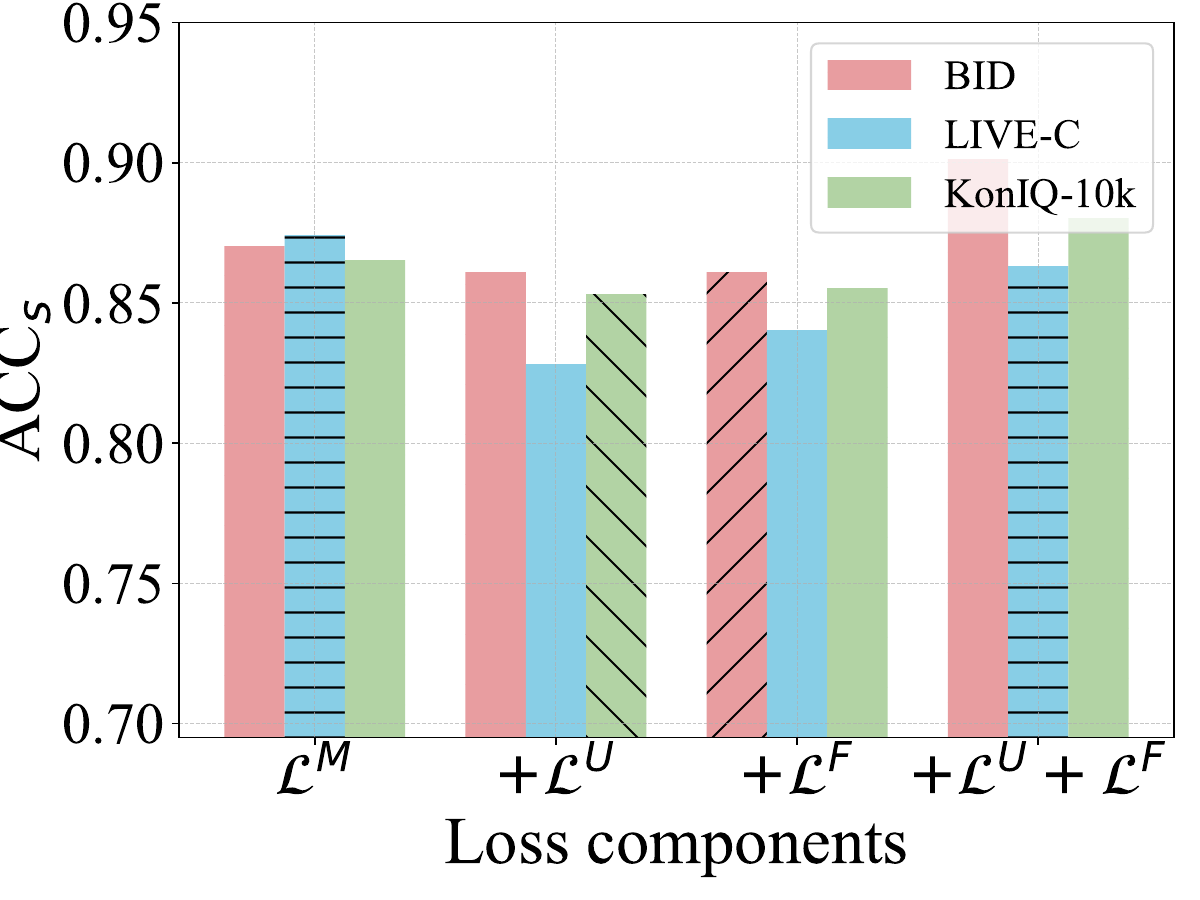}}
        \centerline{(d) Scene and distortion aided}\medskip
    \end{minipage}
    \begin{minipage}[b]{0.245\linewidth}
        \centering
        \centerline{\includegraphics[width=\linewidth]{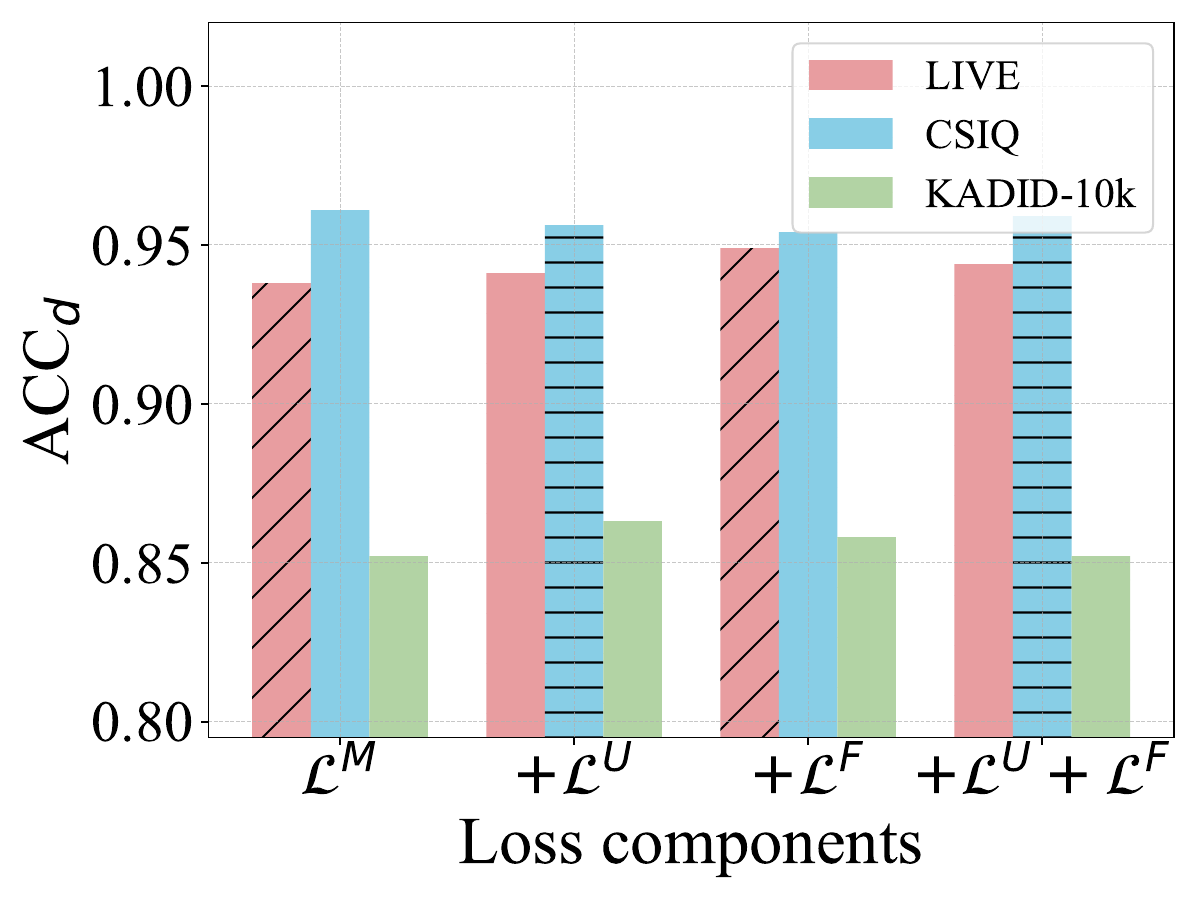}}
        \centerline{(e) Distortion aided}\medskip
    \end{minipage}
    \begin{minipage}[b]{0.245\linewidth}
        \centering
        \centerline{\includegraphics[width=\linewidth]{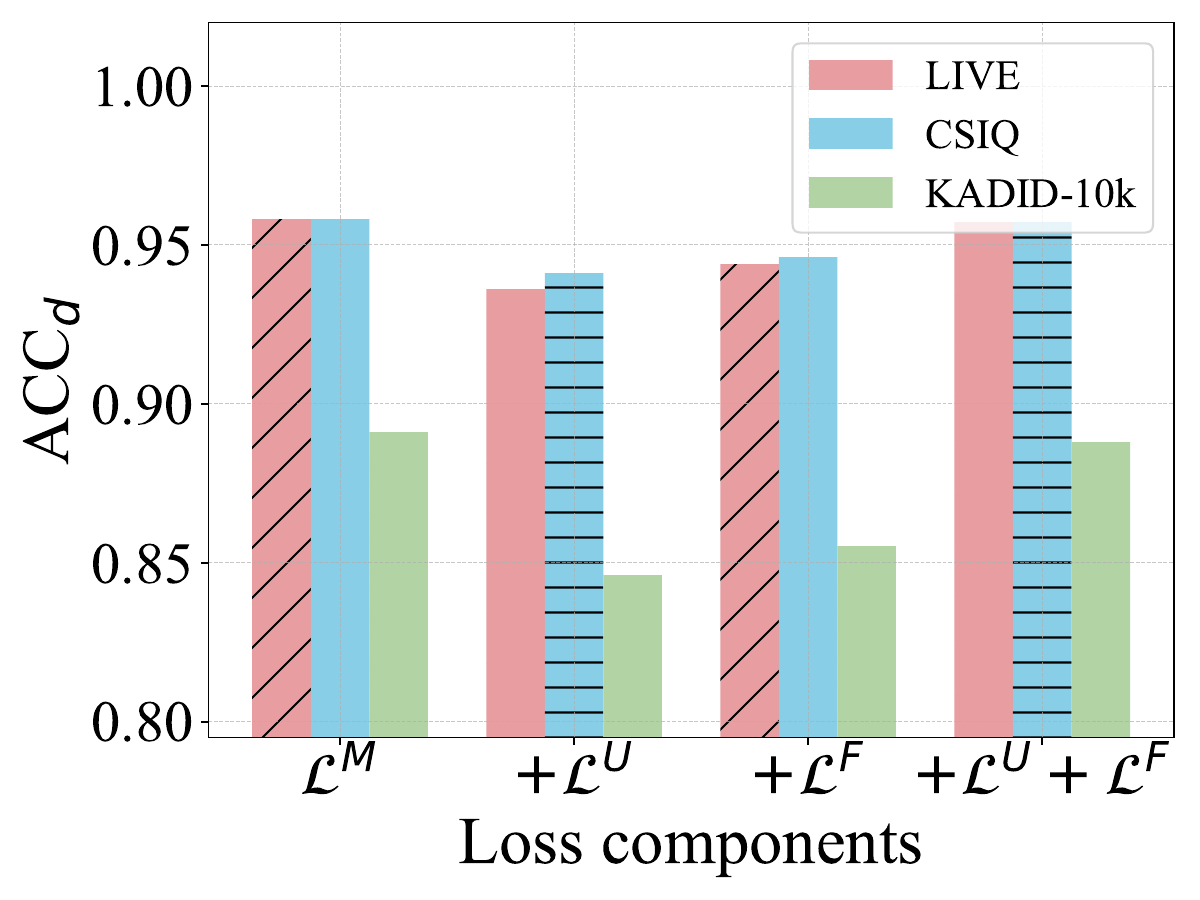}}
        \centerline{(f) Scene and distortion aided}\medskip
    \end{minipage}
    \begin{minipage}[b]{0.245\linewidth}
        \centering
        \centerline{\includegraphics[width=\linewidth]{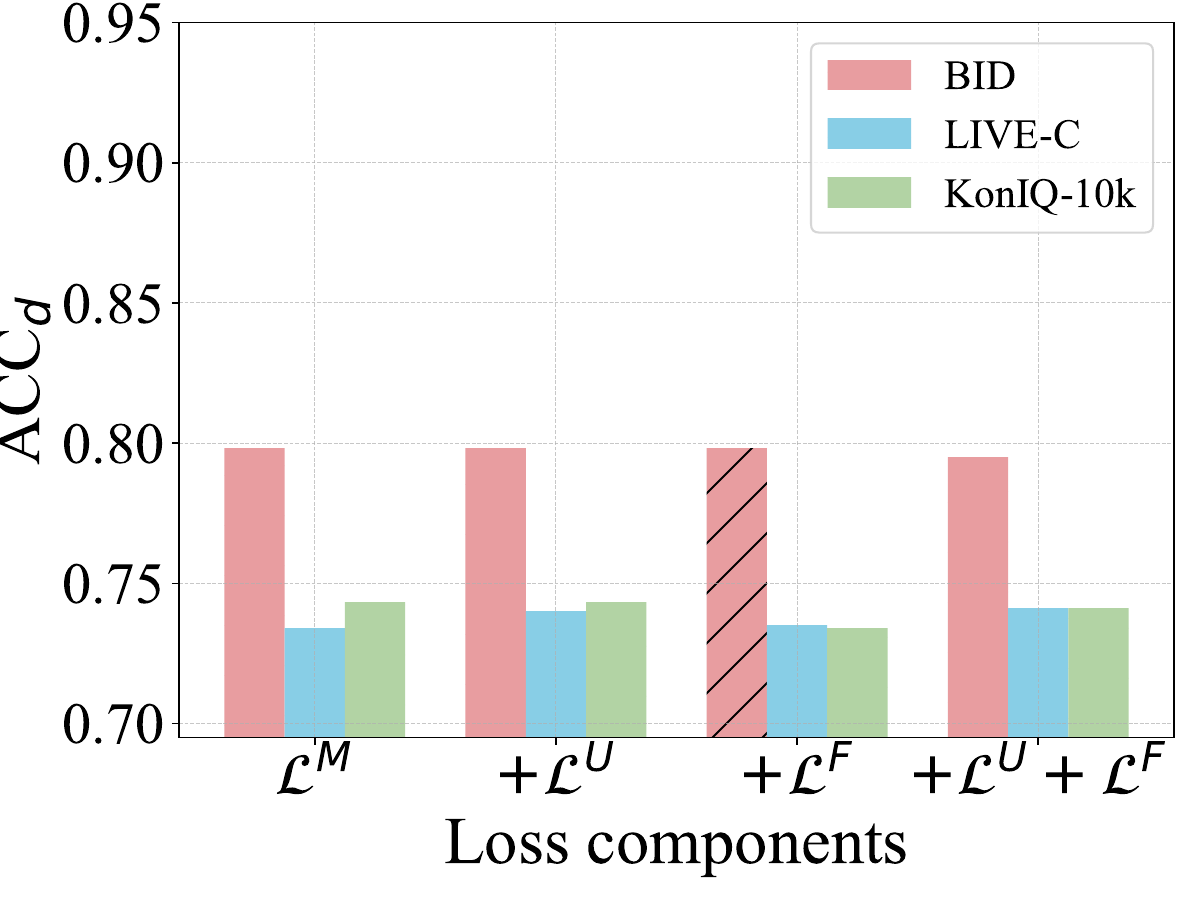}}
        \centerline{(g) Distortion aided}\medskip
    \end{minipage}
    \begin{minipage}[b]{0.245\linewidth}
        \centering
        \centerline{\includegraphics[width=\linewidth]{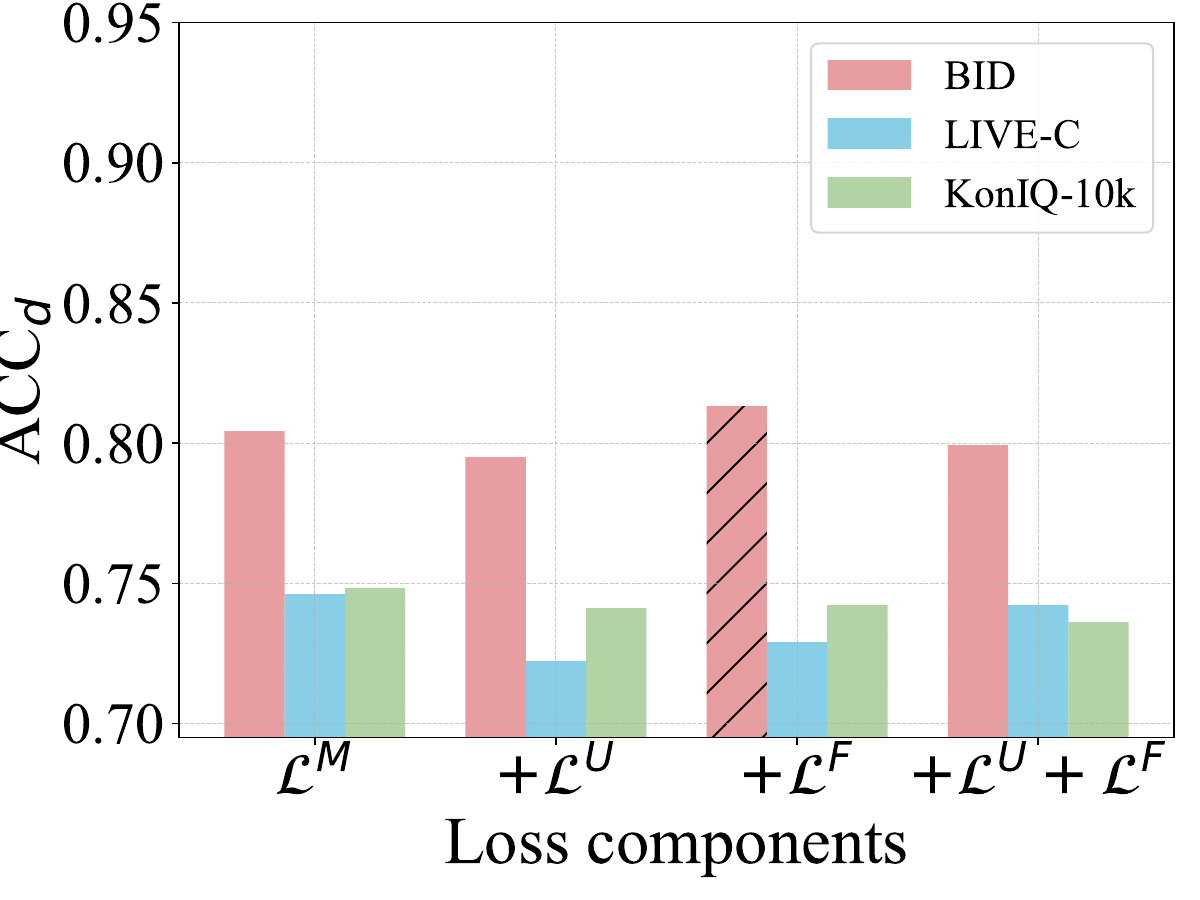}}
        \centerline{(h) Scene and distortion aided}\medskip
    \end{minipage}
    \caption{Accuracy performance in scene classification (a-d) and distortion classification (e-h) among different combinations of task assistance for both synthetic and authentic distortion datasets.}
    \label{fig:ablation_acc}
\end{figure*}

As shown in Section \ref{sec:ablation} and Table \ref{tab:ablation}, there are altogether 16 ablation settings in terms of the combination of auxiliary tasks and loss components.
The SRCC and PLCC performance within different combinations of auxiliary tasks on the six IQA datasets is shown in Figure~\ref{fig:ablation_srcc} and Figure~\ref{fig:ablation_plcc}. The accuracy for specific auxiliary tasks (scene and distortion classification) is shown in Figure~\ref{fig:ablation_acc}.
In most cases, with the addition of cross region loss ($+\mathcal{L}^U$) and cross-grained loss ($+\mathcal{L}^F$), the performance of the model improves.

\subsection{Uncertainty Results}\label{sup:uncertainty}

\begin{figure*}[htbp]
    \centering
    \begin{minipage}[b]{0.245\linewidth}
        \centering
        \centerline{\includegraphics[width=\linewidth]{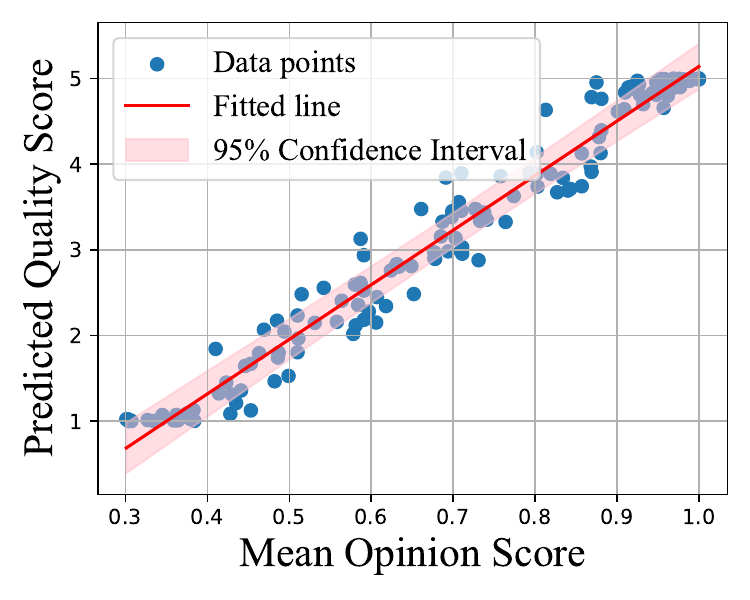}}
        \centerline{(a) DEFNet in CSIQ}\medskip
    \end{minipage}
    \begin{minipage}[b]{0.245\linewidth}
        \centering
        \centerline{\includegraphics[width=\linewidth]{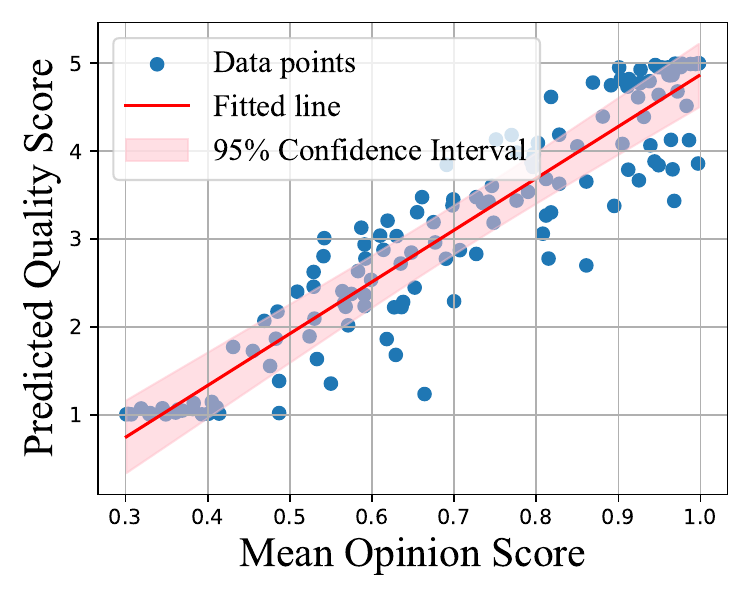}}
        \centerline{(b) LIQE in CSIQ}\medskip
    \end{minipage}
    \begin{minipage}[b]{0.245\linewidth}
        \centering
        \centerline{\includegraphics[width=\linewidth]{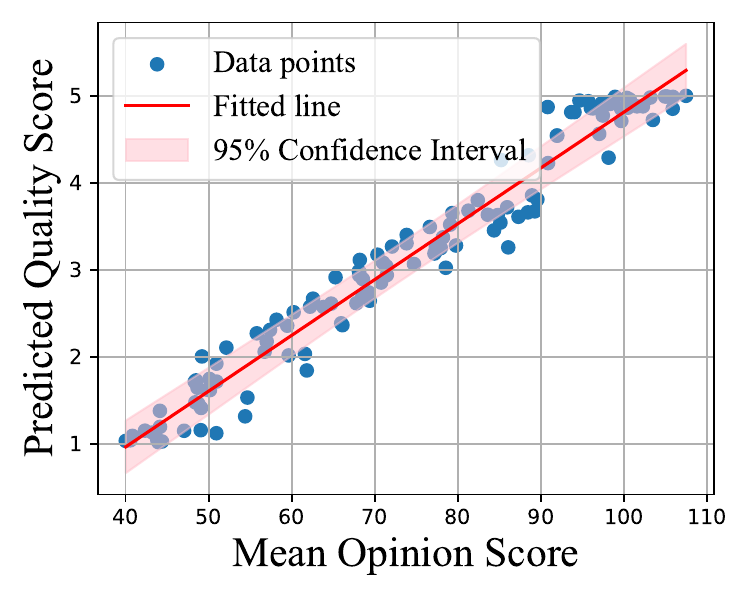}}
        \centerline{(c) DEFNet in LIVE}\medskip
    \end{minipage}
    \begin{minipage}[b]{0.245\linewidth}
        \centering
        \centerline{\includegraphics[width=\linewidth]{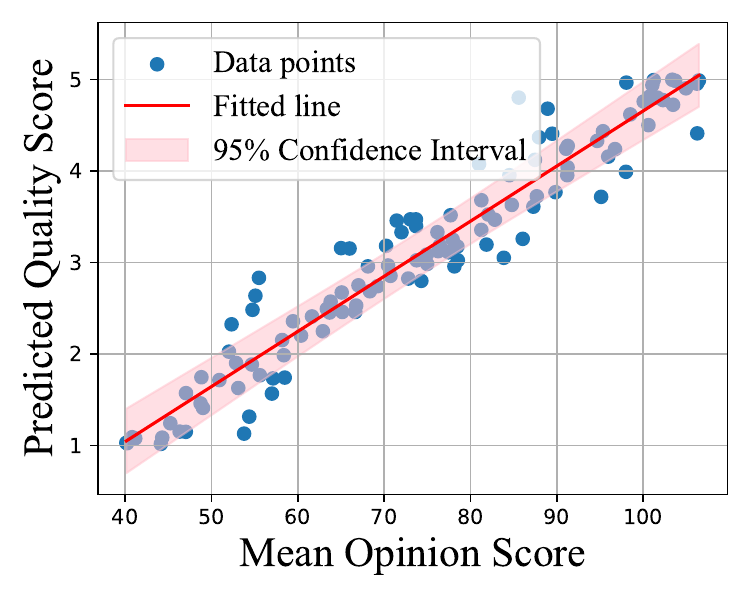}}
        \centerline{(d) LIQE in LIVE}\medskip
    \end{minipage}
    \caption{Scatter plots and 95\% confidence intervals for DEFNet and LIQE \cite{Zhang_2023_CVPR} in CSIQ and LIVE datasets.}
    \label{fig:ci}
\end{figure*}

Scatter plots between the predicted and ground-truth scores, and their 95\% confidence intervals are shown in Figure \ref{fig:ci}.
The experimental results indicate that the introduction of the two-level trustworthy evidential fusion reduces the uncertainty of model predictions.
By applying evidence theory to image quality assessment tasks and utilizing the four dimensions of data distribution, final predictions are made for the three tasks of image quality assessment. 
By combining aleatoric and epistemic uncertainty with evidential learning, an optimization is carried out for the final prediction, allowing the model to better focus on the parts with significant fluctuations in the prediction results and focus on learning relevant regions. In addition, reallocating uncertainty between cross sub-regions and different granularity fusion can also enable targeted optimization of the model.

\subsection{Model Complexity}\label{sup:complexity}

In this section, we analyze the model complexity of the proposed DEFNet and several state-of-the-art methods. Table~\ref{tab:params} presents the comparison results in terms of the number of parameters (\# Params), which provide insights into the computational efficiency and resource demands of each model. 
It can be observed that DEFNet, with 84.22M parameters, is higher than LIQE in terms of the number and complexity of model parameters. 
In comparison with TReS~\cite{Golestaneh_2022_WACV}, CDINet~\cite{10440553} and TSFE~\cite{lou2023refining}, DEFNet is still competitive.
Despite this, there is still room for improvement in parameter and complexity optimization.

\begin{table}[t]
  \centering
  \fontsize{9}{10}\selectfont %{字体尺寸}{行距}
  \caption{Model complexity comparison.}
  \label{tab:params}
  \begin{tabular}{lc}
      \toprule
      Methods  & \# Params (M) \\
      \midrule
      % % PQR~\cite{zeng2017probabilistic}   &  23.51     \\
      % DBCNN~\cite{8576582}               &  15.31     \\
      % VCRNet~\cite{pan2022vcrnet}        &  11.41     \\
      % HyperIQA~\cite{Su_2020_CVPR}       &  27.38     \\
      TReS~\cite{Golestaneh_2022_WACV}   &  152.45    \\
      CDINet~\cite{10440553}             &  99.62      \\
      TSFE~\cite{lou2023refining}        &  91.75      \\
      LIQE~\cite{Zhang_2023_CVPR}        &  59.02      \\
      \midrule
      DEFNet                             &  84.22   \\
      \bottomrule  
  \end{tabular}
\end{table}

%% file: main.bbl
\begin{thebibliography}{71}
\providecommand{\natexlab}[1]{#1}
\providecommand{\url}[1]{\texttt{#1}}
\expandafter\ifx\csname urlstyle\endcsname\relax
  \providecommand{\doi}[1]{doi: #1}\else
  \providecommand{\doi}{doi: \begingroup \urlstyle{rm}\Url}\fi

\bibitem[Amini et~al.(2020)Amini, Schwarting, Soleimany, and Rus]{NEURIPS2020_aab08546}
Alexander Amini, Wilko Schwarting, Ava Soleimany, and Daniela Rus.
\newblock Deep evidential regression.
\newblock In \emph{Advances in Neural Information Processing Systems}, pages 14927--14937, 2020.

\bibitem[Bai et~al.(2025)Bai, Yin, Dong, Zhang, Pun, and Chen]{bai2025lensnet}
Jiesong Bai, Yuhao Yin, Yihang Dong, Xiaofeng Zhang, Chi-Man Pun, and Xuhang Chen.
\newblock Lensnet: An end-to-end learning framework for empirical point spread function modeling and lensless imaging reconstruction.
\newblock \emph{arXiv preprint arXiv:2505.01755}, 2025.

\bibitem[Bosse et~al.(2016)Bosse, Maniry, Wiegand, and Samek]{bosse2016deep}
Sebastian Bosse, Dominique Maniry, Thomas Wiegand, and Wojciech Samek.
\newblock A deep neural network for image quality assessment.
\newblock In \emph{IEEE international conference on image processing}, pages 3773--3777. IEEE, 2016.

\bibitem[Chen et~al.(2024)Chen, Mo, Hou, Wu, Liao, Sun, Yan, and Lin]{chen2024topiq}
Chaofeng Chen, Jiadi Mo, Jingwen Hou, Haoning Wu, Liang Liao, Wenxiu Sun, Qiong Yan, and Weisi Lin.
\newblock Topiq: A top-down approach from semantics to distortions for image quality assessment.
\newblock \emph{IEEE Transactions on Image Processing}, 33:\penalty0 2404--2418, 2024.

\bibitem[Chen et~al.(2025{\natexlab{a}})Chen, Li, Shen, Mahmud, Pham, Pun, and Wang]{chen2025high}
Xuhang Chen, Zhuo Li, Yanyan Shen, Mufti Mahmud, Hieu Pham, Chi-Man Pun, and Shuqiang Wang.
\newblock High-fidelity functional ultrasound reconstruction via a visual auto-regressive framework.
\newblock \emph{arXiv preprint arXiv:2505.21530}, 2025{\natexlab{a}}.

\bibitem[Chen et~al.(2025{\natexlab{b}})Chen, Xu, Lou, Li, Ding, and Huang]{chen2025semi}
Yanyuan Chen, Dexuan Xu, Yiwei Lou, Hang Li, Weiping Ding, and Yu Huang.
\newblock Semi-supervised medical image classification via cross-training and dual-teacher fusion model.
\newblock \emph{Information Fusion}, page 103389, 2025{\natexlab{b}}.

\bibitem[Ciancio et~al.(2011)Ciancio, Targino~da Costa, da~Silva, Said, Samadani, and Obrador]{5492198}
Alexandre Ciancio, André Luiz N~Targino Targino~da Costa, Eduardo A.~B. da Silva, Amir Said, Ramin Samadani, and Pere Obrador.
\newblock No-reference blur assessment of digital pictures based on multifeature classifiers.
\newblock \emph{IEEE Transactions on Image Processing}, 20\penalty0 (1):\penalty0 64--75, 2011.

\bibitem[Dempster(2008)]{Dempster2008}
Arthur~P Dempster.
\newblock Upper and lower probabilities induced by a multivalued mapping.
\newblock In \emph{Classic works of the Dempster-Shafer theory of belief functions}, pages 57--72. Springer, 2008.

\bibitem[Fang et~al.(2020)Fang, Zhu, Zeng, Ma, and Wang]{Fang_2020_CVPR}
Yuming Fang, Hanwei Zhu, Yan Zeng, Kede Ma, and Zhou Wang.
\newblock Perceptual quality assessment of smartphone photography.
\newblock In \emph{Proceedings of the IEEE/CVF Conference on Computer Vision and Pattern Recognition}, pages 3677--3686, 2020.

\bibitem[Ghadiyaram and Bovik(2016)]{7327186}
Deepti Ghadiyaram and Alan~C. Bovik.
\newblock Massive online crowdsourced study of subjective and objective picture quality.
\newblock \emph{IEEE Transactions on Image Processing}, 25\penalty0 (1):\penalty0 372--387, 2016.

\bibitem[Golestaneh et~al.(2022)Golestaneh, Dadsetan, and Kitani]{Golestaneh_2022_WACV}
S.~Alireza Golestaneh, Saba Dadsetan, and Kris~M. Kitani.
\newblock No-reference image quality assessment via transformers, relative ranking, and self-consistency.
\newblock In \emph{Proceedings of the IEEE/CVF Winter Conference on Applications of Computer Vision (WACV)}, pages 1220--1230, 2022.

\bibitem[Gu et~al.(2020)Gu, Cai, Chen, Ye, Jimmy~S, and Dong]{jinjin2020pipal}
Jinjin Gu, Haoming Cai, Haoyu Chen, Xiaoxing Ye, Ren Jimmy~S, and Chao Dong.
\newblock Pipal: a large-scale image quality assessment dataset for perceptual image restoration.
\newblock In \emph{Computer Vision--ECCV 2020: 16th European Conference, Glasgow, UK, August 23--28, 2020, Proceedings, Part XI 16}, pages 633--651. Springer, 2020.

\bibitem[Guo et~al.(2025{\natexlab{a}})Guo, Chen, Wang, and Pun]{underwater2025guo}
Xiaojiao Guo, Xuhang Chen, Shuqiang Wang, and Chi-Man Pun.
\newblock Underwater image restoration through a prior guided hybrid sense approach and extensive benchmark analysis.
\newblock \emph{IEEE Transactions on Circuits and Systems for Video Technology}, 35\penalty0 (5):\penalty0 4784--4800, 2025{\natexlab{a}}.

\bibitem[Guo et~al.(2025{\natexlab{b}})Guo, Dong, Chen, Chen, Li, Zheng, and Pun]{guo2025underwater}
Xiaojiao Guo, Yihang Dong, Xuhang Chen, Weiwen Chen, Zimeng Li, FuChen Zheng, and Chi-Man Pun.
\newblock Underwater image restoration via polymorphic large kernel cnns.
\newblock In \emph{IEEE International Conference on Acoustics, Speech and Signal Processing}, pages 1--5, 2025{\natexlab{b}}.

\bibitem[Guo et~al.(2025{\natexlab{c}})Guo, Luo, Dong, Liang, Li, Zhang, and Chen]{guo2025asymmetric}
Xiaojiao Guo, Shenghong Luo, Yihang Dong, Zexiao Liang, Zimeng Li, Xiujun Zhang, and Xuhang Chen.
\newblock An asymmetric calibrated transformer network for underwater image restoration.
\newblock \emph{The Visual Computer}, pages 1--13, 2025{\natexlab{c}}.

\bibitem[He(2024)]{he2024epl}
Yuanpeng He.
\newblock Epl: Evidential prototype learning for semi-supervised medical image segmentation.
\newblock \emph{arXiv preprint arXiv:2404.06181}, 2024.

\bibitem[He and Deng(2022)]{he2022mmget}
Yuanpeng He and Yong Deng.
\newblock Mmget: A markov model for generalized evidence theory.
\newblock \emph{Computational and Applied Mathematics}, 41\penalty0 (1):\penalty0 9, 2022.

\bibitem[He and Xiao(2021)]{he2021conflicting}
Yuanpeng He and Fuyuan Xiao.
\newblock Conflicting management of evidence combination from the point of improvement of basic probability assignment.
\newblock \emph{International Journal of Intelligent Systems}, 36\penalty0 (5):\penalty0 1914--1942, 2021.

\bibitem[He et~al.(2024{\natexlab{a}})He, Bi, Li, Pun, Jiao, and Jin]{he2024mutual}
Yuanpeng He, Yali Bi, Lijian Li, Chi-Man Pun, Wenpin Jiao, and Zhi Jin.
\newblock Mutual evidential deep learning for semi-supervised medical image segmentation.
\newblock In \emph{IEEE International Conference on Bioinformatics and Biomedicine}, pages 2010--2017. IEEE, 2024{\natexlab{a}}.

\bibitem[He et~al.(2024{\natexlab{b}})He, Li, Zhan, Jiao, and Pun]{he2024generalized}
Yuanpeng He, Lijian Li, Tianxiang Zhan, Wenpin Jiao, and Chi-Man Pun.
\newblock Generalized uncertainty-based evidential fusion with hybrid multi-head attention for weak-supervised temporal action localization.
\newblock In \emph{IEEE International Conference on Acoustics, Speech and Signal Processing}, pages 3855--3859. IEEE, 2024{\natexlab{b}}.

\bibitem[He et~al.(2025)He, Li, Zhan, Pun, Jiao, and Jin]{he2025co}
Yuanpeng He, Lijian Li, Tianxiang Zhan, Chi-Man Pun, Wenpin Jiao, and Zhi Jin.
\newblock Co-evidential fusion with information volume for semi-supervised medical image segmentation.
\newblock \emph{Pattern Recognition}, 166:\penalty0 111639, 2025.

\bibitem[Hosu et~al.(2020)Hosu, Lin, Sziranyi, and Saupe]{8968750}
Vlad Hosu, Hanhe Lin, Tamas Sziranyi, and Dietmar Saupe.
\newblock Koniq-10k: An ecologically valid database for deep learning of blind image quality assessment.
\newblock \emph{IEEE Transactions on Image Processing}, 29:\penalty0 4041--4056, 2020.

\bibitem[Huang et~al.(2019)Huang, Tian, Jiang, and Chen]{huang2019convolutional}
Yuge Huang, Xiang Tian, Rongxin Jiang, and Yaowu Chen.
\newblock Convolutional neural network with uncertainty estimates for no-reference image quality assessment.
\newblock In \emph{Tenth International Conference on Graphics and Image Processing}, pages 401--407. SPIE, 2019.

\bibitem[Ke et~al.(2021)Ke, Wang, Wang, Milanfar, and Yang]{Ke_2021_ICCV}
Junjie Ke, Qifei Wang, Yilin Wang, Peyman Milanfar, and Feng Yang.
\newblock Musiq: Multi-scale image quality transformer.
\newblock In \emph{Proceedings of the IEEE/CVF International Conference on Computer Vision}, pages 5148--5157, 2021.

\bibitem[Larson and Chandler(2010)]{10.1117/1.3267105}
Eric~Cooper Larson and Damon~Michael Chandler.
\newblock {Most apparent distortion: full-reference image quality assessment and the role of strategy}.
\newblock \emph{Journal of Electronic Imaging}, 19\penalty0 (1):\penalty0 011006, 2010.

\bibitem[Li et~al.(2022)Li, Wu, Tian, Li, Dong, and Shi]{LI2022307}
Aobo Li, Jinjian Wu, Shiwei Tian, Leida Li, Weisheng Dong, and Guangming Shi.
\newblock Blind image quality assessment based on progressive multi-task learning.
\newblock \emph{Neurocomputing}, 500:\penalty0 307--318, 2022.

\bibitem[Lin et~al.(2019)Lin, Hosu, and Saupe]{8743252}
Hanhe Lin, Vlad Hosu, and Dietmar Saupe.
\newblock Kadid-10k: A large-scale artificially distorted iqa database.
\newblock In \emph{2019 Eleventh International Conference on Quality of Multimedia Experience}, pages 1--3, 2019.

\bibitem[Liu et~al.(2019)Liu, Johns, and Davison]{Liu_2019_CVPR}
Shikun Liu, Edward Johns, and Andrew~J. Davison.
\newblock End-to-end multi-task learning with attention.
\newblock In \emph{Proceedings of the IEEE/CVF Conference on Computer Vision and Pattern Recognition}, pages 1871--1880, 2019.

\bibitem[Lou et~al.(2023)Lou, Chen, Xu, Zhou, Cao, Wang, and Huang]{lou2023refining}
Yiwei Lou, Yanyuan Chen, Dexuan Xu, Doudou Zhou, Yongzhi Cao, Hanpin Wang, and Yu Huang.
\newblock Refining the unseen: Self-supervised two-stream feature extraction for image quality assessment.
\newblock In \emph{2023 IEEE International Conference on Data Mining}, pages 1193--1198. IEEE, 2023.

\bibitem[Lou et~al.(2024{\natexlab{a}})Lou, Xu, Zhang, Zhang, Cao, Wang, and Huang]{lou2024mr}
Yiwei Lou, Dexuan Xu, Rongchao Zhang, Jiayu Zhang, Yongzhi Cao, Hanpin Wang, and Yu Huang.
\newblock Mr image quality assessment via enhanced mamba: A hybrid spatial-frequency approach.
\newblock In \emph{IEEE International Conference on Bioinformatics and Biomedicine}, pages 3561--3564. IEEE, 2024{\natexlab{a}}.

\bibitem[Lou et~al.(2024{\natexlab{b}})Lou, Zhang, Xu, Cao, Wang, and Huang]{lou2024no}
Yiwei Lou, Jiayu Zhang, Dexuan Xu, Yongzhi Cao, Hanpin Wang, and Yu Huang.
\newblock No-reference mri quality assessment via contrastive representation: Spatial and frequency domain perspectives.
\newblock In \emph{IEEE International Conference on Multimedia and Expo}, pages 1--6. IEEE, 2024{\natexlab{b}}.

\bibitem[Ma et~al.(2021)Ma, Han, Zhang, Fu, Zhou, and Hu]{NEURIPS2021_371bce7d}
Huan Ma, Zongbo Han, Changqing Zhang, Huazhu Fu, Joey~Tianyi Zhou, and Qinghua Hu.
\newblock Trustworthy multimodal regression with mixture of normal-inverse gamma distributions.
\newblock In \emph{Advances in Neural Information Processing Systems}, pages 6881--6893, 2021.

\bibitem[Ma et~al.(2017{\natexlab{a}})Ma, Duanmu, Wu, Wang, Yong, Li, and Zhang]{7752930}
Kede Ma, Zhengfang Duanmu, Qingbo Wu, Zhou Wang, Hongwei Yong, Hongliang Li, and Lei Zhang.
\newblock Waterloo exploration database: New challenges for image quality assessment models.
\newblock \emph{IEEE Transactions on Image Processing}, 26\penalty0 (2):\penalty0 1004--1016, 2017{\natexlab{a}}.

\bibitem[Ma et~al.(2017{\natexlab{b}})Ma, Liu, Liu, Wang, and Tao]{7934456}
Kede Ma, Wentao Liu, Tongliang Liu, Zhou Wang, and Dacheng Tao.
\newblock dipiq: Blind image quality assessment by learning-to-rank discriminable image pairs.
\newblock \emph{IEEE Transactions on Image Processing}, 26\penalty0 (8):\penalty0 3951--3964, 2017{\natexlab{b}}.

\bibitem[Ma et~al.(2019)Ma, Liu, Fang, and Simoncelli]{8803390}
Kede Ma, Xuelin Liu, Yuming Fang, and Eero~P. Simoncelli.
\newblock Blind image quality assessment by learning from multiple annotators.
\newblock In \emph{IEEE International Conference on Image Processing}, pages 2344--2348, 2019.

\bibitem[Ma et~al.(2020)Ma, Duanmu, Wang, Wu, Liu, Yong, Li, and Zhang]{8590800}
Kede Ma, Zhengfang Duanmu, Zhou Wang, Qingbo Wu, Wentao Liu, Hongwei Yong, Hongliang Li, and Lei Zhang.
\newblock Group maximum differentiation competition: Model comparison with few samples.
\newblock \emph{IEEE Transactions on Pattern Analysis and Machine Intelligence}, 42\penalty0 (4):\penalty0 851--864, 2020.

\bibitem[Madhusudana et~al.(2022)Madhusudana, Birkbeck, Wang, Adsumilli, and Bovik]{9796010}
Pavan~C. Madhusudana, Neil Birkbeck, Yilin Wang, Balu Adsumilli, and Alan~C. Bovik.
\newblock Image quality assessment using contrastive learning.
\newblock \emph{IEEE Transactions on Image Processing}, 31:\penalty0 4149--4161, 2022.

\bibitem[Meng et~al.(2023)Meng, Zhao, Tan, and Li]{MENG2023344}
Fanyong Meng, Dengyu Zhao, Chunqiao Tan, and Zijun Li.
\newblock Ordinal-cardinal consensus analysis for large-scale group decision making with uncertain self-confidence.
\newblock \emph{Information Fusion}, 93:\penalty0 344--362, 2023.

\bibitem[Mittal et~al.(2012)Mittal, Moorthy, and Bovik]{mittal2012no}
Anish Mittal, Anush~Krishna Moorthy, and Alan~Conrad Bovik.
\newblock No-reference image quality assessment in the spatial domain.
\newblock \emph{IEEE Transactions on Image Processing}, 21\penalty0 (12):\penalty0 4695--4708, 2012.

\bibitem[Mittal et~al.(2013)Mittal, Soundararajan, and Bovik]{6353522}
Anish Mittal, Rajiv Soundararajan, and Alan~C. Bovik.
\newblock Making a “completely blind” image quality analyzer.
\newblock \emph{IEEE Signal Processing Letters}, 20\penalty0 (3):\penalty0 209--212, 2013.

\bibitem[Pan et~al.(2022)Pan, Yuan, Lei, Fang, Shao, and Kwong]{pan2022vcrnet}
Zhaoqing Pan, Feng Yuan, Jianjun Lei, Yuming Fang, Xiao Shao, and Sam Kwong.
\newblock Vcrnet: Visual compensation restoration network for no-reference image quality assessment.
\newblock \emph{IEEE Transactions on Image Processing}, 31:\penalty0 1613--1627, 2022.

\bibitem[Ponomarenko et~al.(2015)Ponomarenko, Jin, Ieremeiev, Lukin, Egiazarian, Astola, Vozel, Chehdi, Carli, Battisti, and {Jay Kuo}]{PONOMARENKO201557}
Nikolay Ponomarenko, Lina Jin, Oleg Ieremeiev, Vladimir Lukin, Karen Egiazarian, Jaakko Astola, Benoit Vozel, Kacem Chehdi, Marco Carli, Federica Battisti, and C.-C. {Jay Kuo}.
\newblock Image database tid2013: Peculiarities, results and perspectives.
\newblock \emph{Signal Processing: Image Communication}, 30:\penalty0 57--77, 2015.

\bibitem[Qin et~al.(2022)Qin, Lou, Huang, Chen, and Yue]{qin2022ensemble}
Yuanze Qin, Yiwei Lou, Yu Huang, Rigao Chen, and Weihua Yue.
\newblock An ensemble deep learning approach combining phenotypic data and fmri for adhd diagnosis.
\newblock \emph{Journal of Signal Processing Systems}, 94\penalty0 (11):\penalty0 1269--1281, 2022.

\bibitem[Qu et~al.(2021)Qu, Chen, Chung, and Chen]{9505016}
Qiang Qu, Xiaoming Chen, Vera Chung, and Zhibo Chen.
\newblock Light field image quality assessment with auxiliary learning based on depthwise and anglewise separable convolutions.
\newblock \emph{IEEE Transactions on Broadcasting}, 67\penalty0 (4):\penalty0 837--850, 2021.

\bibitem[Radford et~al.(2019)Radford, Wu, Child, Luan, Amodei, Sutskever, et~al.]{radford2019language}
Alec Radford, Jeffrey Wu, Rewon Child, David Luan, Dario Amodei, Ilya Sutskever, et~al.
\newblock Language models are unsupervised multitask learners.
\newblock \emph{OpenAI blog}, 1\penalty0 (8):\penalty0 9, 2019.

\bibitem[Radford et~al.(2021)Radford, Kim, Hallacy, Ramesh, Goh, Agarwal, Sastry, Askell, Mishkin, Clark, Krueger, and Sutskever]{pmlr-v139-radford21a}
Alec Radford, Jong~Wook Kim, Chris Hallacy, Aditya Ramesh, Gabriel Goh, Sandhini Agarwal, Girish Sastry, Amanda Askell, Pamela Mishkin, Jack Clark, Gretchen Krueger, and Ilya Sutskever.
\newblock Learning transferable visual models from natural language supervision.
\newblock In \emph{Proceedings of the 38th International Conference on Machine Learning}, pages 8748--8763. PMLR, 2021.

\bibitem[Saha et~al.(2023)Saha, Mishra, and Bovik]{Saha_2023_CVPR}
Avinab Saha, Sandeep Mishra, and Alan~C. Bovik.
\newblock Re-iqa: Unsupervised learning for image quality assessment in the wild.
\newblock In \emph{Proceedings of the IEEE/CVF Conference on Computer Vision and Pattern Recognition}, pages 5846--5855, 2023.

\bibitem[Shafer(2016)]{SHAFER20167}
Glenn Shafer.
\newblock A mathematical theory of evidence turns 40.
\newblock \emph{International Journal of Approximate Reasoning}, 79:\penalty0 7--25, 2016.
\newblock 40 years of Research on Dempster-Shafer Theory.

\bibitem[Sheikh et~al.(2006)Sheikh, Sabir, and Bovik]{1709988}
H.R. Sheikh, M.F. Sabir, and A.C. Bovik.
\newblock A statistical evaluation of recent full reference image quality assessment algorithms.
\newblock \emph{IEEE Transactions on Image Processing}, 15\penalty0 (11):\penalty0 3440--3451, 2006.

\bibitem[Shen et~al.(2025)Shen, Zhou, Chen, Wei, Feng, Pu, and Jia]{shen2025image}
Wenhao Shen, Mingliang Zhou, Yu Chen, Xuekai Wei, Yong Feng, Huayan Pu, and Weijia Jia.
\newblock Image quality assessment: Investigating causal perceptual effects with abductive counterfactual inference.
\newblock In \emph{Proceedings of the Computer Vision and Pattern Recognition Conference}, pages 17990--17999, 2025.

\bibitem[Su et~al.(2020)Su, Yan, Zhu, Zhang, Ge, Sun, and Zhang]{Su_2020_CVPR}
Shaolin Su, Qingsen Yan, Yu Zhu, Cheng Zhang, Xin Ge, Jinqiu Sun, and Yanning Zhang.
\newblock Blindly assess image quality in the wild guided by a self-adaptive hyper network.
\newblock In \emph{Proceedings of the IEEE/CVF Conference on Computer Vision and Pattern Recognition}, pages 3667--3676, 2020.

\bibitem[Thurstone(2017)]{thurstone2017law}
Louis~L Thurstone.
\newblock A law of comparative judgment.
\newblock In \emph{Scaling}, pages 81--92. Routledge, 2017.

\bibitem[Tsai et~al.(2007)Tsai, Liu, Qin, Chen, and Ma]{tsai2007frank}
Ming-Feng Tsai, Tie-Yan Liu, Tao Qin, Hsin-Hsi Chen, and Wei-Ying Ma.
\newblock Frank: a ranking method with fidelity loss.
\newblock In \emph{Proceedings of the 30th annual international ACM SIGIR conference on Research and development in information retrieval}, pages 383--390, 2007.

\bibitem[Wang et~al.(2023)Wang, Xiong, Li, Suo, and Gao]{wang2023learning}
Xiaoqi Wang, Jian Xiong, Bo Li, Jinli Suo, and Hao Gao.
\newblock Learning hybrid representations of semantics and distortion for blind image quality assessment.
\newblock In \emph{ICASSP 2023-2023 IEEE International Conference on Acoustics, Speech and Signal Processing (ICASSP)}, pages 1--5. IEEE, 2023.

\bibitem[Wu et~al.(2018)Wu, Li, Ngan, and Ma]{7937920}
Qingbo Wu, Hongliang Li, King~N. Ngan, and Kede Ma.
\newblock Blind image quality assessment using local consistency aware retriever and uncertainty aware evaluator.
\newblock \emph{IEEE Transactions on Circuits and Systems for Video Technology}, 28\penalty0 (9):\penalty0 2078--2089, 2018.

\bibitem[Wu et~al.(2024)Wu, Liao, and Zhang]{WU2024102014}
Xingli Wu, Huchang Liao, and Chonghui Zhang.
\newblock Preference disaggregation analysis for sorting problems in the context of group decision-making with uncertain and inconsistent preferences.
\newblock \emph{Information Fusion}, 101:\penalty0 102014, 2024.

\bibitem[Xia et~al.(2025)Xia, He, Gao, and Hu]{xia2025blind}
Jili Xia, Lihuo He, Xinbo Gao, and Bo Hu.
\newblock Blind image quality assessment for in-the-wild images by integrating distorted patch selection and multi-scale-and-granularity fusion.
\newblock \emph{Knowledge-Based Systems}, 309:\penalty0 112772, 2025.

\bibitem[Yan et~al.(2019)Yan, Bare, and Tan]{8666733}
Bo Yan, Bahetiyaer Bare, and Weimin Tan.
\newblock Naturalness-aware deep no-reference image quality assessment.
\newblock \emph{IEEE Transactions on Multimedia}, 21\penalty0 (10):\penalty0 2603--2615, 2019.

\bibitem[Yao et~al.(2023)Yao, Cao, Feng, Cheng, and Han]{9718583}
Xiwen Yao, Qinglong Cao, Xiaoxu Feng, Gong Cheng, and Junwei Han.
\newblock Learning to assess image quality like an observer.
\newblock \emph{IEEE Transactions on Neural Networks and Learning Systems}, 34\penalty0 (11):\penalty0 8324--8336, 2023.

\bibitem[Yi et~al.(2025)Yi, Zheng, Liang, Dong, Fang, Wu, and Chen]{yi2025mac}
Fanghai Yi, Zehong Zheng, Zexiao Liang, Yihang Dong, Xiyang Fang, Wangyu Wu, and Xuhang Chen.
\newblock Mac-lookup: Multi-axis conditional lookup model for underwater image enhancement.
\newblock \emph{arXiv preprint arXiv:2507.02270}, 2025.

\bibitem[Ying et~al.(2020)Ying, Niu, Gupta, Mahajan, Ghadiyaram, and Bovik]{Ying_2020_CVPR}
Zhenqiang Ying, Haoran Niu, Praful Gupta, Dhruv Mahajan, Deepti Ghadiyaram, and Alan Bovik.
\newblock From patches to pictures (paq-2-piq): Mapping the perceptual space of picture quality.
\newblock In \emph{Proceedings of the IEEE/CVF Conference on Computer Vision and Pattern Recognition}, pages 3575--3585, 2020.

\bibitem[Zeng et~al.(2017)Zeng, Zhang, and Bovik]{zeng2017probabilistic}
Hui Zeng, Lei Zhang, and Alan~C Bovik.
\newblock A probabilistic quality representation approach to deep blind image quality prediction.
\newblock \emph{arXiv preprint arXiv:1708.08190}, 2017.

\bibitem[Zhang et~al.(2024)Zhang, Xu, Chen, Lou, and Huang]{zhang2024curriculum}
Jiayu Zhang, Dexuan Xu, Yanyuan Chen, Yiwei Lou, and Yue Huang.
\newblock Curriculum learning for self-iterative semi-supervised medical image segmentation.
\newblock In \emph{IEEE International Conference on Bioinformatics and Biomedicine}, pages 1342--1349. IEEE, 2024.

\bibitem[Zhang et~al.(2015)Zhang, Zhang, and Bovik]{7094273}
Lin Zhang, Lei Zhang, and Alan~C. Bovik.
\newblock A feature-enriched completely blind image quality evaluator.
\newblock \emph{IEEE Transactions on Image Processing}, 24\penalty0 (8):\penalty0 2579--2591, 2015.

\bibitem[Zhang et~al.(2020)Zhang, Ma, Yan, Deng, and Wang]{8576582}
Weixia Zhang, Kede Ma, Jia Yan, Dexiang Deng, and Zhou Wang.
\newblock Blind image quality assessment using a deep bilinear convolutional neural network.
\newblock \emph{IEEE Transactions on Circuits and Systems for Video Technology}, 30\penalty0 (1):\penalty0 36--47, 2020.

\bibitem[Zhang et~al.(2021)Zhang, Ma, Zhai, and Yang]{9369977}
Weixia Zhang, Kede Ma, Guangtao Zhai, and Xiaokang Yang.
\newblock Uncertainty-aware blind image quality assessment in the laboratory and wild.
\newblock \emph{IEEE Transactions on Image Processing}, 30:\penalty0 3474--3486, 2021.

\bibitem[Zhang et~al.(2023)Zhang, Zhai, Wei, Yang, and Ma]{Zhang_2023_CVPR}
Weixia Zhang, Guangtao Zhai, Ying Wei, Xiaokang Yang, and Kede Ma.
\newblock Blind image quality assessment via vision-language correspondence: A multitask learning perspective.
\newblock In \emph{Proceedings of the IEEE/CVF Conference on Computer Vision and Pattern Recognition (CVPR)}, pages 14071--14081, 2023.

\bibitem[Zheng et~al.(2024)Zheng, Luo, Zhou, Ling, and Yue]{10440553}
Limin Zheng, Yu Luo, Zihan Zhou, Jie Ling, and Guanghui Yue.
\newblock Cdinet: Content distortion interaction network for blind image quality assessment.
\newblock \emph{IEEE Transactions on Multimedia}, 26:\penalty0 7089--7100, 2024.

\bibitem[Zhou et~al.(2025)Zhou, Shen, Wei, Luo, Jia, Zhuang, and Jia]{zhou2025blind}
Mingliang Zhou, Wenhao Shen, Xuekai Wei, Jun Luo, Fan Jia, Xu Zhuang, and Weijia Jia.
\newblock Blind image quality assessment: Exploring content fidelity perceptibility via quality adversarial learning.
\newblock \emph{International Journal of Computer Vision}, pages 1--17, 2025.

\bibitem[Zhou et~al.(2024)Zhou, Tan, Zhao, and Yue]{zhou2024multitask}
Tianwei Zhou, Songbai Tan, Baoquan Zhao, and Guanghui Yue.
\newblock Multitask deep neural network with knowledge-guided attention for blind image quality assessment.
\newblock \emph{IEEE Transactions on Circuits and Systems for Video Technology}, 34\penalty0 (8):\penalty0 7577--7588, 2024.

\bibitem[Zhou et~al.(2022)Zhou, Xu, Xu, and Quan]{10.1145/3503161.3547982}
Zihan Zhou, Yong Xu, Ruotao Xu, and Yuhui Quan.
\newblock No-reference image quality assessment using dynamic complex-valued neural model.
\newblock In \emph{Proceedings of the 30th ACM International Conference on Multimedia}, page 1006–1015, New York, NY, USA, 2022. Association for Computing Machinery.

\end{thebibliography}
